%% file: GLIO2_arxiv.tex
\documentclass{ieeetj}
\input{glio2_preamble}
\input{glio2_arxiv_layout}

\begin{document}
\bstctlcite{IEEEexample:BSTcontrol}
\title{GLIO2: A GPU-Parallelized Tightly-Coupled LiDAR--Inertial--GNSS System for Robust and Real-Time Global Localization and Mapping}
\author{\mbox{Qi~Zhang, Xikun~Liu, Qijun~Qin, Xiangru~Wang, Junzhe~Wang, Naigui~Xiao, Jianhao~Jiao, and Weisong~Wen}}
\affil{Department of Aeronautical and Aviation Engineering, The Hong Kong Polytechnic University, Hong Kong SAR, China}
\corresp{Corresponding author: Weisong Wen (email: \href{mailto:welson.wen@polyu.edu.hk}{welson.wen@polyu.edu.hk}).}
\authornote{The work described in this paper was fully supported by a grant from the Research Grants Council of the Hong Kong Special Administrative Region, China (Project No. PolyU 25217524).}

\markboth{}{}

\begin{abstract}
Globally consistent, real-time state estimation in large-scale, perceptually degraded environments is essential for autonomous vehicles and aerial robots, and requires fusing LiDAR, inertial, and GNSS measurements. Existing fusion methods, however, share a scan-to-map front-end with two failure modes. First, each scan is aligned to an incrementally built map that drifts under degeneracy, and once the estimate diverges the error is irrecoverable. Second, even without divergence, a registration biased by dynamic objects or wrong correspondences is propagated as a single pose constraint with an over-confident covariance, leaving its correspondences unavailable for GNSS to re-weight or relinearize. We propose GLIO2, a tightly-coupled LiDAR--Inertial--GNSS system whose GPU-parallel front-end jointly optimizes scan-to-multiscan LiDAR, IMU pre-integration, and raw GNSS measurements in a single sliding-window factor graph, sustaining real-time operation on edge hardware. A complementary offline back-end reuses the same cached factors to refine the entire trajectory in batch, completing the $30$-min, $4.51$-km UrbanNav~\textit{Whampoa} sequence in about $24$\,s. Across three public benchmarks (UrbanNav, MARS-LVIG, M3DGR) and self-collected UAV and vehicle data, GLIO2 attains the best overall accuracy among evaluated systems. On a $5.66$-km bridge traversed at up to $96$\,km/h, where every competing baseline diverges under LiDAR degeneracy, it maintains $1.6$\,m horizontal accuracy. On an NVIDIA Jetson Orin~NX, the full pipeline runs at about $25$\,Hz ($39.60$\,ms per scan). The source code and datasets will be released.
\end{abstract}


\maketitle

\section{Introduction}
\label{sec:intro}

\begin{figure}[!t]
  \centering
  \includegraphics[width=\columnwidth]{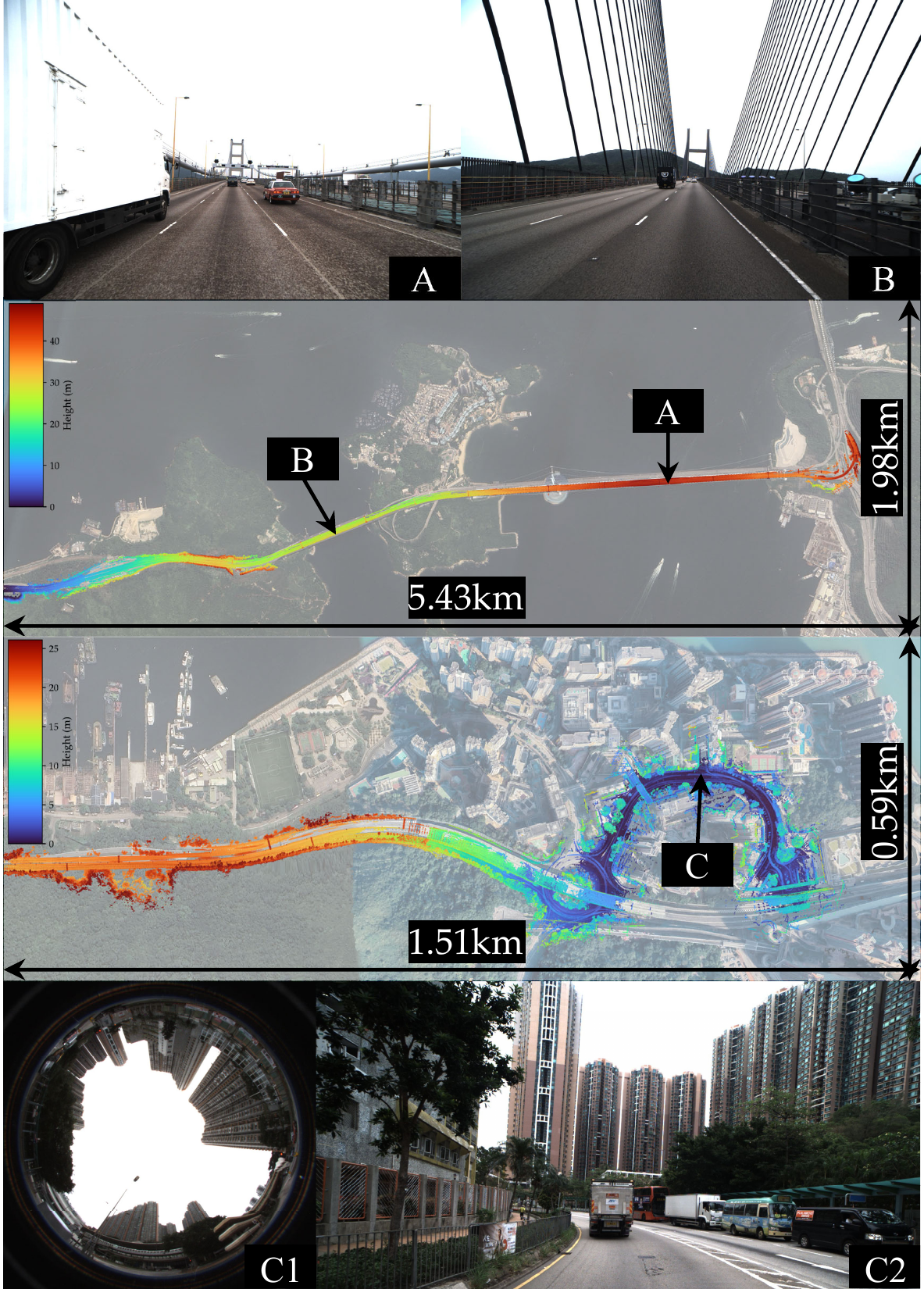}
  \caption{GLIO2 in two challenging self-collected scenes, with the point-cloud map (colored by height) projected onto satellite imagery. \textbf{Top}: the $5.66$\,km \textit{Bridge}, crossed at up to $96$\,km/h, where the straight bridge and the heavy moving traffic cause LiDAR degeneracy along the driving direction (front views~A, B). \textbf{Bottom}: the $1.44$\,km dense-canyon \textit{Urban} route, where high-rise buildings block and reflect the GNSS signals (location~C, with fisheye sky view~C1 and front view~C2). In each scene GLIO2 keeps the global map continuous and aligned with the imagery despite the complementary LiDAR and GNSS degradation.}
  \label{fig:map_deg1}
\end{figure}

\IEEEPARstart{A}{n} autonomous vehicle in a dense urban canyon and a micro-aerial vehicle surveying open farmland face the same estimation problem: the state must be globally consistent over the whole mission, locally smooth at control rate, robust to sensor degradation, and computed in real time on the limited compute carried onboard. LiDAR, inertial measurement units (IMUs), and global navigation satellite systems (GNSS) each meet part of these demands and fail another.

LiDAR-inertial odometry (LIO)~\cite{fastlio2,glim,lio-sam} exploits the rich structural information in point clouds to produce high-frequency, locally smooth, and locally accurate motion estimates together with dense 3D geometric maps. The IMU predicts inter-scan motion to deskew each scan and bridge the gaps between LiDAR frames, though it drifts rapidly when integrated alone. Relying primarily on local geometric constraints, however, LIO accumulates drift over long trajectories and degrades severely under geometric degeneracy (weak, repetitive, or insufficient structure such as open fields, long tunnels, and flat or sparsely featured scenes), where the registration loses observability along its under-constrained directions~\revwording{\cite{zhang2016degeneracy}}.

GNSS, in contrast, provides drift-free absolute positioning in a globally referenced Earth-fixed frame, bounding the global error that LIO cannot~\cite{GNSS_IMU_FGO_Welson}. Yet its measurements arrive at only $1$--$10$\,Hz and are corrupted by multipath, non-line-of-sight (NLOS) reception, and outright outages in the urban canyons, tunnels, and indoor spaces a robot must traverse, where a single contaminated observation can inject errors of tens of meters.

The two modalities are thus complementary: GNSS anchors the estimate to an Earth-fixed frame and bounds the global drift that LIO accumulates, while LIO carries the estimate through GNSS outages and down-weights GNSS outliers with its more precise local geometry. Fusing LiDAR, IMU, and GNSS in a single tightly-coupled factor graph~\cite{isam2} exploits this complementarity, jointly relinearizing the raw heterogeneous measurements as new information arrives~\cite{glio,ligo,GNSS_IMU_FGO_Welson}. Building such an estimator that is at once globally and locally consistent, robust under sensor degradation, and real-time on edge platforms, however, raises three central challenges.

\textit{a) Front-end fragility}: LiDAR--GNSS systems such as LIO-SAM~\cite{lio-sam}, LIGO~\cite{ligo}, GIL~\cite{li2021gil}, and GLIO~\cite{glio} all build their LiDAR estimate on a scan-to-map (S2M) front-end, which fuses past scans into a single deterministic map and aligns each new scan to it. In scenes such as a sea-crossing bridge, a flat open plain, or a UAV flying close to a repetitive high-rise facade, severe geometric degeneracy rapidly destabilizes this scan-to-map LiDAR localization, so the front-end can no longer produce a correct estimate; once it diverges, the back-end GNSS estimate cannot be recovered either. Even GLIO's offline scan-to-multiscan (S2MS) batch is insufficient, as it still requires a non-divergent front-end as its initial guess. The first challenge, then, is the inherent limitation of the front-end design adopted by these mainstream methods.

\textit{b) Frozen, over-confident LiDAR}: Mainstream S2M methods first reduce each scan to a fixed LiDAR constraint (a $6$-DoF pose-with-covariance prior in LIGO~\cite{ligo}, a relative-pose factor in LIO-SAM~\cite{lio-sam}) and fuse it with GNSS. In urban scenes, however, dynamic objects, passing traffic, and changing structure create wrong correspondences that bias this constraint. It then enters the fusion under an over-confident covariance that outweighs the GNSS correction, and because the past point clouds are never re-associated or relinearized, the bias is frozen into the trajectory rather than corrected. The second challenge is thus to admit LiDAR into the GNSS fusion as \revwording{raw relative constraints, so that LiDAR information constrains the local geometry without being imposed as a potentially overconfident prior on the global frame. Active correspondences can then be re-associated and their residuals relinearized as the joint estimate updates.}

\textit{c) Real-time cost}: \revwording{Addressing the first two challenges motivates replacing S2M with S2MS registration and keeping} the raw point residuals in one joint optimization, so that LiDAR and GNSS relinearize together. The price is computational. Each scan forms dense voxel correspondences against several keyframes, and every nonlinear iteration must recompute all correspondences, fused covariances, and Jacobians as the keyframe poses move. \revwording{Even on a high-performance CPU (e.g., the Intel Core Ultra~7 265KF), each per-scan sliding-window update must relinearize hundreds of thousands of point residuals across the active keyframes; on denser scans, our CPU implementation already consumes the full $100$\,ms budget of a $10$\,Hz LiDAR and peaks at $200$--$300$\,ms (Sec.~\ref{sec:ablation}). GLIO~\cite{glio}, for example, performs its dense S2MS and raw-GNSS joint optimization in the offline back-end, while its online stage retains the S2M front-end described above.} The third challenge is thus to sustain this dense, raw-domain joint optimization online, \revwording{within a per-scan budget that our CPU implementation cannot consistently meet.}

Motivated by these observations, we present GLIO2, a unified tightly-coupled LiDAR--Inertial--GNSS system. A GPU-parallel front-end fuses dense LiDAR, IMU, and raw GNSS in one real-time sliding-window graph, and an offline batch back-end refines the trajectory by reusing the same factors. The first two contributions respond to the three challenges; the third validates and releases the system.

The contributions of this work are as follows.
\begin{itemize}
    \item \textbf{A unified online LiDAR--Inertial--GNSS front-end.} GLIO2 fuses scan-to-multiscan LiDAR, raw double-differenced pseudorange and Doppler, and IMU pre-integration in a single real-time sliding-window graph (Sec.~\ref{sec:system_overview}). \revwording{Each scan contributes relative scan-to-keyframe constraints without being reduced to a fixed-map LiDAR pose prior.} The raw correspondences are re-associated and relinearized at every iteration, \revwording{allowing registration errors to be revised as the joint estimate updates.} Tightly coupling both modalities on the raw measurements keeps the estimate robust to LiDAR degeneracy and to GNSS dropouts (Sec.~\ref{sec:ablation}).

    \item \textbf{Real-time onboard operation with a fast offline back-end.} \revwording{GPU-parallel per-point linearization enables the joint LiDAR--Inertial--GNSS optimization to run at the LiDAR rate on an NVIDIA Jetson Orin~NX.} \revwording{The offline back-end reuses compact relative-pose LiDAR factors generated online to refine the full trajectory without re-registering the point clouds. It improves trajectory accuracy and runs approximately $12$--$20\times$ faster than the GLIO back-end on the three tested sequences (Sec.~\ref{sec:batch}, Fig.~\ref{fig:backend_bubble}).}

    \item \textbf{Validation and open-source release.} On three public benchmarks (UrbanNav~\cite{hsu2021urbannav}, MARS-LVIG~\cite{MARS-LVIG}, M3DGR~\cite{zhang2025towards}) and self-collected UAV and vehicle data, under both GNSS NLOS/multipath and LiDAR degeneracy, \revwording{GLIO2 achieves the lowest 3D ATE RMSE among the evaluated LiDAR--Inertial--GNSS systems after batch refinement} (Sec.~\ref{sec:experiments}). A single-receiver variant runs without a base station (Appendix~\ref{app:single_receiver}). The source code and self-collected datasets will be released.
\end{itemize}

\revwording{To our knowledge, GLIO2 is the first system to jointly optimize dense scan-to-multiscan LiDAR factors, IMU pre-integration, and raw GNSS observations fully online at the LiDAR rate on an embedded GPU. We evaluate it on eleven sequences drawn from three public benchmarks (UrbanNav~\cite{hsu2021urbannav}, MARS-LVIG~\cite{MARS-LVIG}, M3DGR~\cite{zhang2025towards}) and four self-collected UAV and vehicle runs, spanning dense urban canyons, indoor--outdoor transitions, a $5.66$\,km sea-crossing bridge, and feature-poor farmland. The online front-end completes all eleven sequences, with $3$D ATE RMSEs of $3.54$\,m on UrbanNav-\textit{Tst}, $7.76$\,m on \textit{Whampoa}, $1.98$\,m on our dense-canyon \textit{Urban} route, and $3.59$\,m on the LiDAR-degenerate \textit{Bridge}, where every evaluated LiDAR-based baseline diverges. With offline batch refinement, these errors decrease to $2.58$, $2.99$, $1.33$, and $2.55$\,m, respectively. Across all eleven sequences, the batch solution achieves the lowest $3$D ATE RMSE among the evaluated LiDAR--inertial--GNSS systems. The front-end averages $39.6$--$55.5$\,ms per scan across the tested LiDARs on an NVIDIA Jetson Orin~NX, below the $100$\,ms scan period, while the offline back-end re-optimizes the full $30$-min \textit{Whampoa} trajectory in $24.2$\,s.}

\section{Related Work}
\label{sec:related_work}

\subsection{LiDAR Registration Paradigms}
\label{subsec:lidar_factors}
Scan matching provides the geometric constraints in LiDAR odometry, and the way a scan is matched determines the form of constraint it produces and thus what LiDAR contributes to the joint estimator.

\textbf{\textit{Scan-to-map}} (S2M) registration is adopted by most real-time LiDAR-inertial odometry (LIO) systems, which registers each new scan against an incrementally built reference map (e.g., a surfel or implicit-surface map~\cite{deschaud2018imls,behley2018suma}) and produces a single absolute pose~\cite{fastlio2,point-lio,lio-sam}. Building on classical pairwise scan registration~\cite{icp,loam}, S2M is efficient and locally accurate. However, it treats the fused map as a fixed reference, so potentially erroneous earlier map points are never corrected by later scans, and each scan is summarized by one absolute pose rather than by its raw correspondences.

\textbf{\textit{Scan-to-multiscan}} (S2MS) registration instead keeps several keyframe poses as optimization variables and jointly refines them through relative constraints. The estimated poses update the world-frame geometry, while each keyframe's point cloud remains in its local frame. S2MS builds on the joint multi-scan alignment of Lu and Milios~\cite{lu1997globally} and nScan-matching~\cite{nscan_matching}, and modern formulations bring a distribution-to-distribution registration metric (GICP~\cite{gicp}, NDT~\cite{biber2003ndt}, and its voxelized form VGICP~\cite{vgicp}) into a multi-scan factor graph. The main cost is computational: as the keyframe poses move, the correspondences, residuals, and Jacobians must be re-associated and relinearized for many scan pairs at every iteration, so dense S2MS has traditionally run at low rate or offline, as in the global LiDAR bundle adjustment of BALM~\cite{liu2021balm} and the GNSS-coupled global batch optimization of GLIO~\cite{glio}. Lightweight and GPU-accelerated formulations reduce this cost: FF-LINS~\cite{fflins} uses a frame-to-frame association on a solid-state LiDAR, while Koide \textit{et~al.} apply GPU acceleration to dense factors for real-time LiDAR-inertial odometry~\cite{koide2022globally} and faster global mapping~\cite{koide2021globally}, an approach consolidated in GLIM~\cite{glim}.
\subsection{GNSS Observables}
\label{subsec:gnss_observables}
A GNSS receiver outputs three raw observables, the pseudorange (PSR), the Doppler (DOP), and the carrier phase (CP). Prior work fuses these observables in different ways.

The simplest is \emph{Single Point Positioning} (SPP), which estimates an absolute position from a single receiver's pseudoranges, modeling the ionospheric and tropospheric delays and the satellite and receiver clock errors~\cite{enge1994global}. GVINS~\cite{cao2022gvins}, the raw-GNSS visual-inertial estimator of Liu \textit{et~al.}~\cite{liu2021rawgnssvio} and gnssFGO~\cite{zhang2024gnssfgo} tightly couple raw pseudorange and Doppler, the Doppler providing a global velocity observation. \emph{Precise Point Positioning} (PPP) applies precise orbit and clock products and estimates the receiver clock and tropospheric delay, removing the first-order ionosphere either by the ionosphere-free combination (GIL~\cite{li2021gil}) or with external ionosphere maps (P3-LINS~\cite{li2023p3-lins}). \emph{Differential} methods instead suppress most of these errors directly: double-differencing (DD) the pseudorange against a reference station~\cite{odijk2017differential} cancels the satellite and receiver clock errors and largely removes the spatially correlated atmospheric delays over a short baseline, leaving meter-level constraints, as in GLIO~\cite{glio}. The carrier phase is far more precise but carries an integer ambiguity; the dominant strategy resolves the double-differenced ambiguity~\cite{chang2005mlambda} in a between-station/between-satellite formulation (\emph{Real-Time Kinematic}, RTK), as in GLINS~\cite{glins}, GIVL-SLAM~\cite{wang2024givl}, and FGO-GIL~\cite{li2023fgo}, reaching centimeter accuracy. An alternative differences the carrier phase across epochs (\emph{Time-Differenced Carrier Phase}, TD-CP); since the integer ambiguity is time-invariant, this differencing cancels it when no cycle slip occurs, leaving a smooth epoch-to-epoch motion constraint, as in LIGO~\cite{ligo}.

\subsection{LiDAR--Inertial--GNSS Integration Systems}
\label{subsec:lig_systems}
A LiDAR--Inertial--GNSS system couples LiDAR, IMU, and GNSS measurements, which supplies the globally referenced, drift-free positioning that LiDAR and inertial sensing alone cannot. We group existing systems by how GNSS is coupled to the estimator: loosely-coupled, tightly-coupled filter-based, and tightly-coupled factor-graph methods (Tab.~\ref{tab:related_comparison}).
 
\textbf{\textit{Loosely-coupled}} methods combine LiDAR and GNSS at the state-estimation level: a position solved from GNSS enters as a constraint in the LiDAR-inertial estimator. LIO-SAM~\cite{lio-sam} and LIO-Fusion~\cite{wu2023lio} add the GPS/RTK solution as a position factor in the graph, with LiDAR entering as the relative pose of a LIO front-end, and He \textit{et~al.}~\cite{he2020integrated} add GNSS-derived poses to a mapping graph as pose factors. Others apply quality management to the RTK output and fuse only the validated fixed solutions~\cite{zhang2021continuous,chang2019gnss}, or extend this state-level fusion to GNSS together with multi-sensors~\cite{moore2016generalized,mascaro2018gomsf,nubert2022graph,qin2019general}. This bounds the long-term LiDAR drift, but reduces GNSS to a single pose: one epoch corrupted by multipath or NLOS can drag the whole solution off, or be rejected outright even when many of its satellites remain usable.

\revwording{\textbf{\textit{Tightly-coupled filter-based}} methods fuse the raw GNSS observables in a recursive filter.} GIL~\cite{li2021gil}, P3-LINS~\cite{li2023p3-lins}, and the GNSS/IMU/LiDAR estimator of Li~\textit{et~al.}~\cite{li2022tightly} couple raw PPP measurements with IMU and LiDAR and reach precise positioning without a reference station. GIL and P3-LINS adopt an error-state EKF, Li~\textit{et~al.} use a very short window MSCKF. All three share the same drawback: a recursive filter absorbs each measurement once and marginalizes past states rather than re-optimizing them. This is especially damaging for GNSS, whose measurements are noisy and often corrupted by NLOS and multipath. A single committed outlier biases the current state, and from there it propagates into every estimate that follows.
 
\textbf{\textit{Tightly-coupled factor-graph}} methods fuse the raw observables in a sliding-window factor graph, keeping a window of states available for relinearization, so GNSS outliers can be re-weighted as later measurements arrive. FGO-GIL~\cite{li2023fgo} and GIVL-SLAM~\cite{wang2024givl} both register each LiDAR keyframe against a fixed submap and directly fuse with GNSS RTK in a factor graph optimization. Because the submap is held fixed during optimization, scan-to-submap matching returns an absolute pose in the map frame, which enters the factor graph as an absolute constraint on the state. GNSS RTK provides an absolute constraint as well. The two anchor the state to different references and differ in accuracy, so their fusion comes down to weighting one absolute constraint against the other, \revwording{with their contributions depending on relative weighting and consistent frame alignment.}

gnssFGO~\cite{zhang2024gnssfgo}, Beuchert \textit{et~al.}~\cite{10161522}, and LIGO~\cite{ligo} all adopt a hierarchical tightly-coupled design, optimizing a relative pose from a LiDAR-odometry front-end together with raw GNSS in a sliding-window graph, which avoids the absolute-constraint competition. gnssFGO uses a scan-to-scan front-end in a continuous-time formulation, Beuchert \textit{et~al.} use ICP scan-to-submap, and LIGO runs a Point-LIO~\cite{point-lio} EKF front-end. In all of them the LiDAR registration is solved once and frozen into the resulting pose factor, and the back-end never relinearizes the raw point cloud. So when the scans are geometrically degenerate, or a moving object causes a wrong registration, the pose carries an error, yet it is passed to the back-end with an over-confident covariance; the GNSS observations that remain informative can no longer correct it, and the estimate either diverges or keeps an error it cannot recover from.

GLINS~\cite{glins} \revwording{uses a landmark-based tightly coupled formulation.} Instead of registering against a fixed map, GLINS promotes stable planes from the LiDAR voxel map to explicit landmark states and jointly optimizes them with the poses, IMU biases, and raw GNSS RTK observations in a single sliding-window graph. Each scan therefore contributes a relative constraint, and the map co-evolves with the global GNSS correction. \revwording{This joint refinement also creates a possible path for error propagation: if a GNSS bias that survives outlier rejection distorts the estimated landmark geometry, subsequent scan-to-landmark constraints can carry that distortion into later pose estimates.}

GLIO~\cite{glio} comes closest to our goal: a real-time (RT) S2M front-end feeds an offline back-end that fuses S2MS LiDAR with raw double-differenced PSR and DOP in a global batch optimization, yielding strong global consistency. But that consistency is obtained only offline. The S2MS batch still requires a non-divergent S2M front-end as its initial guess, so once the front-end diverges under sustained degeneracy the batch has nothing valid to refine; and because the back-end re-registers the point clouds, it is slow.

GLIO2 addresses these problems within a single tightly-coupled factor graph that fuses raw double-differenced pseudorange and Doppler with dense scan-to-multiscan LiDAR over one large sliding window. First, on the LiDAR side, \revwording{GLIO2 retains voxel maps in individual keyframe frames without maintaining a fused, fixed world-frame reference map. It forms direct dense correspondences between the scans in the window and exposes them as relative constraints without introducing a fixed-map LiDAR pose prior.} Second, because these correspondences and their residuals are re-associated and relinearized at every iteration, a registration error from geometric degeneracy or a moving object \revwording{can be revised as the joint estimate updates while the associated scans remain active in the window.} \revwording{Third, GLIO2 builds on the GPU-accelerated scan-matching factors of GLIM~\cite{glim} and extends dense LiDAR--inertial smoothing with raw GNSS factors and jointly estimated global-frame states, allowing the combined optimization to run online at the LiDAR rate.}
\begin{table}[!tb]
\centering
\caption{Summary of the Most Relevant Methods for LiDAR--Inertial--GNSS Integration Systems.}
\label{tab:related_comparison}
\setlength{\tabcolsep}{3pt}
\renewcommand{\arraystretch}{1.2}
\resizebox{\columnwidth}{!}{%
\begin{tabular}{llllc}
\toprule
System & LiDAR Factor & GNSS & Coupling & RT \\
\midrule
LIO-SAM~\cite{lio-sam}                   & S2M        & GNSS Solution              & loose FGO              & \textbf{\checkmark} \\
LIO-Fusion~\cite{wu2023lio}              & S2M        & GNSS Solution              & loose FGO              & \textbf{\checkmark} \\
He~\textit{et~al.}~\cite{he2020integrated} & S2M      & GNSS Solution              & loose FGO              & $\times$ \\
\midrule
GIL~\cite{li2021gil}                     & S2M        & PPP: PSR+CP           & tight EKF              & $\times$ \\
P3-LINS~\cite{li2023p3-lins}             & S2M        & PPP: PSR+CP+DOP       & tight EKF              & $\times$ \\
Li~\textit{et~al.}~\cite{li2022tightly}  & S2MS& PPP: PSR+CP & tight MSCKF & \textbf{\checkmark} \\
LIGO~\cite{ligo}                         & S2M    & PSR+DOP+TD-CP         & tight FGO              & \textbf{\checkmark} \\
Beuchert~\textit{et~al.}~\cite{10161522} & S2M       & PSR+DD-CP           & tight FGO              & \textbf{\checkmark} \\
gnssFGO~\cite{zhang2024gnssfgo}          & Scan-to-Scan       & PSR+DOP        & tight FGO          & \textbf{\checkmark} \\
FGO-GIL~\cite{li2023fgo}                 & S2M        & RTK: PSR+CP           & tight FGO              & $\times$ \\
GLINS~\cite{glins}                       & planar landmark    & RTK: PSR+CP+DOP       & tight FGO              & \textbf{\checkmark} \\
GLIO~\cite{glio}                         & S2M+S2MS (offline) & DD PSR+DOP           & tight FGO              & $\times$ \\
GIVL-SLAM~\cite{wang2024givl}            & S2M        & RTK: PSR+CP           & tight FGO              & $\times$ \\
\midrule
\textbf{GLIO2 (ours)}                    & \textbf{dense S2MS} & \textbf{DD PSR+DOP} & \textbf{tight FGO} & \textbf{\checkmark} \\
\bottomrule
\end{tabular}}
\end{table}

\section{System Overview}
\label{sec:system_overview}

\begin{figure*}[!t]
  \centering
  \includegraphics[width=\textwidth]{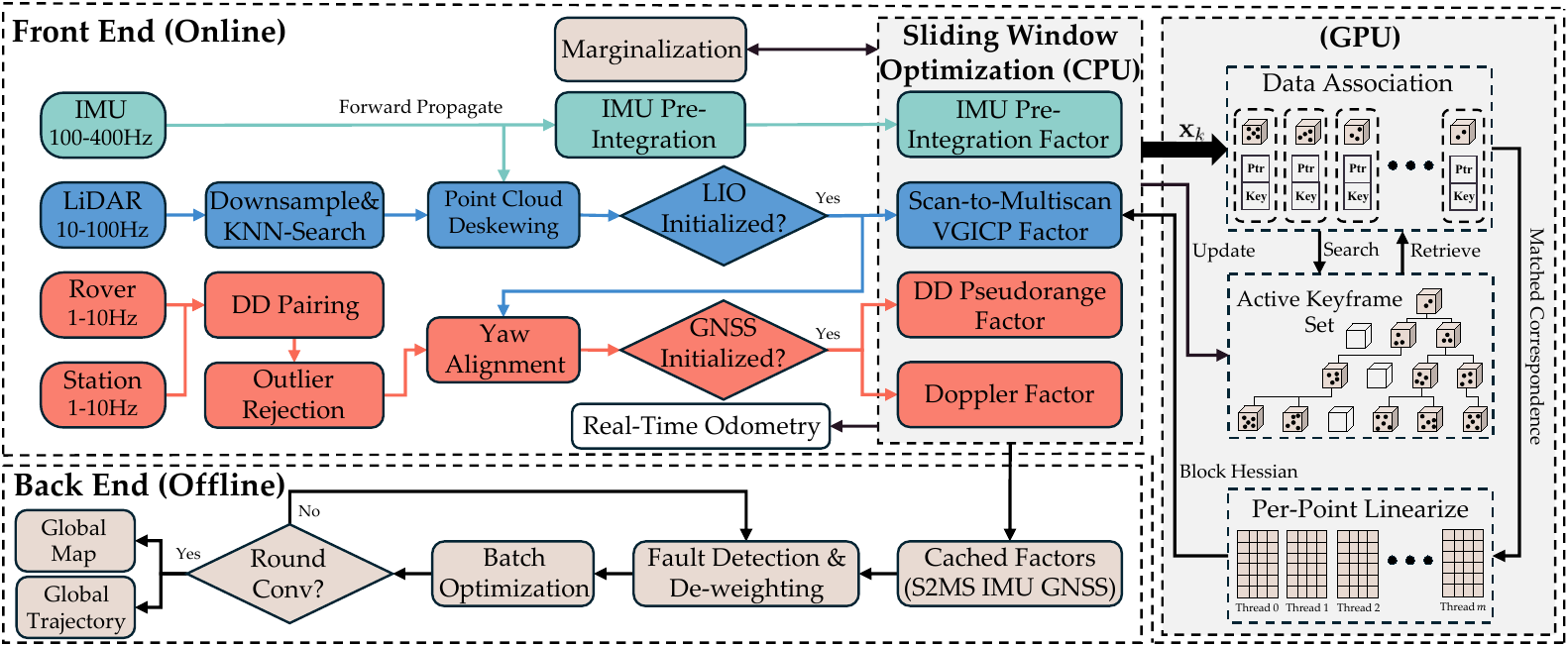}
  \caption{System overview of GLIO2. The front-end IMU-deskews each LiDAR scan and bootstraps by LIO initialization, where the IMU pre-integration and scan-to-multiscan (S2MS) voxelized-GICP (VGICP) LiDAR factors supply motion and local geometry; GNSS is then aligned to the LIO trajectory, and its double-differenced (DD) pseudorange and Doppler factors enter the graph. The sliding-window graph is solved on the CPU at the LiDAR rate with the dense per-point linearization on the GPU: at each linearization point $\mathbf{x}_k$, the points are re-associated against the hash-indexed keyframe voxel maps and the block Hessians returned, the CPU solves for the next $\mathbf{x}_k$ until convergence, and the converged poses update the active keyframe set while out-of-window states are marginalized. The offline back-end reuses the cached relative factors, applies several rounds of GNSS fault detection and de-weighting (FDD), and batch-optimizes the whole trajectory.}
  \label{fig:system_overview}
\end{figure*}

\subsection{System Architecture}
The overview of GLIO2 is shown in Fig.~\ref{fig:system_overview}. GLIO2 fuses three asynchronous streams, LiDAR, IMU, and raw GNSS from the rover and a reference station, in a single tightly-coupled factor graph (Sec.~\ref{sec:methodology}). The methodology comprises the factor graph formulation (Sec.~\ref{sec:factor_graph}), the four measurement factors (Sec.~\ref{sec:dd_psr}--\ref{sec:lidar_factor}), the active keyframe set (Sec.~\ref{sec:active_map}), marginalization (Sec.~\ref{sec:marginalization}), and initialization (Sec.~\ref{sec:initialization}) of the online front-end, and the global batch optimization of the back-end (Sec.~\ref{sec:batch}).

\subsection{Notation and Coordinate Frames}
\begin{figure}[t]
\centering
\includegraphics[width=\columnwidth]{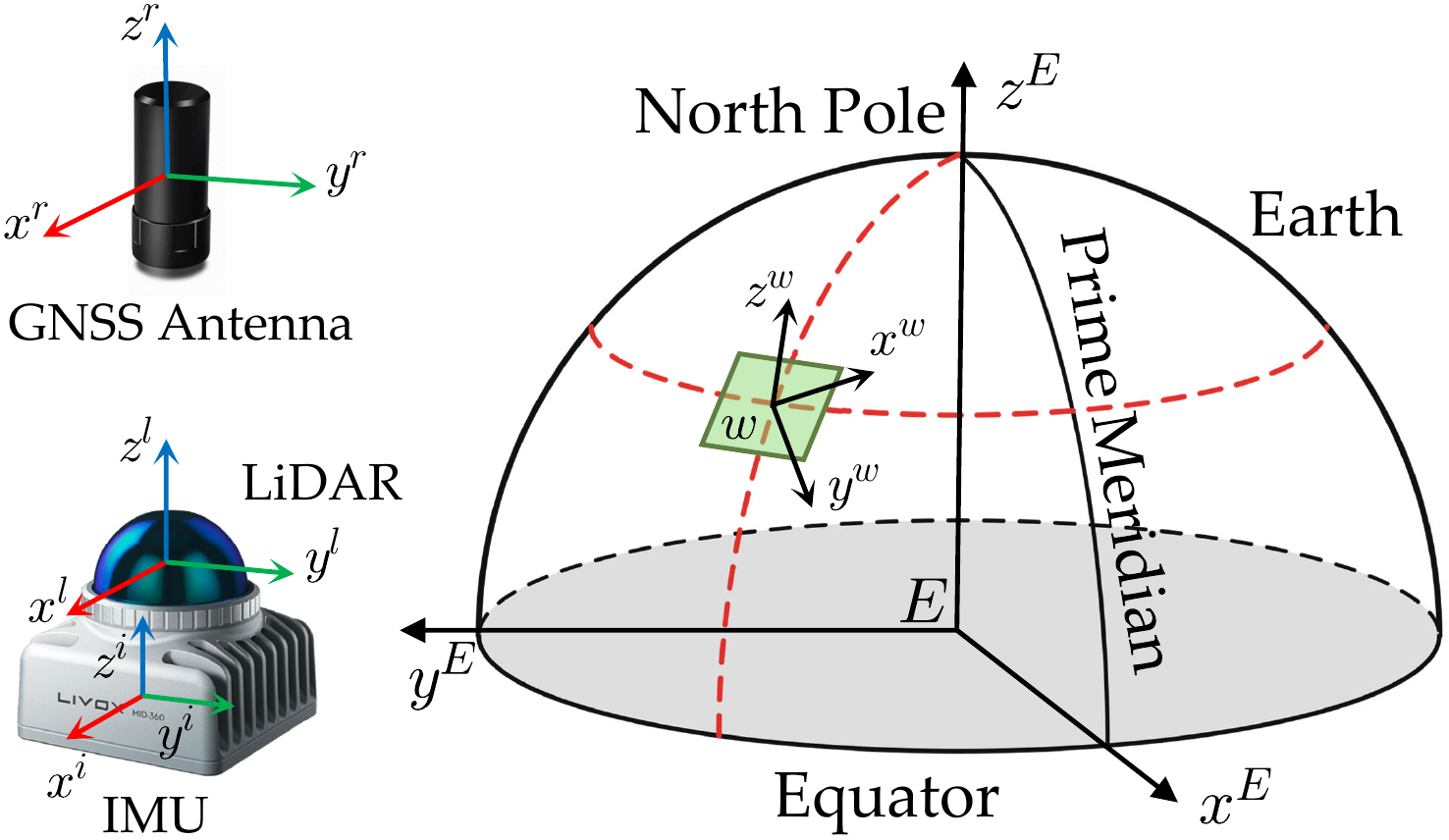}
\caption{The five coordinate frames of GLIO2: the Earth-centered ECEF frame ($E$), the local-world frame ($w$), and the LiDAR ($l$), IMU ($i$), and GNSS-receiver ($r$) frames on the platform. The LiDAR and IMU share a fixed extrinsic; the local-world frame relates to ECEF through the estimated extrinsic ${}_{w}^{E}\mathbf{T}$.}
\label{fig:coordinate}
\end{figure}

To integrate GNSS, LiDAR, and IMU measurements, the proposed framework employs five spatial reference frames. A 3D quantity $(\cdot)$ expressed in the Earth-Centered, Earth-Fixed (ECEF), local-world, LiDAR, IMU, or GNSS-receiver frame \revwording{is denoted by the left superscripts} ${}^{E}(\cdot)$, ${}^{w}(\cdot)$, ${}^{l}(\cdot)$, ${}^{i}(\cdot)$, and ${}^{r}(\cdot)$, respectively. As shown in Fig.~\ref{fig:coordinate}, the ECEF frame $E$ is centered at the Earth, the local-world frame $w$ is at the start of the trajectory, and the LiDAR ($l$), IMU ($i$), and GNSS-receiver ($r$) frames are rigidly mounted on the platform. Throughout this paper, the notation ${}_{b}^{a}(\cdot)$ denotes a quantity of frame $b$ expressed in frame $a$. A lone left superscript ${}^{a}(\cdot)$ indicates that the quantity is expressed in frame $a$. A right subscript $(\cdot)_{k}$ marks the state or measurement at discrete time step $k$. Tab.~\ref{tab:nomenclature} summarizes the recurring notation; factor-specific symbols are defined where they first appear.

\begin{table}[!t]
\caption{Nomenclature of Coordinates and System States}
\label{tab:nomenclature}
\centering
\begin{tabular}{ll}
\toprule
Notations & Explanation \\
\midrule
${}^{E}(\cdot), {}^{w}(\cdot), {}^{l}(\cdot)$ & ECEF, local-world, and LiDAR coordinate frames \\
${}^{i}(\cdot), {}^{r}(\cdot)$ & IMU and GNSS receiver coordinate frames \\
${}_{b}^{a}(\cdot)$ & Variable of frame $b$ represented in frame $a$ \\
$(\cdot)_{k}$ & Physical quantity at discrete time step $k$ \\
$\lfloor\cdot\rfloor_{\times},\;(\cdot)^{\wedge}$ & Skew operator; hat map, $\mathbb{R}^{6}\!\to\!\mathfrak{se}(3)$ \\
$\boxplus,\;\boxminus,\;\oplus$ & Manifold plus/minus; orthogonal direct sum \\
$\hat{(\cdot)}$ & Estimate (posterior) of a quantity $(\cdot)$ \\
$\mathcal{X}_{k}$ & Full states optimized in the unified factor graph \\
$\mathbf{R}, \mathbf{p}, \mathbf{v}$ & Rotation matrix, translation vector, and velocity \\
$\mathbf{b}_{a}, \mathbf{b}_{g}$ & Accelerometer and gyroscope biases in $\mathbb{R}^{3}$ \\
${}^{w}\mathbf{g}$ & Gravity in local-world frame ($\|{}^{w}\mathbf{g}\|{=}g_{0}$) \\
$\boldsymbol{\dot{\delta t}}$ & GNSS receiver clock drift rates ($\mathbb{R}^{4}$) \\
${}^{i}\mathbf{l}_{r}$ & Lever arm, IMU to GNSS antenna \\
$\rho, D$ & GNSS pseudorange and Doppler observations \\
$\mathbf{e}_{r}^{s}$ & Unit line-of-sight vector, receiver to satellite $s$ \\
\bottomrule
\end{tabular}
\end{table}

\subsection{System State Representation}
\label{sec:system_state}
In our tightly-coupled framework, the full state $\mathcal{X}_{k}$ at time step $k$ stacks two window-shared quantities (the extrinsic transformation between the local-world and ECEF frames and the local-world gravity) with a sliding window of $W$ time-varying navigation states,
\begin{equation}
\mathcal{X}_{k} = \bigl\{\, {}_{w}^{E}\mathbf{T}_{k},\; {}^{w}\mathbf{g},\;
\mathbf{x}_{k-W+1}^{f},\, \mathbf{x}_{k-W+2}^{f},\,
\ldots,\; \mathbf{x}_{k}^{f}\,\bigr\},
\end{equation}
where ${}_{w}^{E}\mathbf{T}_{k} = ({}_{w}^{E}\mathbf{R}_{k},\; {}_{w}^{E}\mathbf{p}_{k}) \in \mathrm{SE}(3)$ is the extrinsic that aligns the local-world frame with the ECEF frame and $W$ is the fixed-lag window length. The gravity ${}^{w}\mathbf{g}$ is a single quantity shared by every node in the window: its magnitude is fixed to the local constant $g_{0}=9.81~\mathrm{m/s^{2}}$, so only its $2$-DoF direction is estimated (a minimal perturbation on the sphere $g_{0}\mathbb{S}^{2}$), seeded at initialization and refined online rather than held fixed (Sec.~\ref{sec:initialization}).

The navigation state $\mathbf{x}_{n}^{f}$ at time step $n$, over which all sensor modalities are jointly optimized, is the manifold tuple
\begin{equation}
\mathbf{x}_{n}^{f} = \bigl(
{}_{i}^{w}\mathbf{R}_{n},\, {}_{i}^{w}\mathbf{p}_{n},\, {}_{i}^{w}\mathbf{v}_{n},\,
{}^{i}\mathbf{b}_{g_{n}},\, {}^{i}\mathbf{b}_{a_{n}},\,
\boldsymbol{\dot{\delta t}}_{n}
\bigr),
\end{equation}
which lives on $\mathrm{SO}(3)\times\mathbb{R}^{16}$, where ${}_{i}^{w}\mathbf{R}_{n} \in \mathrm{SO}(3)$, ${}_{i}^{w}\mathbf{p}_{n} \in \mathbb{R}^{3}$,
and ${}_{i}^{w}\mathbf{v}_{n} \in \mathbb{R}^{3}$ denote the orientation, position, and velocity of the IMU frame with respect to the local-world frame. The vectors
${}^{i}\mathbf{b}_{g_{n}} \in \mathbb{R}^{3}$ and ${}^{i}\mathbf{b}_{a_{n}} \in \mathbb{R}^{3}$
denote the gyroscope and accelerometer biases, and $\boldsymbol{\dot{\delta t}}_{n} \in \mathbb{R}^{4}$ holds the per-constellation receiver clock-drift rates (GPS, BDS, Galileo, GLONASS). In the double-difference configuration, the receiver clock offset is excluded from the state: it is canceled exactly by the double-differenced pseudorange factor
(Sec.~\ref{sec:dd_psr}) and is therefore neither observable nor needed by the estimator; in the base-station-free single-receiver variant, where differencing no longer removes it, it is reinstated as a per-constellation state, raising the per-node state to $\mathrm{SO}(3)\times\mathbb{R}^{20}$ (Appendix~\ref{app:single_receiver}). The clock drift rate remains observable through the Doppler
factor (Sec.~\ref{sec:doppler}) in both cases. The temporal evolution of $\boldsymbol{\dot{\delta t}}_{n}$
is modeled as a random walk between consecutive keyframes (Sec.~\ref{sec:doppler}).

\section{Methodology}
\label{sec:methodology}

\subsection{Factor Graph Formulation}
\label{sec:factor_graph}
The optimal state estimate is obtained by solving a Maximum A Posteriori (MAP) problem. Assuming that all sensor measurements are independent and their associated measurement noise follows a zero-mean Gaussian distribution, the MAP inference is equivalent to minimizing the sum of squared Mahalanobis distances of the residuals. 
Inspired by the formulations in tightly-coupled multi-sensor fusion frameworks \cite{glio,ligo,dellaert2017factor}, the objective function of our unified factor graph optimizes the full state $\mathcal{X}$ by minimizing the weighted sum of cost functions from all active factors:
\begin{equation}
\begin{aligned}
\mathcal{X}^{*} = \arg\min_{\mathcal{X}} \Biggl( & \left\| \mathbf{r}_{p} + \mathbf{H}_{p}\,(\mathcal{X}\boxminus\hat{\mathcal{X}}_{0}) \right\|^{2} \\
& + \sum_{(i,j) \in \mathcal{E}_{\mathcal{B}}} \left( \left\| \mathbf{r}_{\mathcal{B},ij} \right\|_{\boldsymbol{\Sigma}_{\mathcal{B},ij}}^{2} + \left\| \mathbf{r}_{\dot{\delta t},ij} \right\|_{\boldsymbol{\Sigma}_{\dot{\delta t},ij}}^{2} \right) \\
& + \sum_{(i,j) \in \mathcal{E}_{\mathcal{L}}} \left\| \mathbf{r}_{\mathcal{L},ij} \right\|_{\boldsymbol{\Sigma}_{\mathcal{L},ij}}^{2} \\
& + \sum_{t \in \mathcal{T}} \left( \left\| \mathbf{r}_{\rho, t} \right\|_{\boldsymbol{\Sigma}_{\rho, t}}^{2} + \left\| \mathbf{r}_{D, t} \right\|_{\boldsymbol{\Sigma}_{D, t}}^{2} \right) \Biggr)
\end{aligned}
\label{eq:map_objective}
\end{equation}
where $\{ \mathbf{r}_{p}, \mathbf{H}_{p} \}$ encapsulates the linearized prior obtained from the marginalization of past states (Sec.~\ref{sec:marginalization}), $\mathcal{X}\boxminus\hat{\mathcal{X}}_{0}$ denotes the minimal perturbation of the current state with respect to the linearization point $\hat{\mathcal{X}}_{0}$ fixed at the time of marginalization, and $\mathbf{H}_{p}$ is the square-root information matrix of the prior (the Cholesky factor of its information matrix), so the bracketed term is the whitened prior residual. The sets $\mathcal{E}_{\mathcal{B}}$ and $\mathcal{E}_{\mathcal{L}}$ collect the IMU-pre-integration edges between consecutive frames and the scan-to-multiscan LiDAR edges between the current scan and overlapping historical keyframes within the sliding window, respectively. The set $\mathcal{T}$ collects the GNSS epochs. The residuals $\mathbf{r}_{\mathcal{B},ij}, \mathbf{r}_{\mathcal{L},ij}$ denote the IMU and LiDAR factors with edge-specific covariances $\boldsymbol{\Sigma}_{\mathcal{B},ij}, \boldsymbol{\Sigma}_{\mathcal{L},ij}$, while $\mathbf{r}_{\rho, t}$ and $\mathbf{r}_{D, t}$ stack, at epoch $t$, the DD pseudorange residuals and the raw (single-receiver, single-satellite) Doppler residuals with covariances $\boldsymbol{\Sigma}_{\rho, t}$ and $\boldsymbol{\Sigma}_{D, t}$ detailed in Sec.~\ref{sec:dd_psr} and Sec.~\ref{sec:doppler}, respectively. The term $\mathbf{r}_{\dot{\delta t},ij}$ is the receiver clock-drift random-walk factor on the same edges \eqref{eq:clock_rw}, which regularizes constellation channels left momentarily unobserved by the Doppler factor.

\subsection{Double-Difference Pseudorange Factor}
\label{sec:dd_psr}

\begin{figure}[!tb]
  \centering
  \includegraphics[width=\columnwidth]{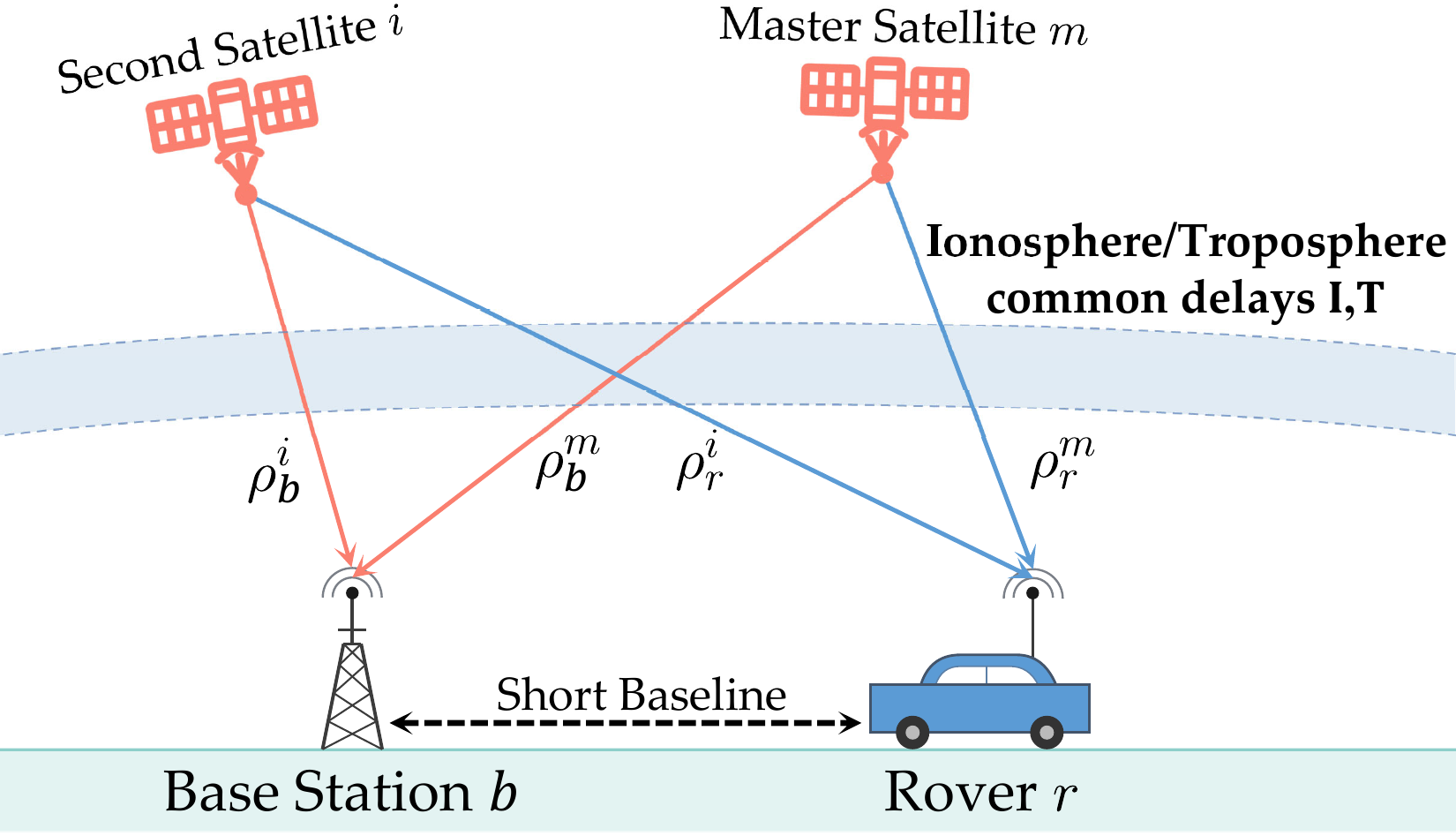}
  \caption{Double-differenced pseudorange geometry among the rover $r$, base station $b$, master satellite $m$, and secondary satellite $i$. Double differencing cancels the error terms common to both receivers and both satellites, namely the clock offsets and atmospheric delays.}
  \label{fig:dd_psr}
\end{figure}

GNSS pseudorange observations are processed via the double-difference (DD) paradigm. The raw pseudorange from the rover $r$ to satellite $s$ models the receiver--satellite geometric range corrupted by clock and propagation errors~\cite{enge1994global},
\begin{equation}
\rho_{r}^{s} = \bigl\|{}^{E}\mathbf{p}^{s}-{}^{E}\mathbf{p}_{r}\bigr\|
+ c\bigl(\delta t_{r}-\delta t^{s}\bigr)
+ I_{r}^{s} + T_{r}^{s} + \Omega_{r}^{s} + \epsilon_{r}^{s},
\label{eq:raw_pseudorange}
\end{equation}
where $\|{}^{E}\mathbf{p}^{s}-{}^{E}\mathbf{p}_{r}\|$ is the geometric range between the satellite position ${}^{E}\mathbf{p}^{s}$, obtained from the broadcast ephemeris, and the rover antenna position ${}^{E}\mathbf{p}_{r}$ given by \eqref{eq:antenna_pos}; $c$ is the speed of light, $\delta t_{r}$ and $\delta t^{s}$ are the receiver and satellite clock biases, $I_{r}^{s}$ and $T_{r}^{s}$ the ionospheric and tropospheric delays, $\Omega_{r}^{s}$ the Sagnac correction, and $\epsilon_{r}^{s}$ the residual measurement noise, dominated by multipath and NLOS errors in urban canyons; the base station $b$ follows the same model. The double difference therefore cancels these errors in two steps: differencing two satellites $(m,i)$ at one receiver removes that receiver's clock bias; differencing this between the rover $r$ and the base $b$ then removes the satellite clock biases and, over the short rover--base baseline, the ionospheric, tropospheric, and Sagnac terms (Fig.~\ref{fig:dd_psr}), leaving only the geometric ranges plus noise. The base-station ECEF coordinates ${}^{E}\mathbf{p}_{b}$ are taken from the RTCM correction stream of a known benchmark, and the receiver clock bias is therefore not carried as a state; only its drift rate $\boldsymbol{\dot{\delta t}}_{r}$, required by the Doppler factor (Sec.~\ref{sec:doppler}), is retained.

The rover antenna position is obtained by propagating the navigation state through the
extrinsic transformation:
\begin{equation}
{}^{E}\mathbf{p}_{r} = {}_{w}^{E}\mathbf{R}\bigl({}_{i}^{w}\mathbf{R}
\cdot {}^{i}\mathbf{l}_{r} + {}_{i}^{w}\mathbf{p}\bigr) + {}_{w}^{E}\mathbf{p}
\label{eq:antenna_pos}
\end{equation}

For a single satellite pair $(m,i)$, the scalar DD pseudorange residual is:
\begin{equation}
\begin{aligned}
r_{\rho}^{mi}(\mathcal{X}) = & \Bigl[ \bigl(\|{}^{E}\mathbf{p}^{m} - {}^{E}\mathbf{p}_{r}\| - \|{}^{E}\mathbf{p}^{i} - {}^{E}\mathbf{p}_{r}\|\bigr) \\
& \quad - \bigl(\|{}^{E}\mathbf{p}^{m} - {}^{E}\mathbf{p}_{b}\| - \|{}^{E}\mathbf{p}^{i} - {}^{E}\mathbf{p}_{b}\|\bigr) \Bigr] - \hat{\rho}_{rb}^{mi}
\end{aligned}
\label{eq:dd_pseudorange}
\end{equation}
where $\hat{\rho}_{rb}^{mi} = (\rho_{r}^{m} - \rho_{r}^{i}) - (\rho_{b}^{m} - \rho_{b}^{i})$
is the observed DD pseudorange. At epoch $t$, the $K=|\mathcal{S}_{rb,t}|-1$ scalar
residuals over all secondary satellites are stacked into the vector
$\mathbf{r}_{\rho,t} = [r_{\rho}^{m i_{1}}, \ldots, r_{\rho}^{m i_{K}}]^{T} \in \mathbb{R}^{K}$
that enters \eqref{eq:map_objective}.

Each receiver--satellite pseudorange is weighted by the elevation- and $\mathrm{SNR}$-dependent goGPS variance~\cite{gogps_weighting},
\begin{equation}
\sigma^{2}(\mathrm{SNR},E_{s}) = \sigma_{0}^{2}\,\frac{a_{0}+a_{1}\,10^{-\mathrm{SNR}/10}}{\sin^{2}\!E_{s}},
\label{eq:gogps_weight}
\end{equation}
where $\sigma_{0}^{2}$ is the nominal pseudorange noise variance carrying the physical units and $a_{0},a_{1}$ are dimensionless shape coefficients. At each epoch, the master satellite $m$ is the highest-elevation one jointly tracked by both receivers (least affected by multipath and NLOS) within the common set $\mathcal{S}_{rb,t}$ of satellites above a $15^{\circ}$ elevation mask, giving $K = |\mathcal{S}_{rb,t}|-1$ double differences. Because all $K$ of them are formed against this same master $m$, their covariance is not diagonal but carries a master-shared off-diagonal term~\cite{rtklib},
\begin{equation}
\boldsymbol{\Sigma}_{\rho,t} =
\begin{bmatrix}
\sigma_{\rho,m}^{2} + \sigma_{\rho,1}^{2} & \sigma_{\rho,m}^{2} & \cdots & \sigma_{\rho,m}^{2} \\
\sigma_{\rho,m}^{2} & \sigma_{\rho,m}^{2} + \sigma_{\rho,2}^{2} & \cdots & \sigma_{\rho,m}^{2} \\
\vdots & \vdots & \ddots & \vdots \\
\sigma_{\rho,m}^{2} & \sigma_{\rho,m}^{2} & \cdots & \sigma_{\rho,m}^{2} + \sigma_{\rho,K}^{2}
\end{bmatrix}
\label{eq:dd_cov}
\end{equation}
Here $\sigma_{\rho,s}^{2}=\sigma_{r,s}^{2}+\sigma_{b,s}^{2}$ stacks the rover and base code variances of satellite $s$, each from \eqref{eq:gogps_weight}. \revwording{Ignoring the shared-reference correlation misrepresents the joint information of the double differences; we therefore whiten $\mathbf{r}_{\rho,t}$ using the full covariance $\boldsymbol{\Sigma}_{\rho,t}$ before it enters \eqref{eq:map_objective}.}

The unit line-of-sight vector $\mathbf{e}_{r}^{s}$ and the Jacobians of $r_{\rho}^{mi}$ with respect to the navigation state and the global extrinsic are given in Appendix~\ref{app:gnss_jacobians}; the extrinsic-translation Jacobian makes ${}_{w}^{E}\mathbf{p}$ observable through the same line-of-sight geometry as the IMU position.

\subsection{Doppler Factor}
\label{sec:doppler}
DD pseudorange constrains the absolute position but is insensitive to the receiver clock drift and carries little velocity information. The Doppler factor supplies an instantaneous velocity constraint and is incorporated in raw, single-receiver--single-satellite form rather than double-differenced: the broadcast satellite clock drift $\dot{\delta t}^{s}$ is accurate enough to apply as a feed-forward correction, whereas double-differencing would cancel the absolute receiver clock drift $\boldsymbol{\dot{\delta t}}_{r}\in\mathbb{R}^{4}$ and leave it unobservable.

The Doppler shift $D_{r}^{s}$ measures the range rate between satellite $s$ and the receiver projected onto the line of sight,
\begin{equation}
\lambda D_{r}^{s} = -(\mathbf{e}_{r}^{s})^{T}\bigl({}^{E}\mathbf{v}^{s} - {}^{E}\mathbf{v}_{r}\bigr) - c\bigl(\dot{\delta t}_{r}^{\gamma(s)} - \dot{\delta t}^{s}\bigr) + n_{D}^{s},
\label{eq:doppler_model}
\end{equation}
with the unit line-of-sight vector $\mathbf{e}_{r}^{s}$ from the receiver to satellite $s$
\begin{equation}
\mathbf{e}_{r}^{s} = \frac{{}^{E}\mathbf{p}^{s} - {}^{E}\mathbf{p}_{r}}{\bigl\|{}^{E}\mathbf{p}^{s} - {}^{E}\mathbf{p}_{r}\bigr\|},
\label{eq:los}
\end{equation}
where $\lambda$ is the carrier wavelength, ${}^{E}\mathbf{v}^{s}$ the satellite velocity from the broadcast ephemeris, $\dot{\delta t}^{s}$ the broadcast satellite clock drift, and $n_{D}^{s}$ the noise. The index $\gamma(s)\in\{\mathrm{GPS},\mathrm{BDS},\mathrm{GAL},\mathrm{GLO}\}$ is the constellation of satellite $s$, and $\dot{\delta t}_{r}^{\gamma(s)}=[\boldsymbol{\dot{\delta t}}_{r}]_{\gamma(s)}$ selects the corresponding channel of the four-DoF receiver clock-drift state.

The rover antenna velocity in ECEF is obtained by differentiating
\eqref{eq:antenna_pos}, treating the local-world and ECEF frames as Earth-fixed so that
${}_{w}^{E}\mathbf{R}$ has zero time derivative:
\begin{equation}
{}^{E}\mathbf{v}_{r} = {}_{w}^{E}\mathbf{R}\bigl({}_{i}^{w}\mathbf{v} + {}_{i}^{w}\mathbf{R}\bigl\lfloor {}^{i}\boldsymbol{\omega} \bigr\rfloor_{\times} {}^{i}\mathbf{l}_{r}\bigr),
\label{eq:antenna_vel}
\end{equation}
where ${}^{i}\boldsymbol{\omega} = \boldsymbol{\omega}_{m} - {}^{i}\mathbf{b}_{g}$ is the
bias-corrected gyroscope reading.

Applying the satellite clock drift as a feed-forward correction, the raw Doppler residual is
\begin{equation}
r_{D}^{s}(\mathcal{X}) = -(\mathbf{e}_{r}^{s})^{T}\bigl({}^{E}\mathbf{v}^{s} - {}^{E}\mathbf{v}_{r}\bigr) - c\bigl(\dot{\delta t}_{r}^{\gamma(s)} - \dot{\delta t}^{s}\bigr) - \lambda D_{r}^{s}.
\label{eq:doppler_residual}
\end{equation}
One scalar residual is formed per tracked satellite and routed to its constellation's clock-drift channel; the per-satellite residuals are stacked at epoch $t$ into $\mathbf{r}_{D,t}$ in \eqref{eq:map_objective}, with diagonal covariance $\boldsymbol{\Sigma}_{D,t}=\mathrm{diag}(\sigma_{D,s}^{2})$ from the same goGPS model \eqref{eq:gogps_weight} scaled to Doppler noise units.

\paragraph*{Clock-drift random walk}
The receiver clock drift is modeled as a random walk between consecutive keyframes; for each edge $(i,j)\in\mathcal{E}_{\mathcal{B}}$ a four-channel factor
\begin{equation}
\mathbf{r}_{\dot{\delta t},ij} = \boldsymbol{\dot{\delta t}}_{r,j} - \boldsymbol{\dot{\delta t}}_{r,i},
\qquad
\boldsymbol{\Sigma}_{\dot{\delta t},ij}=\sigma_{\dot{\delta t}}^{2}\,\Delta t_{ij}\,\mathbf{I}_{4},
\label{eq:clock_rw}
\end{equation}
keeps the clock-drift block full rank whenever a constellation is momentarily unobserved by the Doppler factor.

The Jacobians of $r_{D}^{s}$ are given in Appendix~\ref{app:gnss_jacobians}. The factor carries the constellation-aware mapping $\gamma(s)$ and, through the lever-arm term, couples to the body orientation ${}_{i}^{w}\mathbf{R}$ and the gyroscope bias ${}^{i}\mathbf{b}_{g}$, so the four-channel clock-drift state is consistently observed.

\subsection{IMU Pre-Integration Factor}
\label{sec:imu_factor}
IMU measurements between consecutive frames $i$ and $j$ (edge
$(i,j)\in\mathcal{E}_{\mathcal{B}}$) are summarized into a single relative-motion
constraint by on-manifold pre-integration~\cite{forster2017preintegration,vins-mono}, with the bias Jacobians propagated so that the increments are corrected on a bias update without re-integration. The resulting $15$-dimensional residual $\mathbf{r}_{\mathcal{B},ij}$
stacks the pre-integrated rotation, velocity, and position errors and the
gyroscope and accelerometer bias random-walk increments, weighted by the propagated
covariance $\boldsymbol{\Sigma}_{\mathcal{B},ij}$, and enters \eqref{eq:map_objective}
as $\|\mathbf{r}_{\mathcal{B},ij}\|_{\boldsymbol{\Sigma}_{\mathcal{B},ij}}^{2}$. As the only high-rate factor in the window, it constrains the states between LiDAR and GNSS updates and supplies the inertial information that supports poorly observed LiDAR directions.

\subsection{LiDAR Factor via Scan-to-Multiscan Registration}
\label{sec:lidar_factor}

The information matrix $\mathbf{H}^{\mathcal{L}}$ of the relative scan-to-multiscan factor is strictly singular along the global pose common mode, the $6$-DoF gauge freedom long known in bundle adjustment~\cite{triggs1999bundle} and pose-graph SLAM~\cite{grisetti2010tutorial,kummerle2011g2o,olson2006fast,rosen2019sesync}, conventionally removed by anchoring one pose. \revwording{Here, gauge freedom means that applying the same rigid-body rotation and translation to all active poses leaves the relative LiDAR residuals unchanged. This classical invariance means that increasing LiDAR weights adds no direct information along this common rigid-motion mode.} \revwording{Fig.~\ref{fig:hessian_blocks} illustrates this separation of relative LiDAR information and the GNSS global reference (Sec.~\ref{sec:lidar_fusion}).}

We adopt the right (body) perturbation $\mathbf{T}\boxplus\boldsymbol{\xi}=\mathbf{T}\exp(\boldsymbol{\xi}^{\wedge})$, where $\boldsymbol{\xi}=[\boldsymbol{\phi}^{T},\boldsymbol{\rho}^{T}]^{T}\in\mathbb{R}^{6}$ collects a rotational increment $\boldsymbol{\phi}$ (ordered first) and a translational increment $\boldsymbol{\rho}$, and $(\cdot)^{\wedge}$ is the hat map onto $\mathfrak{se}(3)$ (Tab.~\ref{tab:nomenclature}). The analysis rests on a single identity of the $\mathrm{SE}(3)$ adjoint $\mathrm{Ad}_{\mathbf{T}}=\left[\begin{smallmatrix}\mathbf{R}&\mathbf{0}\\\lfloor\mathbf{p}\rfloor_{\times}\mathbf{R}&\mathbf{R}\end{smallmatrix}\right]\in\mathbb{R}^{6\times6}$, which transfers a right (body) perturbation to its equivalent left (world) form~\cite{sola2018micro}:
\begin{equation}
\mathbf{T}\exp(\boldsymbol{\xi}^{\wedge})=\exp\!\bigl((\mathrm{Ad}_{\mathbf{T}}\boldsymbol{\xi})^{\wedge}\bigr)\,\mathbf{T}.
\label{eq:adjoint}
\end{equation}
The adjoint is invertible, with $\mathrm{Ad}_{\mathbf{T}}^{-1}=\mathrm{Ad}_{\mathbf{T}^{-1}}$, a property used repeatedly below.

\subsubsection{Factor Definition and Linearization}
Let $i$ be the current scan and $j$ an overlapping keyframe. For voxel correspondence $k$, let $\mathbf{p}_k$ be the current-scan voxel mean (in frame $i$) and $\mathbf{q}_k$ the matched keyframe voxel mean (in frame $j$), with voxel covariances $\mathbf{C}_k^{p},\mathbf{C}_k^{q}$. With the relative pose $\mathbf{T}_{ij}\triangleq\mathbf{T}_i^{-1}\mathbf{T}_j=(\mathbf{R}_{ij},\mathbf{t}_{ij})\in\mathrm{SE}(3)$, transforming the keyframe mean into frame $i$ gives the residual
\begin{equation}
\mathbf{r}_k=\mathbf{T}_{ij}\,\mathbf{q}_k-\mathbf{p}_k,
\label{eq:vgicp_residual}
\end{equation}
whose distribution-to-distribution VGICP~\cite{vgicp} information fuses the two voxel covariances,
\begin{equation}
\boldsymbol{\Omega}_k=\bigl(\mathbf{C}_k^{p}+\mathbf{R}_{ij}\,\mathbf{C}_k^{q}\,\mathbf{R}_{ij}^{T}\bigr)^{-1}.
\label{eq:vgicp_info}
\end{equation}
Expressing both means in frame $i$ is mathematically equivalent to registering the current scan to the keyframe; it merely selects a reference frame in which the relative pose $\mathbf{T}_{ij}$ appears directly. The cost of one scan-to-keyframe factor is $e_{ij}=\sum_k \mathbf{r}_k^{T}\boldsymbol{\Omega}_k\mathbf{r}_k$, and the current scan forms $m$ such factors against its overlapping keyframes $\{j_1,\dots,j_m\}$ rather than registering to a single fused map; $e_{ij}$ is the per-edge expansion of the LiDAR term $\|\mathbf{r}_{\mathcal{L},ij}\|^{2}_{\boldsymbol{\Sigma}_{\mathcal{L},ij}}$ in \eqref{eq:map_objective}.

Linearizing at $\mathbf{T}_i\boxplus\boldsymbol{\xi}_i$ and $\mathbf{T}_j\boxplus\boldsymbol{\xi}_j$, the right perturbation propagates as $\mathbf{T}_{ij}\mapsto\exp(-\boldsymbol{\xi}_i^{\wedge})\mathbf{T}_{ij}\exp(\boldsymbol{\xi}_j^{\wedge})$; substituting into \eqref{eq:vgicp_residual} and keeping first-order terms,
\begin{equation}
\begin{aligned}
\mathbf{r}_k\approx{}&(\mathbf{T}_{ij}\mathbf{q}_k-\mathbf{p}_k)
+\underbrace{\bigl[\,\lfloor\mathbf{T}_{ij}\mathbf{q}_k\rfloor_{\times}\;\;{-}\mathbf{I}\,\bigr]}_{\mathbf{J}_{k,i}}\boldsymbol{\xi}_i\\
&+\underbrace{\bigl[\,{-}\mathbf{R}_{ij}\lfloor\mathbf{q}_k\rfloor_{\times}\;\;\mathbf{R}_{ij}\,\bigr]}_{\mathbf{J}_{k,j}}\boldsymbol{\xi}_j,
\end{aligned}
\label{eq:lidar_jac}
\end{equation}
where $\mathbf{J}_{k,i}=\partial\mathbf{r}_k/\partial\boldsymbol{\xi}_i$ and $\mathbf{J}_{k,j}=\partial\mathbf{r}_k/\partial\boldsymbol{\xi}_j$ are $3\times6$ (the first three columns act on $\boldsymbol{\phi}$, the last three on $\boldsymbol{\rho}$). Following the analytic GPU VGICP kernel of GLIM~\cite{glim}, the three $6\times6$ Gauss--Newton Hessian blocks and the gradient of this factor are
\begin{align}
\mathbf{H}_{ii}&=\textstyle\sum_k\mathbf{J}_{k,i}^{T}\boldsymbol{\Omega}_k\mathbf{J}_{k,i},&
\mathbf{H}_{ij}&=\textstyle\sum_k\mathbf{J}_{k,i}^{T}\boldsymbol{\Omega}_k\mathbf{J}_{k,j},\nonumber\\
\mathbf{H}_{jj}&=\textstyle\sum_k\mathbf{J}_{k,j}^{T}\boldsymbol{\Omega}_k\mathbf{J}_{k,j},\nonumber\\
\mathbf{b}_i&=\textstyle\sum_k\mathbf{J}_{k,i}^{T}\boldsymbol{\Omega}_k\mathbf{r}_k,&
\mathbf{b}_j&=\textstyle\sum_k\mathbf{J}_{k,j}^{T}\boldsymbol{\Omega}_k\mathbf{r}_k.
\label{eq:lidar_hessian_blocks}
\end{align}
The fused information $\boldsymbol{\Omega}_k$ is treated as constant within each Gauss--Newton step (its dependence on $\mathbf{T}_{ij}$ is not differentiated) and refreshed between iterations at the updated relative pose.

\subsubsection{Gauge Null Space and Rank of the LiDAR Hessian}
\label{sec:hessian_contribution}
The residual $\mathbf{r}_k$ depends on the two endpoint poses only through $\mathbf{T}_{ij}=\mathbf{T}_i^{-1}\mathbf{T}_j$. A common global left transform $\mathbf{T}_i\mapsto\mathbf{G}\mathbf{T}_i,\ \mathbf{T}_j\mapsto\mathbf{G}\mathbf{T}_j$ ($\mathbf{G}\in\mathrm{SE}(3)$) leaves $\mathbf{T}_{ij}$, and hence $\mathbf{r}_k$, unchanged. Infinitesimally, by~\eqref{eq:adjoint}, a global left perturbation $\boldsymbol{\eta}$ corresponds, in each body frame, to $\boldsymbol{\xi}_i=\mathrm{Ad}_{\mathbf{T}_i^{-1}}\boldsymbol{\eta}$ and $\boldsymbol{\xi}_j=\mathrm{Ad}_{\mathbf{T}_j^{-1}}\boldsymbol{\eta}$. Invariance forces $\mathbf{J}_{k,i}\boldsymbol{\xi}_i+\mathbf{J}_{k,j}\boldsymbol{\xi}_j=\mathbf{0}$ for this family; since it holds for all $\boldsymbol{\eta}$, we obtain the pointwise gauge identity
\begin{equation}
\mathbf{J}_{k,i}\,\mathrm{Ad}_{\mathbf{T}_i^{-1}}+\mathbf{J}_{k,j}\,\mathrm{Ad}_{\mathbf{T}_j^{-1}}=\mathbf{0}
\;\Longleftrightarrow\;
\mathbf{J}_{k,j}=-\mathbf{J}_{k,i}\,\mathrm{Ad}_{\mathbf{T}_{ij}},
\label{eq:gauge_identity}
\end{equation}
which is verified directly from the explicit Jacobians in \eqref{eq:lidar_jac}. Substituting \eqref{eq:gauge_identity} gives $\mathbf{H}_{ij}=-\mathbf{H}_{ii}\mathrm{Ad}_{\mathbf{T}_{ij}}$ and $\mathbf{H}_{jj}=\mathrm{Ad}_{\mathbf{T}_{ij}}^{T}\mathbf{H}_{ii}\mathrm{Ad}_{\mathbf{T}_{ij}}$, so a single factor contributes the symmetric $12\times12$ block
\begin{equation}
\mathbf{H}_{\mathrm{edge}}^{ij}=\begin{bmatrix}\mathbf{H}_{ii}&\mathbf{H}_{ij}\\[2pt]\mathbf{H}_{ij}^{T}&\mathbf{H}_{jj}\end{bmatrix}
=\begin{bmatrix}\mathbf{I}\\[2pt]-\mathrm{Ad}_{\mathbf{T}_{ij}}^{T}\end{bmatrix}\mathbf{H}_{ii}\begin{bmatrix}\mathbf{I}&-\mathrm{Ad}_{\mathbf{T}_{ij}}\end{bmatrix},
\label{eq:hedge_factorization}
\end{equation}
where ``edge'' denotes the factor-graph edge joining nodes $i$ and $j$. Because $\mathrm{Ad}_{\mathbf{T}_{ij}}$ is invertible and, for a geometrically non-degenerate factor, $\mathbf{H}_{ii}\succ\mathbf{0}$,
\begin{equation}
\begin{aligned}
\operatorname{rank}\mathbf{H}_{\mathrm{edge}}^{ij}&=\operatorname{rank}\mathbf{H}_{ii}\le6,\\
\ker\mathbf{H}_{\mathrm{edge}}^{ij}&=\bigl\{[(\mathrm{Ad}_{\mathbf{T}_{ij}}\mathbf{w})^{T},\mathbf{w}^{T}]^{T}\mid\mathbf{w}\in\mathbb{R}^{6}\bigr\},
\end{aligned}
\label{eq:edge_rank}
\end{equation}
i.e., moving both poses along a common rigid motion $(\boldsymbol{\xi}_i,\boldsymbol{\xi}_j)=(\mathrm{Ad}_{\mathbf{T}_{ij}}\mathbf{w},\mathbf{w})$ changes neither residual nor cost. The rank identity in \eqref{eq:edge_rank} holds for any $\mathbf{H}_{ii}$; the kernel identity requires $\mathbf{H}_{ii}\succ\mathbf{0}$: a rank-deficient $\mathbf{H}_{ii}$ enlarges the kernel by its own null directions, the geometric deficiency distinguished below.

\begin{figure}[t]
\centering
\includegraphics[width=\columnwidth]{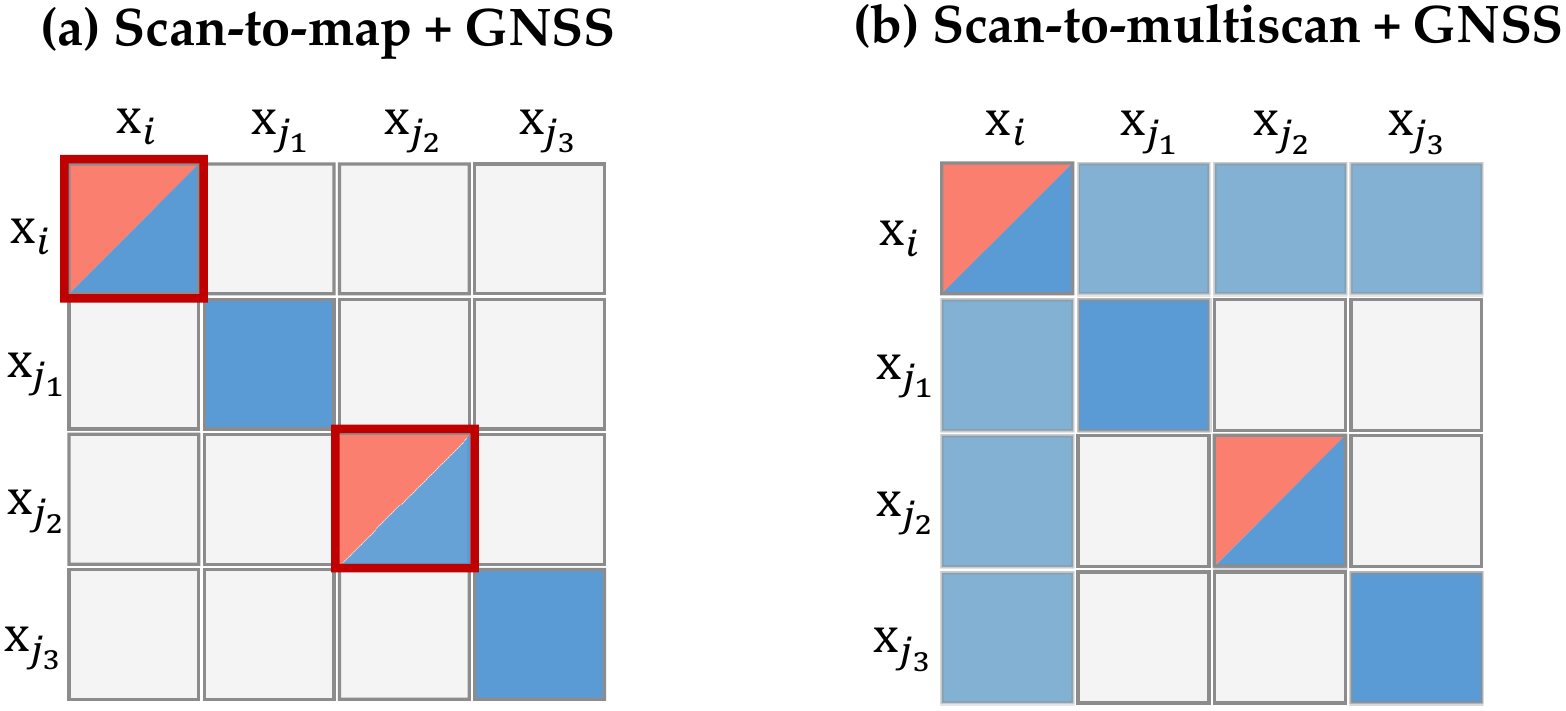}
\caption{Block structure of the window information matrix (GNSS in red, LiDAR in blue), contrasting the two LiDAR factors; nodes $\mathbf{x}_i,\mathbf{x}_{j_1},\mathbf{x}_{j_2},\mathbf{x}_{j_3}$ are the current scan and three keyframes, and each cell is a $6\times6$ pose subblock. \textbf{(a)}~Scan-to-map enters LiDAR as a unary absolute constraint on the diagonal blocks (red-boxed), acting on the same global degrees of freedom as GNSS. \textbf{(b)}~Scan-to-multiscan enters LiDAR as relative constraints coupling $\mathbf{x}_i$ to each keyframe (off-diagonal blocks), so $\mathcal{G}\subseteq\ker(\mathbf{H}^{\mathcal{L}})$.}
\label{fig:hessian_blocks}
\end{figure}

Consider now a window of $N$ poses. The relative factors form a connected pose graph, the pose substructure of the full GLIO2 factor graph, which also carries IMU, GNSS, and extrinsic factors. Assembling the per-edge blocks \eqref{eq:hedge_factorization} by node index yields the aggregate LiDAR information matrix $\mathbf{H}^{\mathcal{L}}\in\mathbb{R}^{6N\times6N}$ (symmetric positive semidefinite). Define the $6$-dimensional global-gauge subspace
\begin{equation}
\mathcal{G}=\bigl\{\bigl[(\mathrm{Ad}_{\mathbf{T}_1^{-1}}\boldsymbol{\eta})^{T},\dots,(\mathrm{Ad}_{\mathbf{T}_N^{-1}}\boldsymbol{\eta})^{T}\bigr]^{T}\!\mid\!\boldsymbol{\eta}\in\mathbb{R}^{6}\bigr\}.
\label{eq:gauge_subspace}
\end{equation}
For any edge $(i,j)$, applying $[\mathbf{I},-\mathrm{Ad}_{\mathbf{T}_{ij}}]$ to $(\mathrm{Ad}_{\mathbf{T}_i^{-1}}\boldsymbol{\eta},\mathrm{Ad}_{\mathbf{T}_j^{-1}}\boldsymbol{\eta})$ yields $\mathrm{Ad}_{\mathbf{T}_i^{-1}}\boldsymbol{\eta}-\mathrm{Ad}_{\mathbf{T}_{ij}}\mathrm{Ad}_{\mathbf{T}_j^{-1}}\boldsymbol{\eta}=\mathbf{0}$, so every edge contributes zero on $\mathcal{G}$ and $\mathcal{G}\subseteq\ker\mathbf{H}^{\mathcal{L}}$, hence $\operatorname{nullity}\ge6$. Conversely, if the pose graph is connected and each $\mathbf{H}_{ii}\succ\mathbf{0}$ (geometric non-degeneracy), then $\mathbf{H}^{\mathcal{L}}\boldsymbol{v}=\mathbf{0}$ iff every edge satisfies $\boldsymbol{v}_i=\mathrm{Ad}_{\mathbf{T}_{ij}}\boldsymbol{v}_j$; on a connected graph this recursion propagates uniquely from any root, and its solution set is exactly the $6$-dimensional orbit $\mathcal{G}$. Therefore
\begin{equation}
\operatorname{rank}\mathbf{H}^{\mathcal{L}}=6N-6,
\label{eq:lidar_rank}
\end{equation}
a constant structural deficiency of $6$: the entire local trajectory can undergo any global rigid motion without changing any relative registration residual (Fig.~\ref{fig:hessian_blocks}(b)).

Two different rank deficiencies arise here and play opposite roles. (i)~The structural gauge deficiency \eqref{eq:lidar_rank}, the classical gauge/datum freedom of relative pose-graph and bundle-adjustment problems~\cite{triggs1999bundle,grisetti2010tutorial}, is always $6$-dimensional ($\mathcal{G}$), independent of scene geometry, and arises from the invariance of relative factors to the global common mode. (ii)~Geometric degeneracy: each $\mathbf{H}_{ii}=\sum_k\mathbf{J}_{k,i}^{T}\boldsymbol{\Omega}_k\mathbf{J}_{k,i}$ is a sum of rank$\le3$ terms whose rank reflects the diversity of the correspondence directions $\{\mathbf{T}_{ij}\mathbf{q}_k\}$ weighted by $\{\boldsymbol{\Omega}_k\}$; since VGICP weights along the voxel surface normals, a full-rank $\mathbf{H}_{ii}$ requires sufficient surface-normal diversity, which corridors, tunnels, or aerial surveys over flat terrain fail to provide, weakening $\mathbf{H}_{ii}$ along the poorly observed directions. Deficiency~(i), lying in $\mathcal{G}$, is permanent and is the mechanism exploited here; (ii), eroding the relative directions within $\mathcal{G}^{\perp}$ that LiDAR should otherwise observe, is scene-dependent and is what IMU/GNSS must compensate.

\subsubsection{Implications for Tightly-Coupled Fusion}
\label{sec:lidar_fusion}
Decompose the tangent space as $\mathbb{R}^{6N}=\mathcal{G}\oplus\mathcal{G}^{\perp}$, the global gauge plus the relative shape subspace. The inclusion $\mathcal{G}\subseteq\ker(\mathbf{H}^{\mathcal{L}})$, equivalently $\mathbf{H}^{\mathcal{L}}\mathbf{P}_{\mathcal{G}}=\mathbf{0}$ with $\mathbf{P}_{\mathcal{G}}$ the projector onto $\mathcal{G}$, follows from the gauge identity \eqref{eq:gauge_identity} unconditionally; it holds even when individual $\mathbf{H}_{ii}$ are rank-deficient under degeneracy. The converse, that the nullity is exactly six \eqref{eq:lidar_rank}, is what additionally requires a connected pose graph with full-rank $\mathbf{H}_{ii}$. The two consequences below need only the unconditional inclusion, so $\mathbf{H}^{\mathcal{L}}$ vanishes on $\mathcal{G}$ whatever the scene geometry.

\revwording{\textbullet~\textbf{Global-frame consistency.} The assembled LiDAR information $\mathbf{H}^{\mathcal{L}}$ vanishes on $\mathcal{G}$, so relative LiDAR factors introduce no direct prior on the common rigid motion of the window and leave the global datum to GNSS. LiDAR weights still influence the estimated relative geometry, so biased or excessively weighted correspondences can affect the fused trajectory.}

\textbullet~\textbf{Degeneracy robustness.} Geometric degeneracy weakens only directions within $\mathcal{G}^{\perp}$, never the datum $\mathcal{G}$, where LiDAR contributes nothing. The datum is supplied instead by the other modalities through the shared extrinsic ${}_{w}^{E}\mathbf{T}$: the DD-pseudorange factors fix the global position (Sec.~\ref{sec:dd_psr}), gravity fixes roll and pitch, and the global heading, unobservable from a single epoch, is made observable by GNSS-constrained motion through the Doppler factor (Sec.~\ref{sec:doppler}) and the yaw initialization (Sec.~\ref{sec:initialization}).

The total information $\mathbf{H}=\mathbf{H}^{\mathcal{L}}+\mathbf{H}^{\mathcal{I}}+\mathbf{H}^{\mathcal{G}}+\mathbf{H}^{\pi}$ \revwording{combines complementary sensor constraints, with its conditioning depending on measurement geometry and relative weighting. A fixed-map S2M pose prior instead places an absolute constraint} on $\mathbf{H}_{ii}$ whose information falls in $\mathcal{G}$ and is weighed directly against GNSS (Fig.~\ref{fig:hessian_blocks}(a)).

\subsubsection{GPU-Parallel Linearization}
\label{sec:gpu_acceleration}

Unlike scan-to-map, which refines only the current scan against a fixed target, scan-to-multiscan keeps every active keyframe as an optimization variable, so each nonlinear iteration changes every relative pose $\mathbf{T}_{ij}$ and must re-associate and relinearize all active LiDAR factors. \revwording{We adopt the GPU-parallel scan-matching design of GLIM~\cite{glim} and integrate it with raw GNSS factors and the shared local-world-to-ECEF state for joint optimization.} The CPU smoother emits the current linearization point; the GPU then, in parallel over all points, performs the $O(1)$ spatial-hash data association against each keyframe voxel map and accumulates the per-factor $6\times6$ block Hessians and gradient $\{\mathbf{H}_{ii},\mathbf{H}_{ij},\mathbf{H}_{jj},\mathbf{b}_i,\mathbf{b}_j\}$ of~\eqref{eq:lidar_hessian_blocks}. Following GLIM, all factor data are serialized so that each linearization pass incurs only two CPU--GPU transfers regardless of the number of active factors, \revwording{reducing synchronization overhead.} The sparse factor-graph solve stays on the CPU and feeds the refined poses back as the next linearization point until convergence, re-associating and relinearizing all LiDAR factors on the GPU at each iteration. \revwording{This offloads dense correspondence search and per-point linearization to the GPU while leaving the sparse graph solve on the CPU. After this offloading, the optimization update is the largest measured runtime stage on the Orin~NX (Tab.~\ref{tab:runtime_breakdown}).} The structure of the active keyframe set these factors operate on, and how it is maintained, are described next.

\subsection{Active Keyframe Set}
\label{sec:active_map}
\revwording{GLIO2 does not maintain a fused, fixed world-frame reference map. Its active keyframe set $\mathcal{K}$ contains the keyframes currently kept for matching, a subset of the active states in the fixed-lag window. Each keyframe stores a multi-resolution VGICP voxel map of its raw cloud in its own local frame~\cite{glim} and its pose $\mathbf{T}_j\in\mathrm{SE}(3)$. Scan-to-keyframe matching supplies the relative constraints (Sec.~\ref{sec:lidar_factor}); the global point cloud is formed by transforming and combining the keyframe clouds using their current poses and ${}_{w}^{E}\mathbf{T}$.} A new deskewed scan $P_i$ is voxelized and matched to the keyframes by the overlap ratio $o(P_i,P_j)$~\cite{glim}: those with sufficient overlap form $\mathcal{K}_i$ and supply the scan-to-keyframe factors of Sec.~\ref{sec:lidar_factor}, while non-overlapping ones are skipped.

\textbf{Live refinement.} Because the keyframe poses are optimization states rather than fixed reference frames, each iSAM2~\cite{isam2} update relinearizes them jointly, and the GPU re-associates the cross-scan correspondences at the updated poses (Sec.~\ref{sec:gpu_acceleration}). The per-keyframe voxel maps stay fixed in their own frames, but as the poses move, the world-frame geometry they induce is re-estimated. The active keyframe set thus co-evolves with the trajectory, unlike the frozen reference a scan-to-map front-end commits to.

\textbf{Maintenance.} After optimization, $P_i$ becomes a keyframe only if it is novel ($o<\tau_{\mathrm{ins}}$). A keyframe is pruned from the matching set when its overlap with the current scan falls below $\tau_{\mathrm{prune}}$, or, once the set exceeds its cap $N_{\max}$, as the most redundant member (low overlap with the current scan but high mutual overlap with the rest). Pruning only stops new factors from forming against that keyframe; its state and the factors already in the graph stay active until the state is marginalized out of the window (Sec.~\ref{sec:marginalization}). Algorithm~\ref{alg:map} summarizes one cycle.

\begin{algorithm}[t]
\footnotesize
\DontPrintSemicolon
\LinesNumbered
\SetCommentSty{algcommfont}
\caption{Active keyframe set maintenance.}
\label{alg:map}
\KwIn{deskewed scan $P_i$; active keyframe set $\mathcal{K}$; sliding window $W$}
\KwOut{updated active keyframe set $\mathcal{K}$}
\tcc{Associate and build LiDAR factors}
voxelize $P_i$ into a multi-resolution VGICP map\;
$\mathcal{K}_i \leftarrow \{\, j\in\mathcal{K} : o(P_i,P_j)>\tau_{\mathrm{assoc}} \,\}$\tcp*{select overlapping keyframes}
construct the scan-to-keyframe factors $P_i\!\leftrightarrow\!\mathcal{K}_i$ (Sec.~\ref{sec:lidar_factor})\;
\tcc{Refine the window jointly}
\Repeat{convergence}{
  re-associate and relinearize all LiDAR factors on the GPU (Sec.~\ref{sec:gpu_acceleration})\;
  one iSAM2 update, propagating the refined poses as the next linearization point\;
}
write the converged poses back to $\mathcal{K}$\;
\tcc{Maintain the keyframe set}
\If{$o\bigl(P_i,\textstyle\bigcup_{j\in\mathcal{K}}P_j\bigr)<\tau_{\mathrm{ins}}$}{
  $\mathcal{K}\leftarrow\mathcal{K}\cup\{i\}$\tcp*{insert as a new keyframe}
}
\ForEach{$j\in\mathcal{K}$ \textnormal{with} $o(P_i,P_j)<\tau_{\mathrm{prune}}$}{
  remove $j$ from $\mathcal{K}$\tcp*{retain its existing factors}
}
\If{$|\mathcal{K}|>N_{\max}$}{
  remove the most redundant keyframe\;
}
marginalize the states leaving $W$ into $\{\mathbf{r}_{p},\mathbf{H}_{p}\}$\tcp*{Sec.~\ref{sec:marginalization}}
\end{algorithm}

\subsection{Marginalization}
\label{sec:marginalization}
Fixed-lag operation wraps iSAM2~\cite{isam2} in an incremental fixed-lag smoother: navigation states whose timestamp leaves the window $W$ are Schur-complemented out of their cliques, folding their accumulated constraints into the linearized Gaussian prior $\{\mathbf{r}_{p},\mathbf{H}_{p}\}$ of \eqref{eq:map_objective}, frozen at its linearization point $\hat{\mathcal{X}}_{0}$. The window-shared extrinsic ${}_{w}^{E}\mathbf{T}$ and gravity ${}^{w}\mathbf{g}$ are exempt: their information from evicted states is folded into the prior, yet the variables themselves stay active and keep being relinearized, holding the local-world frame anchored to ECEF across GNSS dropouts.

\subsection{Initialization}
\label{sec:initialization}
GLIO2 bootstraps online in three coarse-to-fine steps, with no dedicated excitation maneuver
or external heading source. \textbf{(i) Leveling.} The accelerometer specific force
averaged over a short stationary interval gives the gravity direction and thus
the initial roll and pitch; this only seeds the $2$-DoF gravity state, which is afterward
refined in the sliding window (Sec.~\ref{sec:system_state}) rather than held fixed, so a
residual leveling error is corrected online instead of corrupting the accelerometer bias.
\textbf{(ii) Global anchor.} An approximate ECEF origin ${}_{w}^{E}\mathbf{p}$ is solved
first from a code GNSS position fix~\cite{rtklib}: a double-differenced (DD) code solution
relative to the base station when one is available, or a single-point-positioning (SPP)
solution otherwise. The four-channel receiver clock-drift state is seeded from the
Doppler observations. \textbf{(iii) Heading.} The
estimator then runs in LIO-only mode for a short ($5$\,s) interval, during which the LiDAR
registration and IMU jointly refine the coarse roll, pitch, and position of the local
trajectory, while the global heading (yaw), unobservable to LIO, remains free. Aligning this trajectory
to the concurrent GNSS positions by a yaw-only least-squares fit then recovers the global
heading and refines the horizontal position of the anchor. The bootstrap is thus coarse-to-fine:
leveling, anchor, and heading are deliberately approximate, and their residuals are absorbed
online by the continued refinement of the gravity state and the shared extrinsic
${}_{w}^{E}\mathbf{T}$ in the sliding window rather than held fixed.

\subsection{Global Batch Optimization}
\label{sec:batch}
\begin{figure}[t]
    \centering
    \includegraphics[width=\columnwidth]{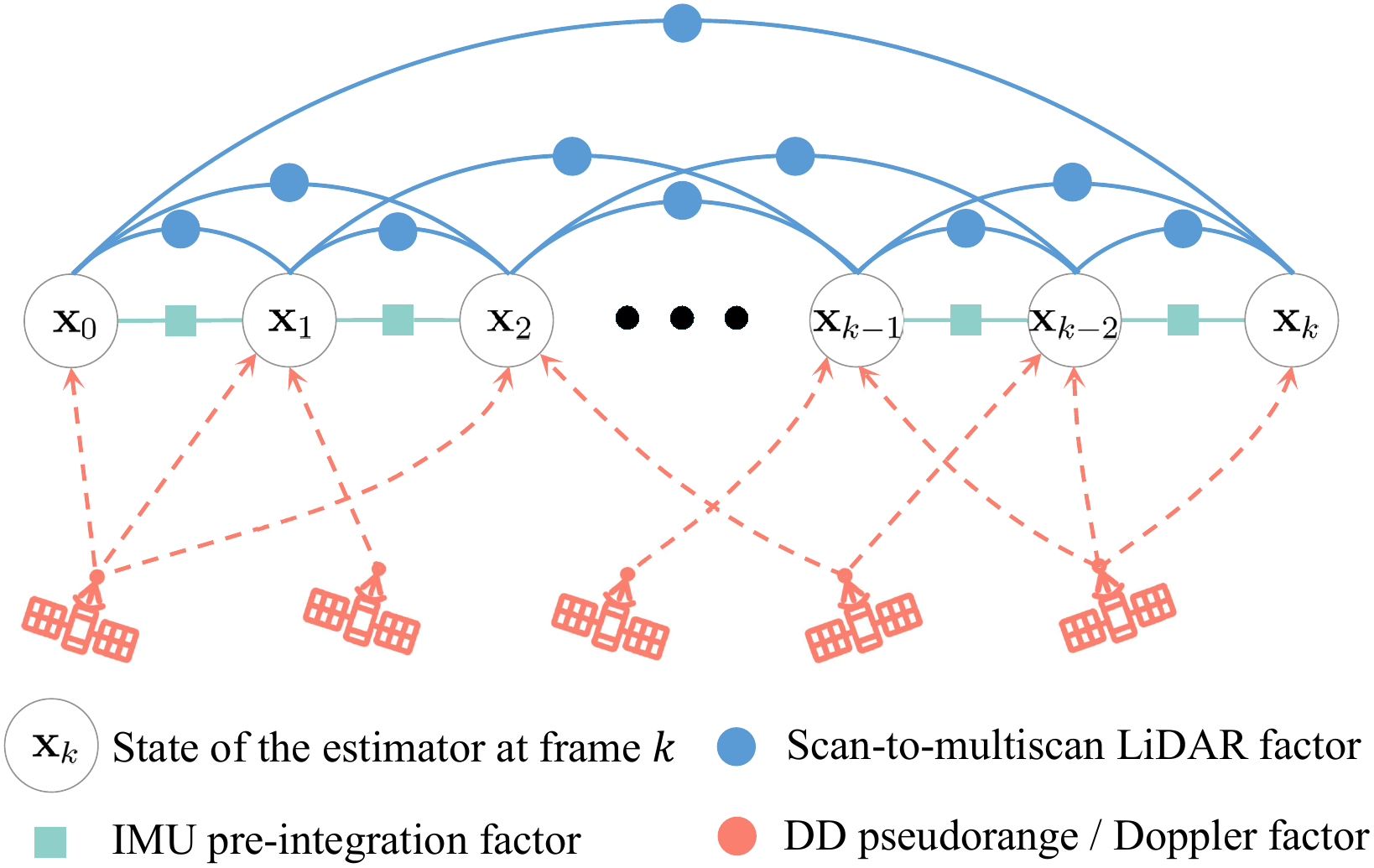}
    \caption{Factor graph of the global batch optimization (Sec.~\ref{sec:batch}): the whole trajectory is re-optimized jointly over all states $\mathbf{x}_0,\dots,\mathbf{x}_k$ under IMU pre-integration factors, the scan-to-multiscan LiDAR between-factors, and the raw double-differenced pseudorange and Doppler factors.}
    \label{fig:batch_graph}
\end{figure}

In most conditions the online front-end is already \revwording{locally accurate and globally referenced} (Sec.~\ref{sec:lidar_fusion}). \revwording{An important source of residual error is a global frame left unobserved for longer than the fixed-lag window:} during prolonged GNSS denial or NLOS, or when the platform stays stationary longer than the window spans. Once such an interval outlasts the window, the smoother marginalizes the affected states (Sec.~\ref{sec:marginalization}) and freezes their linearization into the prior $\{\mathbf{r}_{p},\mathbf{H}_{p}\}$ before GNSS has constrained them, leaving a global bias that no later window can revisit. GLIO2 therefore adds an offline batch optimization over the entire trajectory, jointly over every frame rather than a keyframe subsample (Fig.~\ref{fig:batch_graph}). It keeps the form of~\eqref{eq:map_objective} with three changes: the IMU, scan-to-multiscan, and GNSS factor sets span the full session; the marginalization prior is replaced by a weak prior on the first state and on ${}_{w}^{E}\mathbf{T}$ to fix the residual gauge freedom; and the DD-pseudorange residuals pass through a round-based robust kernel (FDD) that rejects grossly inconsistent satellites and down-weights the rest by their residual, tightening the inlier set over successive rounds to suppress multipath and NLOS faults without discarding the observations that anchor the datum. The problem is solved by Levenberg--Marquardt.

The batch is inexpensive because it reuses the factors the front-end already built. Since the front-end runs a large fixed-lag window and repeatedly relinearizes its dense correspondences online (Sec.~\ref{sec:active_map},~\ref{sec:gpu_acceleration}), the relative scan-to-multiscan factors it emits \revwording{have already been refined within the online window when each state is marginalized.} The back-end keeps, for every frame, only these marginalized relative-pose factors, each carrying the information matrix \revwording{obtained from online registration}, together with the DD-pseudorange and Doppler factors; the raw clouds are discarded. The per-point VGICP search and linearization, the dominant cost of the dense formulation, is thus never repeated, and each LM iteration relinearizes only the compact LiDAR, IMU, and GNSS residuals. \revwording{Because a common rigid correction preserves relative poses and scan correspondences (Sec.~\ref{sec:lidar_fusion}), it does not require LiDAR re-registration. When batch refinement also changes the relative geometry, the compact factors approximate the original dense registration constraints, allowing the trajectory to be refined without repeating the per-point search and linearization; Fig.~\ref{fig:backend_bubble} compares the accuracy and runtime of this back-end with GLIO.}

\section{Experiments}
\label{sec:experiments}

\begin{table*}[!t]
\centering
\caption{Overview of the public and self-collected datasets: platform, sensor suite, reference (ground-truth) source, and the evaluated sequences with their dominant degradation scenario.}
\label{tab:datasets}
\setlength{\tabcolsep}{5pt}
\renewcommand{\arraystretch}{1.2}
\resizebox{\textwidth}{!}{%
\begin{tabular}{llllll}
\toprule
Dataset (carrier) & LiDAR & IMU & GNSS receiver & Reference & Sequences (scenario) \\
\midrule
UrbanNav~\cite{hsu2021urbannav} & Velodyne HDL-32E, 10\,Hz & Xsens MTi-10, 400\,Hz & u-blox ZED-F9P, 1\,Hz & NovAtel SPAN-CPT & Hong Kong Tst, Whampoa (deep-canyon NLOS) \\
MARS-LVIG~\cite{MARS-LVIG} & Livox Avia, 10\,Hz & BMI088, 200\,Hz & u-blox ZED-F9P, 10\,Hz & DJI RTK-fixed & HKisland1, HKairport1 (aerial, LiDAR degeneracy) \\
M3DGR~\cite{zhang2025towards}   & Livox Mid-360, 10\,Hz & ICM40609, 200\,Hz & CUAV C-RTK 9Ps, 10\,Hz & Fixed RTK + ArUco & Gra1, Out1, Den1 (induced degradation, GNSS-denied) \\
\midrule
\textbf{Ours} (UAV) & Livox Mid-360, 10\,Hz & ICM40609, 200\,Hz & u-blox ZED-F9P, 10\,Hz & u-blox ZED-F9P RTK-fixed & Switch (indoor/outdoor), Plain (flat, LiDAR degeneracy) \\
\multirow{2}{*}{\textbf{Ours} (Vehicle)} & Velodyne VLP-32E, 10\,Hz & \multirow{2}{*}{Xsens MTi-30, 400\,Hz} & \multirow{2}{*}{u-blox ZED-F9P, 1\,Hz} & \multirow{2}{*}{NovAtel SPAN-CPT} & \multirow{2}{*}{Urban (urban canyon), Bridge (LiDAR degeneracy)} \\
 & Hesai XT-32, 10\,Hz & & & & \\
\bottomrule
\end{tabular}}
\end{table*}

We evaluate GLIO2 on both public datasets and our self-collected
dataset. The public datasets (UrbanNav~\cite{hsu2021urbannav},
MARS-LVIG~\cite{MARS-LVIG}, and M3DGR~\cite{zhang2025towards})
benchmark GLIO2 against state-of-the-art LiDAR--Inertial--GNSS systems
(Sec.~\ref{sec:public_benchmark}). The self-collected dataset targets the two failure
modes this work addresses, raw GNSS denied or NLOS and LiDAR geometric
degeneracy, across four sequences (\textit{Urban}, \textit{Switch}, \textit{Bridge},
and \textit{Plain}) (Sec.~\ref{sec:private_benchmark}), and drives the ablation study
that isolates the contribution of each module (Sec.~\ref{sec:ablation}). Finally, we
report two real-world deployments of GLIO2, an autonomous facade-cleaning UAV and
a guide-dog navigation system for the visually impaired (Sec.~\ref{sec:application}).

\subsection{Experimental Setup and Datasets}

\subsubsection{Public Datasets}
We assess cross-platform generalization on three public datasets with synchronized GNSS, LiDAR, and inertial streams (Tab.~\ref{tab:datasets}).

\textbullet~\textbf{UrbanNav}~\cite{hsu2021urbannav} is collected by a ground vehicle in the dense urban canyons of Hong Kong, where high-rise buildings induce severe multipath and frequent NLOS reception; we use its \textit{Tst} and \textit{Whampoa} sequences.

\textbullet~\textbf{MARS-LVIG}~\cite{MARS-LVIG} is an aerial-UAV benchmark collected in Hong Kong with a downward-looking LiDAR at $80$--$130$\,m altitude, whose texture-poor coastline and runway scenes drive the scan registration toward degeneracy; we use its raw-GNSS \textit{HKisland1} and \textit{HKairport1} sequences against the onboard RTK-fixed reference.

\textbullet~\textbf{M3DGR}~\cite{zhang2025towards} is a ground-robot benchmark of systematically induced degradation (LiDAR-drift, wheel-slippage, GNSS-denied scenes). In its \textit{Den1} sequence the robot leaves an open-sky area, traverses a GNSS-denied region, and returns, with ground truth from a geodetic RTK receiver outdoors and from the relative transform between ArUco-measured start/end poses for the denied loop; we additionally evaluate its open-sky \textit{Gra1} and \textit{Out1} sequences. As M3DGR provides no reference-station stream, GLIO2 runs in its single-receiver variant (without the base-station double difference; Appendix~\ref{app:single_receiver}), a direct test of base-station-free generalization.

\subsubsection{Sensor Setup}
\begin{figure}[tb]
    \centering
    \includegraphics[width=\columnwidth]{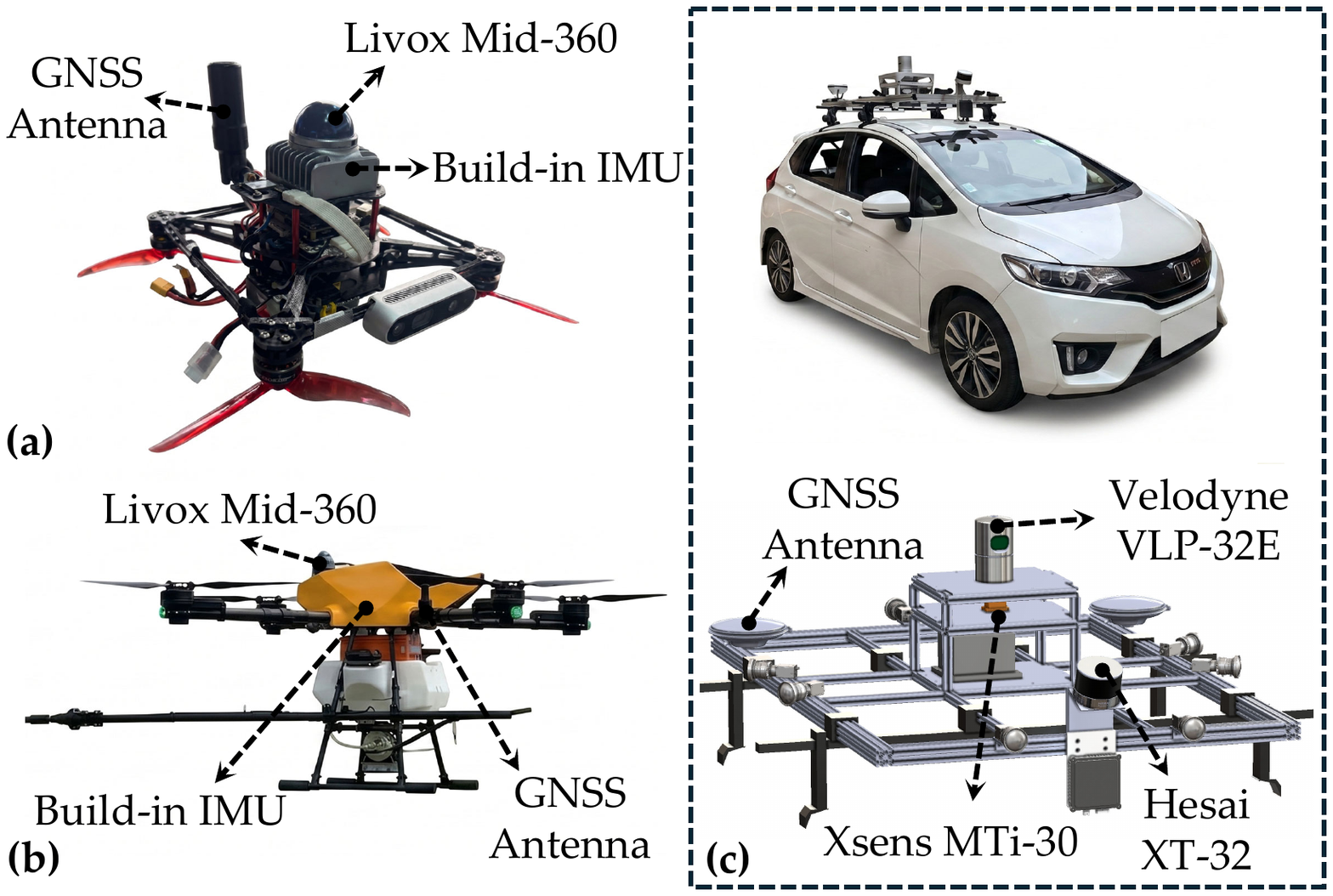}
    \caption{Self-collected platforms: \textbf{(a)},~\textbf{(b)}~the two UAVs; \textbf{(c)}~the ground vehicle (top) and its roof-mounted sensor suite (bottom). Per-platform sensor models and reference sources are listed in Tab.~\ref{tab:datasets}.}
    \label{fig:platforms}
\end{figure}

Our data are collected on aerial and ground platforms (Fig.~\ref{fig:platforms};
sensor suites and reference sources in Tab.~\ref{tab:datasets}). The \textit{Switch}
and \textit{Plain} sequences are flown on two quadrotors that share an identical sensor
suite (a $300$\,mm-class drone for \textit{Switch} and the larger $823$\,mm
facade-cleaning UAV for \textit{Plain}), each running
GLIO2 onboard in real time on an NVIDIA Jetson Orin~NX ($16$\,GB). The ground platform
is a passenger car whose recordings, targeting large-scale collection rather than
onboard deployment, are processed offline for the long-range \textit{Urban} and
\textit{Bridge} sequences.

LiDAR--IMU extrinsics are calibrated per platform: the Livox Mid-360 integrates the
ICM40609 IMU, for which we adopt the factory extrinsics, while the separate
Velodyne and Xsens units are calibrated with the robust real-time LiDAR--inertial
initialization of Zhu \textit{et al.}~\cite{zhu2022robust}. The
IMU-to-GNSS-antenna lever arm ${}^{i}\mathbf{l}_{r}$ is taken from the mechanical
assembly drawings of each platform. All sensor streams are time-synchronized
through the Precision Time Protocol (PTP, IEEE~1588) to a common clock, and all
double-differenced pseudorange factors and the RTK reference obtain
their base-station observations over the network in RTCM format from a regional
continuously operating reference station (CORS) service, removing the need for a
locally deployed base station.

\subsubsection{Self-Collected Datasets}
\label{sec:private_datasets}
Our four self-collected sequences isolate complementary failure modes (Tab.~\ref{tab:datasets}). \textit{Urban} ($1.44$\,km, vehicle) transitions from dense Hong Kong canyons into an open area, stressing long-range global consistency under persistent multipath. \textit{Switch} (UAV) starts outdoors with RTK-fixed GNSS, enters a building that fully blocks the pseudorange channel, and exits again, stressing indoor--outdoor handover. \textit{Bridge} ($5.66$\,km in $239.8$\,s, vehicle at up to $96$\,km/h) crosses a long, straight deck where the sparse along-track structure and the heavy moving traffic cause LiDAR degeneracy along the driving direction, most severely over a $2{,}160$\,m main span, while GNSS reception remains open-sky. \textit{Plain} (low-altitude UAV) flies over flat, feature-poor ground whose near-planar scan constrains only the vertical direction and leaves the horizontal pose under-determined.

\begin{figure}[!tb]
  \centering
  \includegraphics[width=\linewidth]{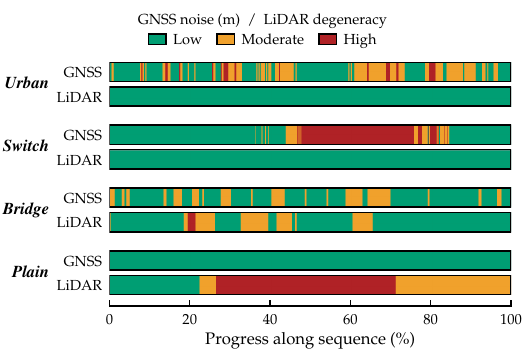}
  \caption{Per-sequence GNSS noise and LiDAR registration degeneracy along each timeline, colored green/yellow/red for low/moderate/high severity. The two modalities degrade in complementary sequences: GNSS in \textit{Urban} and \textit{Switch}, LiDAR on \textit{Bridge} and \textit{Plain}.}
  \label{fig:gnss_noise_levels}
\end{figure}

\label{sec:gnss_noise_char}
To make each sequence's difficulty explicit, we read the GNSS noise directly from the
receiver. For every satellite $s$ the u-blox ZED-F9P publishes a one-sigma pseudorange
noise $\sigma_{s,t}$ (``psr\_std''), which, following LIGO~\cite{ligo}, we adopt as the
primary noise descriptor. At each epoch $t$ we summarize the visible satellite set
$\mathcal{S}_t$ by its mean
\begin{equation}
  \bar{\sigma}_t = \frac{1}{|\mathcal{S}_t|}\sum_{s\in\mathcal{S}_t}\sigma_{s,t},
\end{equation}
and color the top strip of each pair in Fig.~\ref{fig:gnss_noise_levels} green,
yellow, or red as $\bar{\sigma}_t$ falls in $(0,2]$, $(2,5]$, or $>\!5$~m.
\textit{Plain} is open-sky for almost every epoch (green), so its difficulty is
geometric rather than radiometric; \textit{Urban} carries persistent moderate multipath;
\textit{Switch} is clean outdoors but saturates during the indoor passage;
\textit{Bridge} is clean on average with a thin noisy tail. Note that ``psr\_std''
serves only as a sequence-difficulty descriptor and captures measurement precision
alone: the residual NLOS/multipath bias is not exposed by the receiver variance.

We characterize LiDAR degeneracy from the same online scan-to-multiscan registration,
through the conditioning of its distribution-to-distribution VGICP information
matrix $\mathbf{H}_{ii}$~\eqref{eq:lidar_hessian_blocks} restricted to the translational subspace; the
rotational block is inflated by the lever-arm effect and would otherwise dominate the
raw $6$-DoF condition number~\cite{zhang2016degeneracy,tuna2024xicp}. Because the
platforms carry different LiDARs, each sequence is graded on its own translational scale
rather than a shared one. \textit{Urban} and \textit{Switch} stay well-conditioned
throughout, where dense surrounding structure constrains every direction. \textit{Bridge}
is mostly well-constrained with only intermittent moderate spells, where the open,
featureless deck weakens the geometry along the driving direction without degenerating
for long. \textit{Plain} is the sustained degeneracy case, where the flat ground constrains
only the vertical direction and leaves most of the low-altitude flight strongly
ill-conditioned. The two
modalities thus degrade in complementary sequences (GNSS through \textit{Urban} and
\textit{Switch}, LiDAR on \textit{Bridge} and \textit{Plain}), the regime the joint estimator is designed for. This grading is an illustrative per-scan severity indicator, not a
calibrated test; the intervals it flags coincide with the spans where the LiDAR-only and
scan-to-map baselines diverge (Tab.~\ref{tab:comparison_results}), confirming that it
tracks real degradation.

\subsubsection{Evaluation Metrics}
\label{sec:datasets_protocol_ref}
We report the absolute trajectory error (ATE) as the root-mean-square error (RMSE)
in both the horizontal 2D (labeled ``Hor.'' in the tables) and the full 3D (labeled ``3D''), the latter exposing the GNSS-weak vertical
channel that the horizontal metric masks. For the over-confidence analysis (Sec.~\ref{sec:ablation}) we additionally report the estimator's own marginal pose uncertainty, the per-axis posterior standard deviation (``$1\sigma$'') recovered from the optimized information matrix, which reveals when a system is confident yet wrong. Two alignment conventions are used, following the evaluation
protocol of LIGO~\cite{ligo}. GNSS-aware systems recover a global ECEF state, which we map to a local
east--north--up (ENU) frame through the same anchor point in the ECEF frame and compare
without any trajectory alignment.
Pure LiDAR-inertial odometry (FAST-LIO2) lives in a local frame and is
aligned by a single $\mathrm{SE}(3)$ transform fitted to the first $33\%$ of the
trajectory. Loop closure is disabled for all methods. The indoor--outdoor
\textit{Switch} sequence lacks a full absolute reference, so we report ATE over its
outdoor RTK-fixed intervals and assess the indoor part qualitatively via map continuity.
For both UAV sequences (\textit{Switch}, \textit{Plain}) the reference trajectory is
the fixed RTK solution computed onboard by the u-blox ZED-F9P receiver from the RTCM
corrections of a regional CORS station received over the Internet, at the same
$10$\,Hz rate as the raw measurements, following the ground-truth protocol of
LIGO~\cite{ligo}.

All accuracy results are obtained on a desktop workstation with an Intel Core
Ultra~7 265KF CPU, $32$\,GB of memory, and an NVIDIA RTX~5060~Ti GPU; the
application results are obtained on the onboard NVIDIA Jetson Orin~NX
($16$\,GB) edge platform, and runtimes are reported on both platforms
(Tab.~\ref{tab:runtime_breakdown}).

\begin{table*}[!t]
  \centering
  \caption{Horizontal (2D) and 3D ATE RMSE (m) on the public sequences (UrbanNav, MARS-LVIG, M3DGR).}
  \label{tab:public_results}
  \small
  \setlength{\tabcolsep}{3pt}
  \renewcommand{\arraystretch}{1.1}
  \resizebox{\textwidth}{!}{%
  \begin{tabular}{l*{14}{c}}
    \toprule
    \multirow{2}{*}{Sequence}
    & \multicolumn{2}{c}{GLIO2(Online)}
    & \multicolumn{2}{c}{GLIO2(Batch)}
    & \multicolumn{2}{c}{GLIO(Online)~\cite{glio}}
    & \multicolumn{2}{c}{GLIO(Batch)}
    & \multicolumn{2}{c}{LIGO~\cite{ligo}}
    & \multicolumn{2}{c}{LIO-SAM-GPS~\cite{lio-sam}}
    & \multicolumn{2}{c}{RTKLIB~\cite{rtklib}} \\
    \cmidrule(r){2-3} \cmidrule(r){4-5} \cmidrule(r){6-7} \cmidrule(r){8-9} \cmidrule(r){10-11} \cmidrule(r){12-13} \cmidrule(r){14-15}
    & Hor.\ & 3D
    & Hor.\ & 3D
    & Hor.\ & 3D
    & Hor.\ & 3D
    & Hor.\ & 3D
    & Hor.\ & 3D
    & Hor.\ & 3D \\
    \midrule
    UrbanNav-Tst   & 2.153 & \underline{3.539} & \textbf{1.064} & \textbf{2.583} & 5.164 & 19.559 & \underline{1.516} & 3.748 & 5.935 & 8.051 & 8.061 & 18.930 & 6.965 & 11.801 \\
    UrbanNav-Whampoa & 2.834 & 7.756 & \textbf{1.955} & \textbf{2.987} & 7.358 & 16.627 & \underline{2.070} & \underline{5.967} & 8.506 & 16.800 & 10.460 & 22.710 & 21.230 & 41.762 \\
    MARS-LVIG-HKairport1 & 2.545 & \underline{3.207} & \textbf{0.899} &  \textbf{1.701} & $\times$ & $\times$ & $\times$ & $\times$ & 7.002 & 10.472 & 42.980 & 45.355 & \underline{2.194} & 9.672 \\
    MARS-LVIG-HKisland1 &\underline{2.260} & \underline{3.877} & \textbf{2.134} & \textbf{2.473} & $\times$ & $\times$ & $\times$ & $\times$ & 6.999 & 9.317 & 13.111 & 15.530 & 2.772 & 13.038 \\
    M3DGR-Gra1      & \underline{0.527} & \underline{0.534} & \textbf{0.318} & \textbf{0.346} & -- & -- & -- & -- & 0.690 & 1.334 & 4.967 & 9.912 & 7.974 & 12.945 \\
    M3DGR-Out1        & \underline{0.182} & \underline{0.301} & \textbf{0.181} &  \textbf{0.226} & -- & -- & -- & -- & 0.219 & 0.331 & 26.716 & 26.963 & 2.882 & 4.655 \\
    M3DGR-Den1        & \underline{0.762} & \underline{2.785} & \textbf{0.591} & \textbf{0.814} & -- & -- & -- & -- & 11.740 & 11.749 & 8.352 & 8.914 & 2.551 & 17.270 \\
    \bottomrule
  \end{tabular}
  }
  \vspace{2pt}
  \begin{flushleft}
    \footnotesize
    Bold: best result; underline: second best. $\times$: total failure. ``--'': not applicable: M3DGR provides no reference-station observations, so the double-difference GLIO baselines cannot run on its sequences, while GLIO2 falls back to its single-receiver variant there.
  \end{flushleft}
\end{table*}

\begin{table*}[!t]
  \centering
  \caption{Horizontal (2D) and 3D ATE RMSE (m) on the self-collected sequences.}
  \label{tab:comparison_results}
  \small
  \setlength{\tabcolsep}{3pt}
  \renewcommand{\arraystretch}{1.1}
  \resizebox{\textwidth}{!}{%
  \begin{tabular}{l*{16}{c}}
    \toprule
    \multirow{2}{*}{Sequence}
    & \multicolumn{2}{c}{GLIO2(Online)}
    & \multicolumn{2}{c}{GLIO2(Batch)}
    & \multicolumn{2}{c}{GLIO(Online)~\cite{glio}}
    & \multicolumn{2}{c}{GLIO(Batch)}
    & \multicolumn{2}{c}{LIGO~\cite{ligo}}
    & \multicolumn{2}{c}{LIO-SAM-GPS~\cite{lio-sam}}
    & \multicolumn{2}{c}{RTKLIB~\cite{rtklib}}
    & \multicolumn{2}{c}{FAST-LIO2~\cite{fastlio2}} \\
    \cmidrule(r){2-3} \cmidrule(r){4-5} \cmidrule(r){6-7} \cmidrule(r){8-9} \cmidrule(r){10-11} \cmidrule(r){12-13} \cmidrule(r){14-15} \cmidrule(r){16-17}
    & Hor.\ & 3D
    & Hor.\ & 3D
    & Hor.\ & 3D
    & Hor.\ & 3D
    & Hor.\ & 3D
    & Hor.\ & 3D
    & Hor.\ & 3D
    & Hor.\ & 3D \\
    \midrule
    Urban & \underline{1.127} & \underline{1.978} & \textbf{0.919} & \textbf{1.327} & 3.529 & 12.377 & 1.589 & 4.492 & 3.663 & 12.935 & 10.468 & 13.212 & 15.945 & 38.877 & 6.035 & 14.604 \\
    Switch & \underline{0.425} & 0.636 & \textbf{0.213} &\textbf{0.301} & 2.727 & 3.303 & 0.654 & 0.881 & 0.734 & 0.865 & 1.334 & 1.442 & 3.890 & 11.212 & 0.564 & \underline{0.580} \\
    Bridge   & \underline{1.625} & \underline{3.586} & \textbf{1.438} & \textbf{2.552} & $\times$ & $\times$ & $\times$ & $\times$ & $\times$ & $\times$ & $\times$ & $\times$ & 5.634 & 11.367 & $\times$ & $\times$ \\
    Plain    & 0.596 & 0.663 & \textbf{0.142}& \textbf{0.162}& $\times$ & $\times$ & $\times$ & $\times$ & 9.881 & 9.882 & 17.580 & 18.378 & \underline{0.150} & \underline{0.289} & $\times$ & $\times$ \\
    \bottomrule
  \end{tabular}
  }
  \vspace{2pt}
  \begin{flushleft}
    \footnotesize
    Bold: best result; underline: second best. $\times$: total failure.
  \end{flushleft}
\end{table*}

\subsection{Experimental Results}
\revwording{GLIO2 successfully processes all eleven public and self-collected sequences spanning GNSS NLOS and weak LiDAR geometry, and its batch stage further refines the global trajectory.}

\subsubsection{Public Dataset Benchmark Experiments}
\label{sec:public_benchmark}
\revwording{For each baseline, we use the results reported in its original publication for sequences evaluated in that publication. For sequences not covered by the original publication, we run the original implementation using its original parameter settings. The baseline accuracy results therefore comprise both published results and results reproduced in this study.}
We benchmark GLIO2 as an online LiDAR--Inertial--GNSS system against representative methods on the public datasets (Tab.~\ref{tab:public_results}). GLIO2
fuses raw double-differenced pseudorange and Doppler factors when a base station is
available, and falls back to ordinary single-receiver pseudorange and Doppler factors
on datasets without reference-station observations (e.g.\ M3DGR), exposing the
generalization of the same formulation. The external baselines are
LIGO~\cite{ligo}, GLIO~\cite{glio}, the GNSS-factor configuration of
LIO-SAM~\cite{lio-sam} (``LIO-SAM-GPS''), and \revwording{RTKLIB~\cite{rtklib}.}
\revwording{RTKLIB performs kinematic RTK positioning using double-differenced code and carrier-phase observations, forward-backward filtering, and fix-and-hold ambiguity resolution, with the same regional CORS reference station as GLIO2.} GLIO is a two-stage system, and we report both stages: GLIO(Online) is its
real-time scan-to-map front-end, while GLIO(Batch) is its offline back-end, which
re-optimizes the full graph to convergence and is not real-time (its cost is
analyzed in Fig.~\ref{fig:backend_bubble}).

GLIO2 attains the best total RMSE on the UrbanNav \textit{Tst} and \textit{Whampoa}
sequences (the batch variant on both, the online front-end second on \textit{Tst}), confirming that the tightly-coupled formulation transfers to third-party
urban data (Tab.~\ref{tab:public_results}). On the aerial MARS-LVIG sequences the sky is largely open, so the
kinematic-RTK RTKLIB is strong in the horizontal channel ($2.194$\,m on \textit{HKairport1},
$2.772$\,m on \textit{HKisland1}), yet its float carrier-phase solution wanders in the
vertical, inflating its total RMSE to $9.7$--$13.0$\,m. Through its accurate LiDAR
observations, GLIO2 anchors the vertical that RTK leaves to drift; among the
LiDAR--GNSS systems it attains the best total RMSE on both sequences, $3.207$ and
$3.877$\,m online, reduced to $1.701$ and $2.473$\,m by the batch optimization, versus
$10.5$ and $9.3$\,m for the hierarchical LIGO. The downward-looking scans over the
texture-poor runway and coastline keep the horizontal registration geometry weak
(Sec.~\ref{sec:hessian_contribution}), and the loosely-coupled LIO-SAM-GPS drifts to
$13$--$43$\,m horizontally. Finally, the M3DGR sequences provide no base-station
observations, so GLIO2 runs in its single-receiver variant there; it
attains sub-meter horizontal accuracy on all three, the larger \textit{Den1} 3D residual stemming from the vertical error that GNSS denial induces over the loop.

\begin{figure*}[!t]
  \centering
  \includegraphics[width=\textwidth]{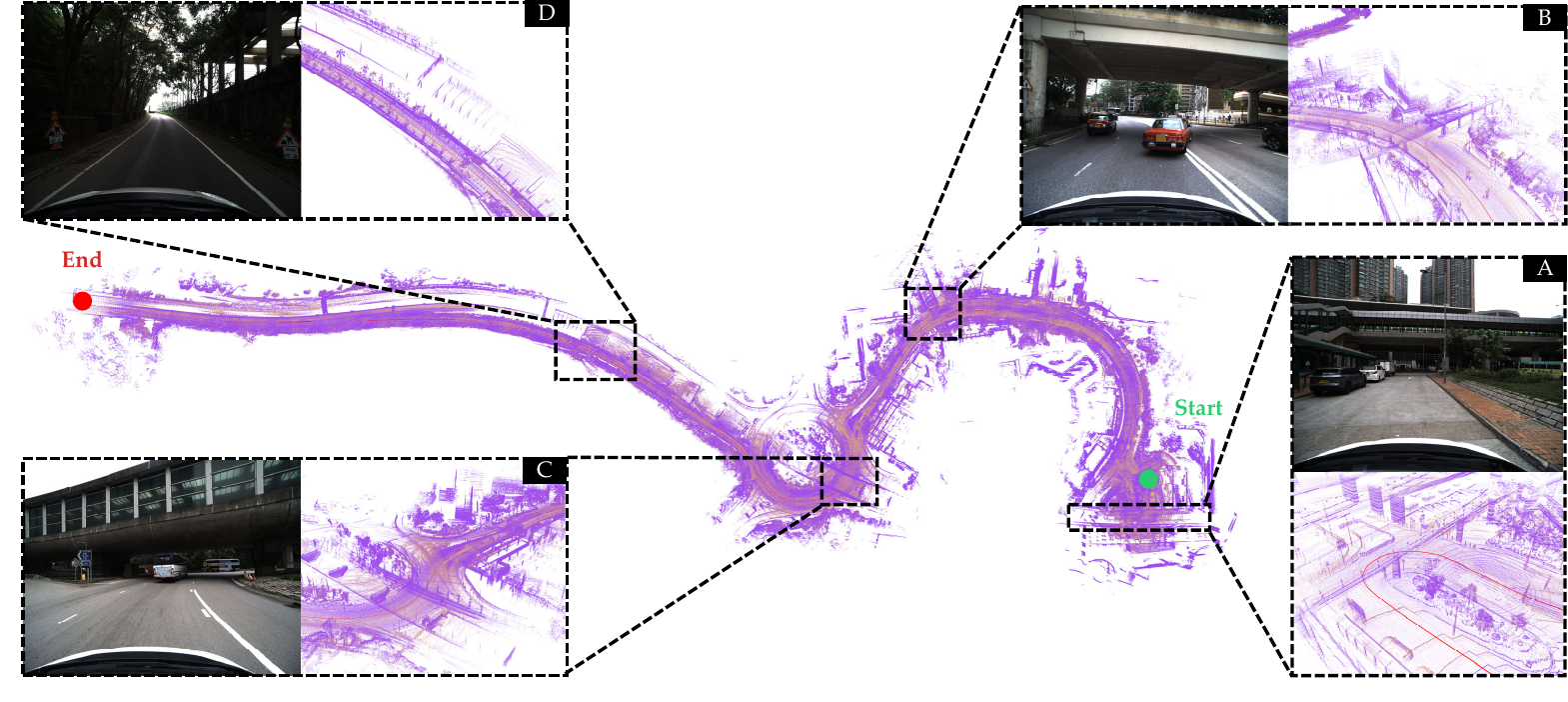}
  \caption{\textit{Urban} sequence: the GLIO2 point-cloud map built online over the full vehicle route (\textit{Start} to \textit{End}). The four callouts \textbf{A}--\textbf{D} pair each onboard camera view with the local map at a representative location of strong GNSS NLOS and occlusion. The global map stays consistent across the route despite the persistent multipath.}
  \label{fig:map_urban}
\end{figure*}

\begin{figure}[!t]
  \centering
  \includegraphics[width=\columnwidth]{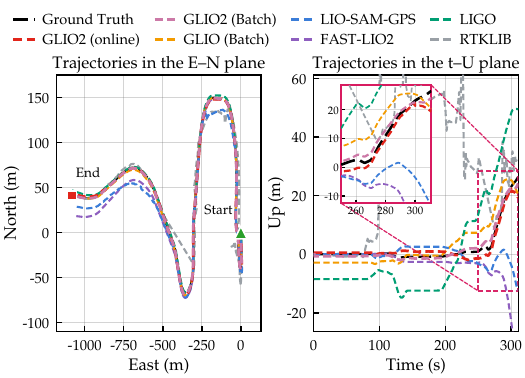}
  \caption{\textit{Urban} sequence: estimated trajectory (top-down and vertical) against the SPAN-CPT reference, with the inset magnifying the vertical drift over the final GNSS-degraded span ($250$--$310$\,s).}
  \label{fig:traj_urban}
\end{figure}

Two failure patterns in Tab.~\ref{tab:public_results} merit closer analysis. First, GLIO (in both configurations) fails on both
MARS-LVIG sequences: the downward-looking scan over the texture-poor terrain is
degenerate from the first frames, so GLIO's scan-to-map front-end
diverges under this start-up degeneracy, and because the corrupted
registration is produced upstream of the back-end, offline re-iteration cannot
repair it.
Second, GLIO's real-time front-end degrades to
$5.2$--$7.4$\,m horizontal across the UrbanNav sequences, while its offline
back-end recovers to $1.5$--$2.1$\,m. The comparison thus
separates an architectural failure that offline computation cannot repair
(front-end degeneracy, on MARS-LVIG) from a front-end accuracy gap on UrbanNav
that the back-end does recover; GLIO2's own
batch optimization (Sec.~\ref{sec:batch}) exploits the latter, re-iterating the same
GPU-linearized scan-to-multiscan and raw-GNSS factors offline to recover accuracy
without inheriting a degeneracy-fragile front-end.

\subsubsection{Self-Collected Dataset Benchmark Experiments}
\label{sec:private_benchmark}
We now evaluate the four self-collected sequences (Sec.~\ref{sec:private_datasets},
Tab.~\ref{tab:comparison_results}), reporting accuracy and the mechanism behind it.

\paragraph{\textit{Urban}}
On the $320$\,s urban-canyon route, GLIO2 is the most accurate fusion system: its
real-time online estimate reaches $1.127$\,m horizontal / $1.978$\,m total, and the
batch optimization (Sec.~\ref{sec:batch}) lowers this to $0.919$\,m / $1.327$\,m by
removing the residual global-anchor error left from initialization. The GNSS-aware
LIGO trails at $3.7$\,m horizontal / $12.9$\,m total, while the LiDAR-only and
loosely-coupled FAST-LIO2 and LIO-SAM-GPS degrade to several or tens of meters and the
GNSS-only RTKLIB, unable to fix its carrier-phase ambiguities in the multipath-ridden
canyon, drifts to $38.9$\,m total on its float solution
(Tab.~\ref{tab:comparison_results}). The advantage here is
global consistency: by keeping the raw GNSS factors active whenever any observation is
usable, with the scan-to-multiscan LiDAR factors constraining the directions GNSS
leaves under-determined, GLIO2 re-establishes the absolute reference as the vehicle
moves between obstructed and open segments, eliminating the large accumulated drift that the
LiDAR-inertial baselines exhibit over this route. The residual online error is
concentrated in the vertical: of the $1.978$\,m total, $1.127$\,m is horizontal,
because the dense surrounding structure keeps the scan-to-multiscan horizontal
constraints strong throughout, whereas over the final long GNSS-blocked stretch the
weakly observed vertical channel drifts slowly until reacquisition, the
$250$--$310$\,s span magnified in the inset of Fig.~\ref{fig:traj_urban}; by
re-optimizing the whole trajectory against all GNSS epochs at once, the batch
refinement recovers most of this vertical drift ($1.978\!\to\!1.327$\,m total). The
online point-cloud map and the trajectory comparison are shown in
Figs.~\ref{fig:map_urban} and~\ref{fig:traj_urban}; the callouts of
Fig.~\ref{fig:map_urban} pair each GNSS-degraded spot with its onboard camera view.

\paragraph{\textit{Switch}}
On the $209$\,s indoor--outdoor flight the building interior supplies rich, low-noise
LiDAR geometry that keeps the horizontal tightly constrained. \revwording{GLIO2 (Batch) achieves the lowest 3D ATE RMSE on this sequence.} The online estimate is $0.425$\,m / $0.636$\,m and the batch optimization
improves it to $0.213$\,m / $0.301$\,m, ahead of LIGO ($0.734$\,m / $0.865$\,m) and
well ahead of the GNSS-only RTKLIB, which breaks down to the meter level
(Tab.~\ref{tab:comparison_results}). The online $3$D error is dominated by the
vertical channel ($0.47$\,m of the $0.636$\,m total), which carries a residual
elevation error from the global initialization; because the batch re-optimizes the
whole trajectory jointly with every GNSS epoch, it absorbs most of this
initialization bias and roughly halves the vertical error. GLIO2 maintains a
continuous trajectory through the GNSS-denied indoor passage and rejoins the second
outdoor segment at re-acquisition without discontinuity; the continuous
indoor--outdoor map in Fig.~\ref{fig:loose_tear} (top center) provides the qualitative
map-continuity evidence announced in Sec.~\ref{sec:datasets_protocol_ref}. On this
sequence the scan-to-multiscan LiDAR factors, rather than the GNSS factors, dominate
the horizontal accuracy.

\paragraph{\textit{Bridge}}
On the LiDAR-degenerate \textit{Bridge} (Sec.~\ref{sec:private_datasets}), only GLIO2
completes the sequence with a usable trajectory, $1.625$\,m / $3.586$\,m online and
$1.438$\,m / $2.552$\,m after the batch optimization. The GNSS-only RTKLIB tracks the
route but at $5.634$\,m / $11.367$\,m, its error concentrated in the weak vertical
channel, while LIGO, LIO-SAM-GPS, FAST-LIO2, and GLIO fail in both their online and
offline batch configurations (Tab.~\ref{tab:comparison_results}). GLIO2 stays accurate because its raw
double-difference factor supplies a measurement that is geometrically independent of
the degenerate scan and constrains the along-bridge direction the registration leaves
unobserved. The bridge route and the GLIO2 point-cloud map are shown in
Fig.~\ref{fig:map_deg1}, where the global map stays straight and continuous across the
degenerate span.

\paragraph{\textit{Plain}}
The \textit{Plain} sequence isolates the same degeneracy on the aerial platform over flat,
feature-poor ground. Over this open-sky site kinematic RTK is inherently strong, and RTKLIB is
the best baseline at $0.150$\,m / $0.289$\,m. GLIO2's real-time online estimate is
larger at $0.596$\,m / $0.663$\,m, dominated by an initialization error rather than by
tracking drift; the batch optimization removes most of this initialization error and
reaches $0.142$\,m / $0.162$\,m, the smallest error on the sequence. Its horizontal
margin over RTKLIB is then only $8$\,mm, but the vertical channel is markedly better:
$0.16$\,m versus $0.29$\,m total, an $8$\,cm versus $25$\,cm vertical error. Beyond this margin, the fusion also adds a continuous high-rate state
stream and removes any dependence on carrier-phase ambiguity fixing.
LIGO and LIO-SAM-GPS drift to $9.9$--$17.6$\,m horizontally; FAST-LIO2 fails to
complete the sequence because, without any GNSS factor, the scan registration cannot
recover from the horizontal under-determination over the flat ground, and GLIO
fails in both configurations for the same front-end reason
(Tab.~\ref{tab:comparison_results}). GLIO2 remains accurate for the same reason as \textit{Bridge}: the double-difference
factor constrains the under-determined directions independently of the degenerate scan,
while the scan-to-multiscan factor exposes only the well-determined relative geometry.

\subsection{Ablation Studies}
\label{sec:ablation}

We ablate three components one at a time on the self-collected sequences: the tightly-coupled
versus loosely-coupled GNSS factor, GPU-based versus CPU-based LiDAR
processing, and the scan-to-multiscan versus scan-to-map LiDAR factor.

\subsubsection{Evaluation of Tightly-Coupled GNSS}
\begin{figure}[!tb]
  \centering
  \includegraphics[width=\columnwidth]{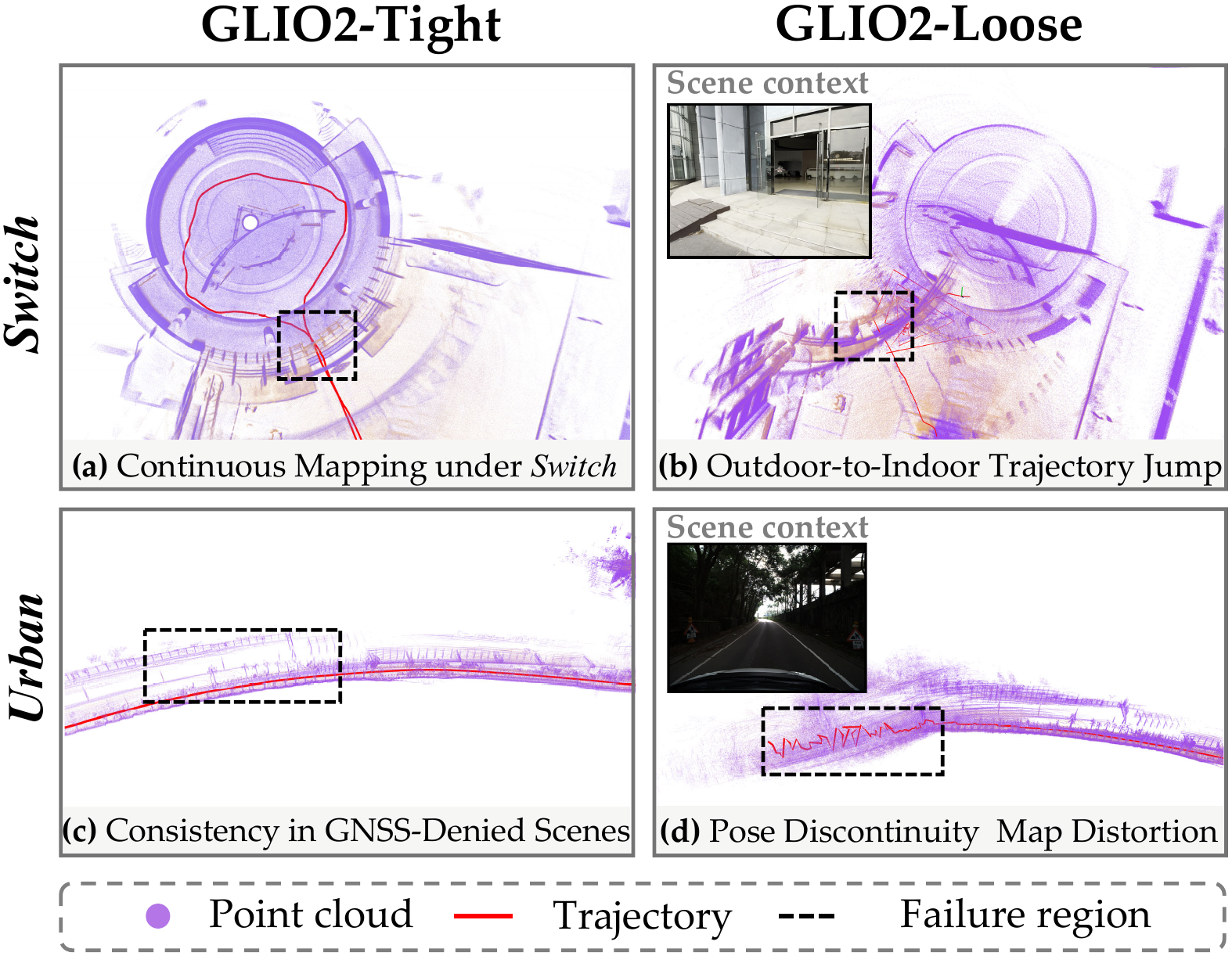}
  \caption{Comparison of GLIO2-Tight and GLIO2-Loose on \textit{Switch} (top) and \textit{Urban} (bottom). GLIO2-Tight \textbf{(a,~c)} keeps a clean, continuous map, whereas GLIO2-Loose \textbf{(b,~d)} injects trajectory jumps that tear the map: at the outdoor-to-indoor transition \textbf{(b)} and under tree-shaded GNSS dropouts \textbf{(d)}.}
  \label{fig:loose_tear}
\end{figure}

This ablation replaces the raw double-differenced pseudorange and Doppler factors with the
u-blox F9P receiver's own position fix, fed into the same sliding-window graph as a
position-level prior (denoted GLIO2-Loose); everything else is held fixed. We evaluate it
on the two GNSS-challenged sequences, \textit{Switch} and
\textit{Urban} (Tab.~\ref{tab:ablation}). The loose variant diverges on \textit{Switch} and drifts to $10.9$\,m on \textit{Urban}, whereas GLIO2-Tight stays robust on both
(Fig.~\ref{fig:loose_tear}). On \textit{Switch}, at the UAV's outdoor-to-indoor transition the
loose prior introduces a trajectory jump that tears the point-cloud map, and the
estimator ultimately diverges; the same failure recurs on \textit{Urban}, triggered there
by the prolonged tree-shaded segments that drop GNSS lock.
GLIO2-Tight instead stays smooth and continuous on both. The difference is that a loose
position prior accepts or rejects the receiver's solution as a whole: once that
solution degrades or is lost it carries no usable covariance, the global anchor is lost,
and the trajectory jumps when the fix returns. Tight raw-observation coupling instead
weights each satellite individually, down-weighting the corrupted observations
while still drawing a geometric constraint from the satellites that remain locked, so a
partial and noisy GNSS view stays usable rather than all-or-nothing.

\subsubsection{Evaluation of GPU Acceleration}
This ablation keeps the objective identical but moves the dense per-point LiDAR
linearization from the GPU back to the CPU. We run it on \textit{Switch} (Livox Mid-360)
and \textit{Urban} (Velodyne VLP-32E), spanning a light and a dense LiDAR
(Tab.~\ref{tab:ablation}). The CPU pipeline cannot hold the $100$\,ms scan
budget: it drops scans and loses accuracy, with 3D RMSE rising to $2.61$\,m on
\textit{Switch} and $4.06$\,m on \textit{Urban}, whereas the GPU pipeline stays
well under budget ($5$--$13\times$ headroom) at full accuracy. \revwording{On the Intel Core Ultra~7 265KF desktop platform, the CPU pipeline averages}
$67.90$\,ms per scan on \textit{Switch} and $99.82$\,ms on \textit{Urban};
the means fit the budget, but a portion of scans reach
$200$--$300$\,ms, two to three times the $100$\,ms period at $10$\,Hz, so the CPU
\revwording{pipeline still drops scans in this configuration.}
\revwording{We attribute the accuracy loss primarily to the constraints lost with the dropped scans.} The
GPU pipeline, by contrast, averages $7.43$\,ms per scan on \textit{Switch} and $20.41$\,ms on
\textit{Urban} while retaining full accuracy (the runtime analysis that follows gives the
per-scan breakdown). \revwording{In our configuration, the GPU linearization inherited from GLIM~\cite{glim} enables processing at the LiDAR rate while retaining the accuracy of the tightly-coupled scan-to-multiscan formulation.}

\begin{table}[tb]
\centering
\caption{GNSS-coupling and GPU ablation: horizontal and 3D ATE RMSE (m) on \textnormal{\textit{Switch}} and \textnormal{\textit{Urban}}.}
\label{tab:ablation}
\small
\setlength{\tabcolsep}{5pt}
\renewcommand{\arraystretch}{1.2}
\begin{tabular}{@{}lcccc@{}}
\toprule
\multirow{2}{*}{Variant} & \multicolumn{2}{c}{Switch} & \multicolumn{2}{c}{Urban} \\
\cmidrule(lr){2-3}\cmidrule(lr){4-5}
 & Hor.\ & 3D & Hor.\ & 3D \\
\midrule
GLIO2-Tight             & \textbf{0.425} & \textbf{0.636} & \textbf{1.127} & \textbf{1.978} \\
w/o tight GNSS (Loose)  & \multicolumn{2}{c}{\textit{diverges}} & 10.884 & 12.705 \\
w/o GPU (CPU)           & 1.570 & 2.608 & 2.569 & 4.064 \\
\bottomrule
\end{tabular}

\end{table}

\subsubsection{Evaluation of Scan-to-Multiscan}

\begin{figure}[tb]
  \centering
  \includegraphics[width=\columnwidth]{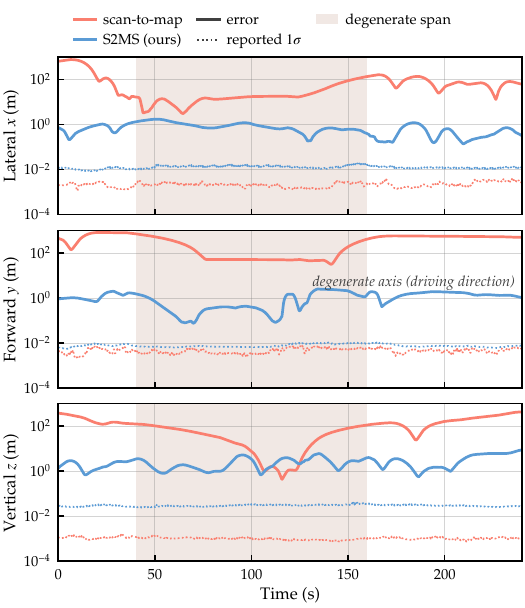}
  \caption{Architecture ablation on the \textit{Bridge}: per-axis position error and the estimator's reported $1\sigma$ for the scan-to-map and S2MS variants in the body frame (lateral $x$; forward $y$, the driving direction; vertical $z$), on a log scale; the shaded band marks the LiDAR-degenerate span. Within it the scan-to-map along-track ($y$) error grows by orders of magnitude while its reported $1\sigma$ stays millimeter-level, \revwording{whereas the S2MS fusion keeps the along-track error at the meter level.}}
  \label{fig:ablation_overconf}
\end{figure}

\revwording{To test how the LiDAR constraint formulation affects fusion, we replace only the relative S2MS factors with a fixed-map pose prior (Sec.~\ref{sec:lidar_fusion}),}
holding the IMU, raw GNSS, and sliding-window graph identical to GLIO2: each scan
enters as an absolute pose-with-covariance prior whose information is the converged
\revwording{matching Hessian (the $\mathbf{H}_{ii}$ block of~\eqref{eq:lidar_hessian_blocks}).} We run it on the degenerate \textit{Bridge} and the
well-conditioned \textit{Urban} (Tab.~\ref{tab:ablation_s2m},
Fig.~\ref{fig:ablation_overconf}). The handoff diverges to $669.6$\,m RMSE on
\textit{Bridge} and drifts to $11.46$\,m on \textit{Urban}, whereas the relative form
stays at $3.586$\,m and $1.978$\,m.

\revwording{The two variants share the VGICP information model in \eqref{eq:vgicp_info}--\eqref{eq:lidar_hessian_blocks}; the ablation changes how the LiDAR constraints enter the graph. The S2M handoff attaches the matching information to each pose as a unary prior, which also constrains the common rigid motion of the window (Fig.~\ref{fig:hessian_blocks}(a)) and is therefore weighed directly against GNSS in those directions. The relative S2MS factors instead contribute zero information along this common-motion subspace (Sec.~\ref{sec:lidar_fusion}), so they constrain the relative geometry without directly anchoring the global frame. The results in Tab.~\ref{tab:ablation_s2m} and Fig.~\ref{fig:ablation_overconf} are consistent with this structural difference. On \textit{Bridge}, the S2M error accumulates along the degenerate driving direction to hundreds of meters while the reported $1\sigma$ along that axis remains at the millimeter level. S2MS keeps the along-track error at the meter level and reports a larger $3$D $1\sigma$ than the S2M variant ($3.31$ versus $0.51$\,cm). Both fusion variants remain overconfident, with reported posterior uncertainties smaller than the observed errors. Nevertheless, the closer agreement between reported uncertainty and observed error in the S2MS variant suggests improved empirical consistency relative to the S2M handoff in this comparison.}

\begin{table}[tb]
\centering
\caption{Architecture ablation on the degenerate \textnormal{\textit{Bridge}} and the well-conditioned \textnormal{\textit{Urban}} sequence.}
\label{tab:ablation_s2m}
\small
\setlength{\tabcolsep}{4pt}
\renewcommand{\arraystretch}{1.15}
\resizebox{\columnwidth}{!}{%
\begin{tabular}{llcccc}
\toprule
Sequence & LiDAR factor & ATE RMSE (m) & 2D $1\sigma$ (cm) & 3D $1\sigma$ (cm) & Outcome \\
\midrule
\multirow{2}{*}{Bridge} & Scan-to-map & $669.6$ & $0.50$ & $0.51$ & diverges \\
 & \textbf{S2MS (ours)} & $\mathbf{3.586}$ & $1.45$ & $3.31$ & \textbf{bounded} \\
\midrule
\multirow{2}{*}{Urban} & Scan-to-map & $11.46$ & $0.22$ & $0.25$ & \revwording{drifts} \\
 & \textbf{S2MS (ours)} & $\mathbf{1.978}$ & $0.86$ & $1.01$ & \textbf{bounded} \\
\bottomrule
\end{tabular}}
\end{table}

\begin{figure*}[!t]
    \centering
    \subfloat[Jetson Orin~NX edge platform.\label{fig:runtime_orin}]{%
        \includegraphics[width=0.49\textwidth]{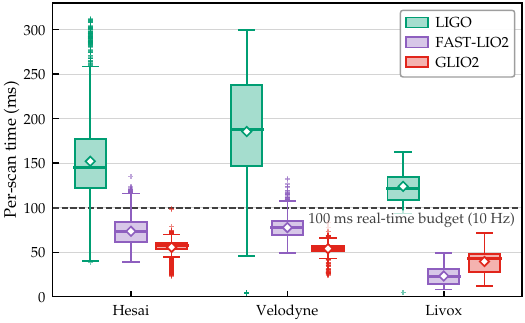}}\hfill
    \subfloat[Desktop platform.\label{fig:runtime_desktop}]{%
        \includegraphics[width=0.49\textwidth]{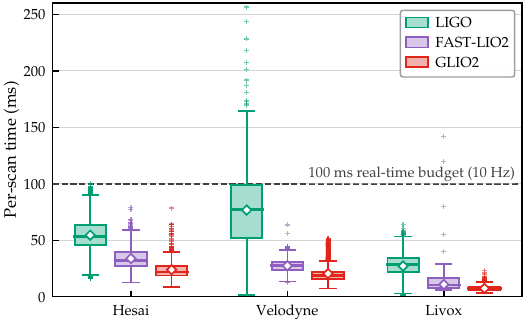}}
    \caption{Per-scan runtime distributions across the three LiDAR types on \textbf{(a)}~the Jetson Orin~NX edge platform and \textbf{(b)}~the desktop platform. The dashed line marks the $100$\,ms real-time budget at the $10$\,Hz LiDAR rate; for LIGO on the Velodyne in \textbf{(a)}, readouts above $300$\,ms ($8.1\%$ of scans) are excluded. In each box the center line is the median and the white diamond ($\Diamond$) is the mean, the whiskers span the $1.5\times$IQR range, and ``+'' marks outliers.}
    \label{fig:runtime}
\end{figure*}

\begin{table}[!t]
\centering
\caption{Per-scan runtime (ms) on the Jetson Orin~NX and desktop platforms: per-scan total of LIGO, FAST-LIO2, and GLIO2, with the GLIO2 stage breakdown. Fastest total per row in bold.}
\label{tab:runtime_breakdown}
\label{tab:runtime_comparison}
\footnotesize
\setlength{\tabcolsep}{2.7pt}
\renewcommand{\arraystretch}{1.1}
\begin{tabular}{lccc@{\hspace{10pt}}ccc}
\toprule
& \multicolumn{3}{c}{\textbf{Per-scan total (ms)}} & \multicolumn{3}{c}{\textbf{GLIO2 breakdown (ms)}} \\
\cmidrule(lr){2-4} \cmidrule(lr){5-7}
\textbf{LiDAR} & LIGO & FAST-LIO2 & GLIO2 & Preproc. & Opt.\ upd. & Residual \\
\midrule
\multicolumn{7}{l}{\textit{Jetson Orin~NX}} \\
\enspace Hesai    & 152.19 & 73.56          & \textbf{55.54} & 15.26 & 32.06 & 8.21 \\
\enspace Velodyne & 185.60 & 77.72          & \textbf{54.02} & 11.36 & 35.24 & 7.42 \\
\enspace Mid-360   & 123.96 & \textbf{23.28} & 39.60          &  5.27 & 28.04 & 6.28 \\
\midrule
\multicolumn{7}{l}{\textit{Desktop}} \\
\enspace Hesai    &  54.50 & 33.81 & \textbf{23.68} &  7.71 & 11.41 & 4.56 \\
\enspace Velodyne &  76.72 & 27.46 & \textbf{20.41} &  5.44 & 10.64 & 4.33 \\
\enspace Mid-360   &  27.22 & 10.84 & \textbf{7.43}  &  2.55 &  2.32 & 2.56 \\
\bottomrule
\end{tabular}
\end{table}

\subsection{Runtime Analysis}

\subsubsection{Front-End Runtime}
We profile the GLIO2 front-end runtime on the Jetson Orin~NX edge
platform and on a desktop platform for reference, across three representative
LiDAR types to probe cross-LiDAR generalization: the Livox Mid-360, the Velodyne
VLP-32E (on which all our \textit{Urban} and \textit{Bridge} accuracy results are
obtained), and the denser Hesai XT-32. The measured modules include LiDAR
preprocessing, optimization update (the iSAM2 update), and the residual odometry
stages. Runtime is measured after initialization to avoid mixing startup overhead
with steady-state performance, and queue waiting is excluded. At $10$\,Hz the
per-scan budget is $100$\,ms; we call the pipeline real-time when these stages together stay below it for most frames.

Tab.~\ref{tab:runtime_breakdown} reports the per-scan runtime
breakdown on both platforms. On the desktop, the total per-scan time
is $7.43$\,ms for the Livox Mid-360, $20.41$\,ms for the Velodyne VLP-32E, and
$23.68$\,ms for the Hesai LiDAR, all well below the $100$\,ms real-time budget;
on the Jetson Orin~NX the corresponding totals are $39.60$\,ms, $54.02$\,ms,
and $55.54$\,ms, still within the $100$\,ms budget.
The same table compares the total per-scan time against LIGO and
FAST-LIO2: GLIO2 achieves the lowest
total time on the Hesai and Velodyne LiDARs across both platforms; on the
Mid-360 it is fastest on the desktop and second to FAST-LIO2 on the Orin~NX. The contrast that matters for deployment is on the edge platform (Fig.~\ref{fig:runtime_orin}): on the Orin~NX, LIGO exceeds the $100$\,ms budget on all three LiDARs
($123.96$--$185.60$\,ms), so it does not sustain the LiDAR rate onboard, whereas
FAST-LIO2 and GLIO2 stay within budget on every LiDAR. On the Mid-360, FAST-LIO2 is
faster ($23.28$ versus $39.60$\,ms), because GLIO2 also linearizes the GNSS factors that
FAST-LIO2 omits and the sparse solid-state cloud gives the per-point GPU parallelization
far less to amortize than the dense mechanical scans do; both still fit the $100$\,ms scan
period, so the difference does not affect real-time feasibility. The stage breakdown in the same
table locates the GLIO2 bottleneck: on the Orin~NX the CPU-side iSAM2 update takes
$28$--$35$\,ms ($58$--$71\%$ of the per-scan total), while preprocessing and the
residual odometry together account for the remaining $11$--$24$\,ms. With the
per-point linearization on the GPU, the sparse solve, not the dense geometry, is
the limiting stage. The desktop (Fig.~\ref{fig:runtime_desktop}) shows the same pattern at smaller scale, the
Mid-360 update dropping to $2.32$\,ms against $10$--$11$\,ms for the denser Hesai
and Velodyne scans. \revwording{Across all three LiDARs, the medians and upper whiskers of GLIO2's runtime distributions remain below the $100$\,ms scan period on the Orin~NX, supporting real-time feasibility on the evaluated sequences (Fig.~\ref{fig:runtime}).}

\subsubsection{Back-End Runtime}
\begin{figure}[!t]
    \centering
    \includegraphics[width=\columnwidth]{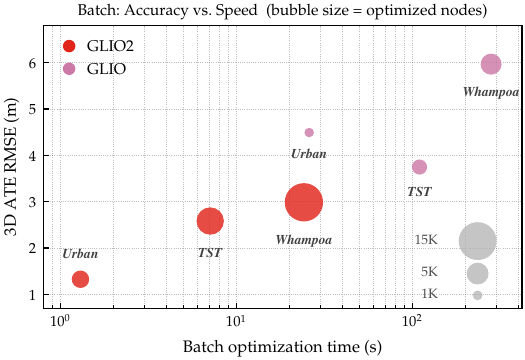}
    \caption{Back-end batch optimization over the three sequences: $3$D~ATE RMSE versus full
    optimization time (log scale), with bubble size the number of optimized
    nodes. GLIO2 reaches lower error in far less time while optimizing
    a larger graph.}
    \label{fig:backend_bubble}
\end{figure}

Beyond the per-scan online budget, the offline batch optimization of
Sec.~\ref{sec:batch} is itself both fast and accurate. On the same machine we
compare it against the back-end of the state-of-the-art GLIO~\cite{glio} on the
\textit{Urban}, UrbanNav~\textit{Tst}, and \textit{Whampoa} sequences, the
longest in duration and trajectory and the densest in point cloud in our
evaluation (Fig.~\ref{fig:backend_bubble}). Because GLIO2 optimizes every frame
while GLIO retains only a keyframe subset (roughly one frame in three), its graph
carries about $3.3\times$ as many nodes, $15{,}346$ versus $4{,}603$ on the
$30$-min \textit{Whampoa} sequence. Despite the larger graph, GLIO2 attains both a lower
$3$D RMSE and a far shorter runtime: $24.2$\,s versus $280$\,s on \textit{Whampoa}
(${\approx}12\times$), $7.1$\,s versus $110$\,s on \textit{Tst} (${\approx}15\times$),
and $1.3$\,s versus $25.9$\,s on the shorter \textit{Urban} (${\approx}20\times$).
GLIO2 thus optimizes a larger graph faster and to higher accuracy.

\section{Applications}
\label{sec:application}
To demonstrate the practicality of GLIO2 in real-world tasks, we deploy it as the
sole onboard localization and mapping source in two complete autonomous systems:
a facade-cleaning UAV and a guide-dog quadruped for the visually impaired.
Together they stress the three properties the estimator is built for: robustness
under simultaneous LiDAR and GNSS degradation, global consistency over a long
map-referenced route, and real-time operation fully onboard.

\subsection{Autonomous Facade-Cleaning UAV}
Cleaning a high-rise glass facade is conventionally done by workers suspended on
ropes, which is hazardous, or by a pilot manually flying a cleaning drone along
the wall, which is inefficient and crash-prone. Autonomy is hard here because
both failure modes this paper targets occur at once. At a $1$--$2.5$\,m standoff
from the facade, the LiDAR sees only flat, self-similar glass and degenerates
along the wall, and the building blocks much of the sky, so few GNSS satellites
remain visible. GLIO2 is the sole localization source in the flight-control loop
of the UAV (the $823$\,mm-class platform of Fig.~\ref{fig:platforms}). It runs
fully onboard the Jetson Orin~NX and returns the IMU-rate state to the controller
well within the per-scan budget. \revwording{At this close standoff, smooth
state estimates are critical to wall clearance: abrupt localization changes,
such as the map-tearing behavior observed with a loose GNSS prior
(Sec.~\ref{sec:ablation}), can induce unsafe control commands.}

\begin{figure}[!tb]
  \centering
  \includegraphics[width=\columnwidth]{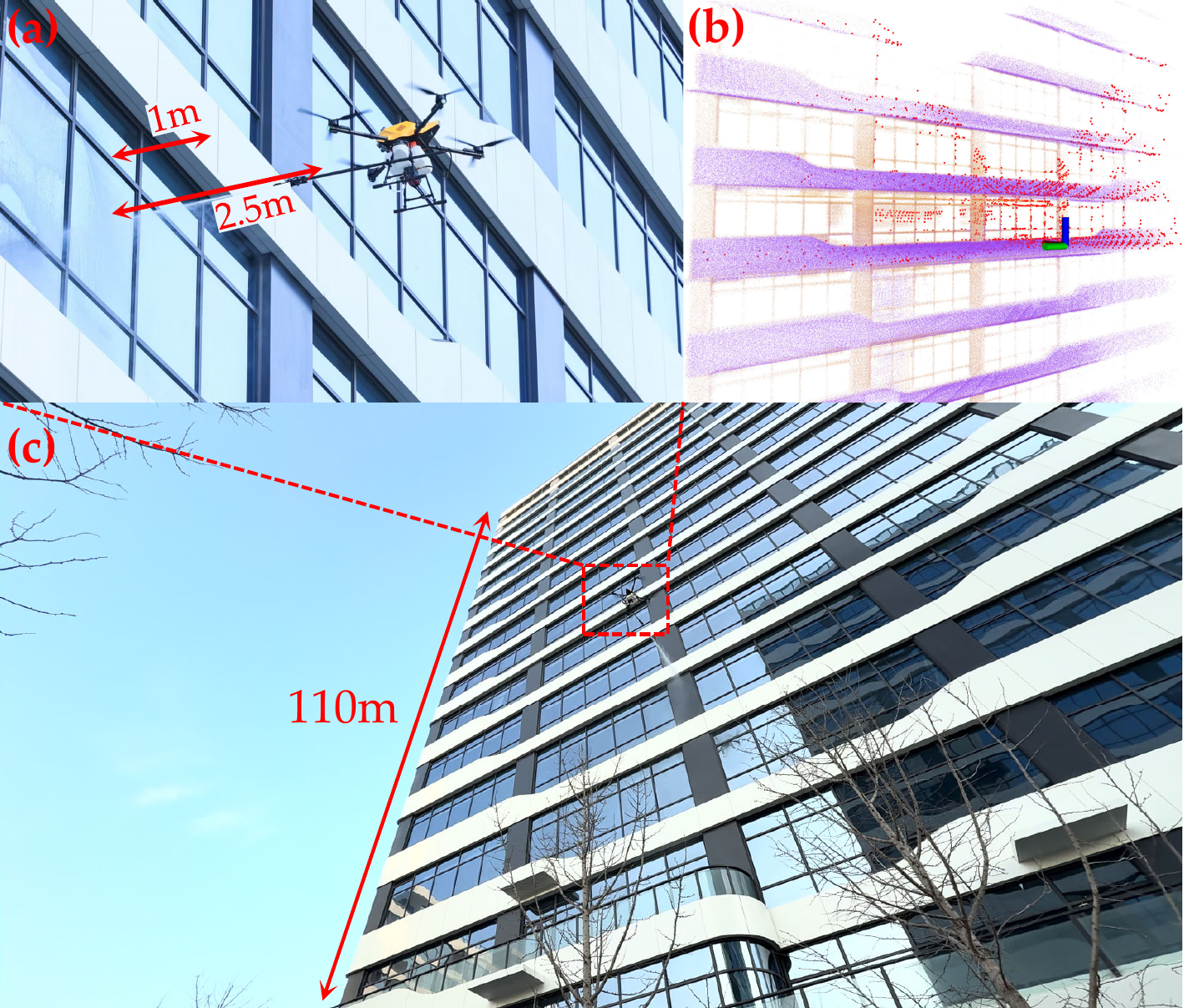}
  \caption{Autonomous facade-cleaning UAV. \textbf{(a)}~Close view of the UAV holding a $1$--$2.5$\,m standoff from the glass curtain wall (annotated distances are UAV-to-wall). \textbf{(b)}~Online GLIO2 LiDAR point-cloud map of the repetitive facade, whose self-similar planar structure causes LiDAR degeneracy along the wall, with the estimated UAV pose. \textbf{(c)}~The $110$\,m high-rise during the cleaning operation, with the UAV working along the facade.}
  \label{fig:application_uav}
\end{figure}

\revwording{We evaluated $10$ cleaning missions, each lasting $5$--$7$\,min, and
defined autonomous success as task completion without manual takeover. The UAV
completed $8$ missions autonomously ($80\%$), while $2$ required operator
takeover.} Fig.~\ref{fig:application_uav} shows the UAV sweeping the glass
curtain wall of a $110$\,m high-rise.

\revwording{Both takeovers followed misclassification of cleaning water spray
as obstacles. The resulting obstacle representation placed the UAV inside an
apparent obstacle, causing the planner to suspend trajectory generation.
GLIO2 continued to provide stable localization, allowing the operator to take
manual control and land. The two missions are counted as failures of autonomous
task completion; localization continuity supported manual recovery. These
trials expose a perception and planning limitation of the integrated cleaning
system under spray interference.}

\subsection{Guide-Dog Navigation for the Visually Impaired}
Guiding a visually impaired user to a named destination requires the route and
goal to be expressed in a global, Earth-fixed frame. Pure LiDAR-inertial
odometry provides no such frame: it estimates motion relative to its start, so a
goal is only an offset from wherever the system happened to initialize, and the
real-world point that offset reaches changes from run to run. Heading is the
worst case, as it is unobservable from LiDAR-inertial data alone: a $5^{\circ}$
error in the initial yaw rotates the whole trajectory and, over a long route,
displaces the destination by tens of meters. This is unsafe for an assistive
task, since the drift can lead the user to the wrong destination. GLIO2 instead estimates the state directly in
the GNSS-anchored Earth frame, so the route and goal are globally referenced and
reproducible across runs, \revwording{supporting navigation to a fixed,
map-referenced destination.}

\begin{figure}[!tb]
  \centering
  \includegraphics[width=\columnwidth]{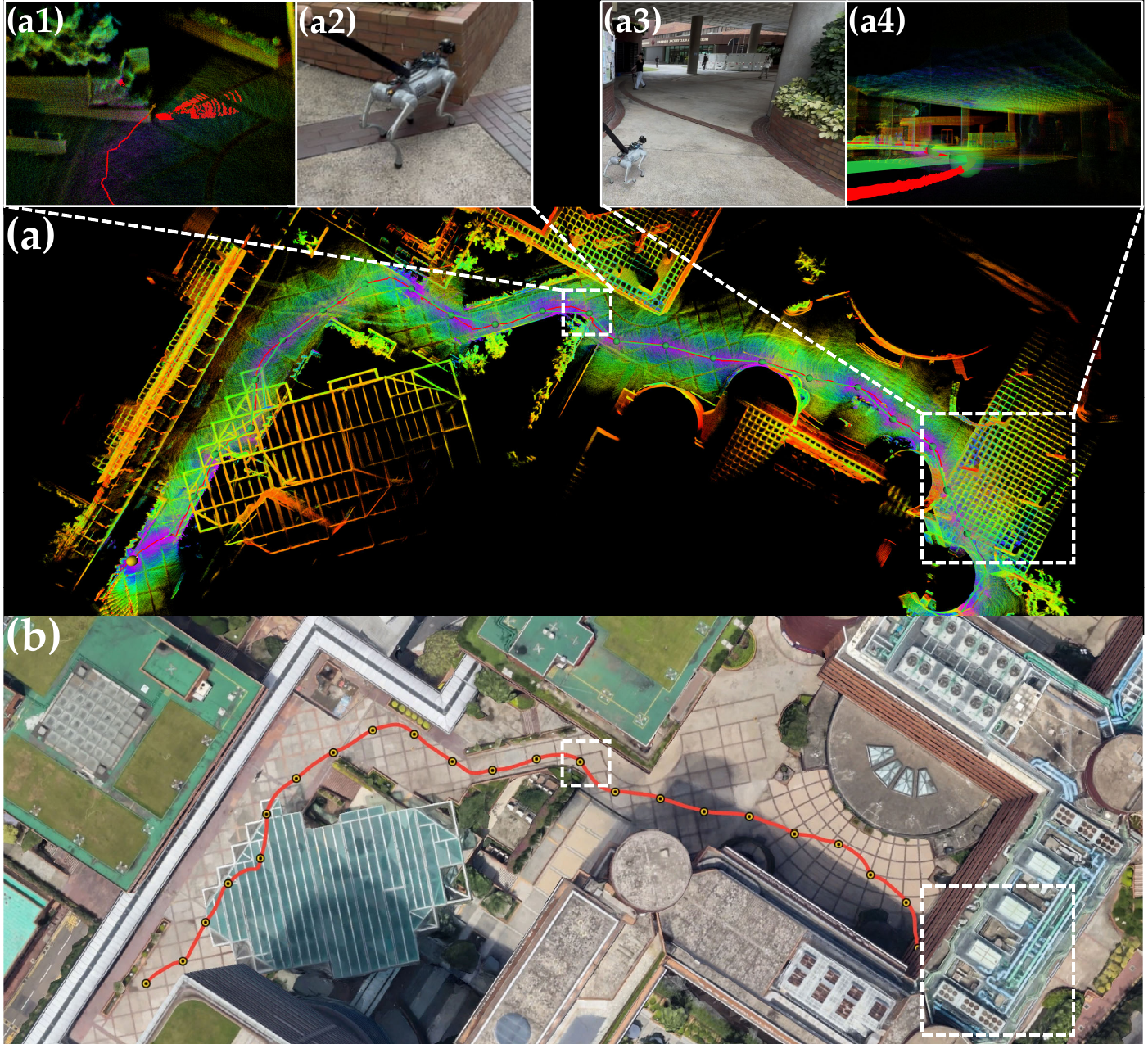}
  \caption{Guide-dog navigation for the visually impaired. \revwording{\textbf{(a)}~Online GLIO2 point-cloud map. \textbf{(b)}~Route on Google Maps imagery. \textbf{(a1)}~Point cloud and \textbf{(a2)}~real-world view of the LiDAR-degenerate section; \textbf{(a3)}~real-world view and \textbf{(a4)}~point cloud of the GNSS-denied section.}}
  \label{fig:application_dog}
\end{figure}

\revwording{We repeated the $210$\,m guided route $5$ times, with each walk
taking approximately $7$\,min. The route traverses a LiDAR-degenerate stretch
(Fig.~\ref{fig:application_dog}(a1), (a2)) followed by a GNSS-denied stretch
(Fig.~\ref{fig:application_dog}(a3), (a4)). All $5$ runs reached the intended
destination without manual intervention. GLIO2 supplied localization throughout,
including during the GNSS-denied segment.}

\section{Conclusion and Future Work}
\label{sec:conclusion}

This paper proposed GLIO2, a unified, online, tightly-coupled
LiDAR--Inertial--GNSS system that fuses IMU pre-integration, dense
scan-to-multiscan LiDAR, and raw double-differenced pseudorange and
Doppler in a single sliding-window factor graph, comprising a GPU-parallel
online front-end, an efficient offline back-end, and a single-receiver variant
for base-station-free settings.

GLIO2's primary strengths include:
\begin{itemize}
\item \textit{Robustness to LiDAR degeneracy:} on a $5.66$-km bridge traversed
at up to $96$\,km/h, where the open deck offers no along-track structure, GLIO2
\revwording{achieved $1.6$\,m horizontal accuracy online while the evaluated LiDAR-based baselines diverged.}
\item \textit{Global consistency under GNSS NLOS:} on the $30$-min, $4.51$-km
\revwording{UrbanNav~\textit{Whampoa} sequence through dense high-rises, GLIO2 (Batch) was the most accurate among the evaluated systems, at $1.96$\,m horizontal and $2.99$\,m 3D.}
\item \textit{Real-time edge deployment:} \revwording{the full pipeline averages $39.60$--$55.54$\,ms per scan across three LiDAR types on an NVIDIA Jetson Orin~NX,} and the offline back-end
re-optimizes an entire $30$-min session in about $24$\,s.
\end{itemize}

The first two follow from keeping LiDAR and GNSS as raw measurements in a single
joint optimization: the dense scan-to-multiscan factors stay re-associable and
relinearizable online together with the raw GNSS factors\revwording{. The smaller discrepancy between reported posterior uncertainty and observed error in the S2MS ablation variant suggests improved empirical consistency relative to the S2M handoff (Tab.~\ref{tab:ablation_s2m}, Fig.~\ref{fig:ablation_overconf}).} \revwording{The third builds on the GPU scan-matching factors of GLIM~\cite{glim}, which offload dense per-point LiDAR computation and enable the joint estimator to operate at the LiDAR rate on the evaluated platforms.}

Several limitations remain and indicate directions for future work:
\begin{itemize}
\item \textit{GNSS-LIO initialization:} While the double-differenced
configuration initializes more accurately and converges faster than the
single-receiver variant, both still depend on adequate GNSS at startup, and a
poor initial GNSS condition or an initialization performed under dynamic motion
degrades the fused estimate. Future work will pursue a more robust
initialization that tolerates degraded and in-motion startup conditions.
\item \textit{GNSS quality management:} A reliable GNSS quality-management
module is essential for a LiDAR--Inertial--GNSS fusion system, yet our current
treatment is limited to elevation/SNR weighting and residual-based fault
detection and de-weighting (FDD). Future work will develop a more refined
quality-management scheme that more accurately identifies and down-weights NLOS
and multipath measurements.
\end{itemize}
The source code and self-collected datasets will be released.

\FloatBarrier
\ifCLASSOPTIONcaptionsoff
  \newpage
\fi

\bibliographystyle{IEEEtran}
\IEEEtriggeratref{46}
\bibliography{IEEEabrv,Bibliography}

\clearpage
\input{appendix_body}
\end{document}

%% file: glio2_preamble.tex
\ifdefined\pdfminorversion\pdfminorversion=7\fi
\usepackage[cmex10]{amsmath}
\usepackage{graphicx}
\graphicspath{{pdf/}}
\DeclareGraphicsExtensions{.pdf,.jpeg,.png}
\usepackage{array}
\usepackage{tabularx}
\usepackage{url}
\usepackage{cite}
\usepackage{booktabs}
\usepackage{amssymb}
\usepackage[ruled,vlined,linesnumbered]{algorithm2e}
\usepackage[table]{xcolor}
\newif\ifmarkchanges
\markchangesfalse
\newcommand{\revwording}[1]{\ifmarkchanges\textcolor{red}{#1}\else#1\fi}

\definecolor{algcomment}{RGB}{106,115,125}
\definecolor{lidargreen}{RGB}{0,158,115}
\definecolor{gpuamber}{RGB}{198,118,10}
\definecolor{oursshade}{RGB}{253,240,238}
\definecolor{headshade}{RGB}{233,236,241}

\usepackage{multirow}
\usepackage{courier}
\usepackage[caption=false,font=footnotesize]{subfig}
\usepackage{placeins}

\usepackage[hidelinks]{hyperref}
\hypersetup{
  pdftitle={GLIO2: A GPU-Parallelized Tightly-Coupled LiDAR-Inertial-GNSS System for Robust and Real-Time Global Localization and Mapping},
  pdfauthor={Qi Zhang, Xikun Liu, Qijun Qin, Xiangru Wang, Junzhe Wang, Naigui Xiao, Jianhao Jiao, and Weisong Wen},
  pdfsubject={GLIO2 arXiv preprint},
  pdfkeywords={Localization, Mapping, Sensor Fusion, Global Navigation}
}

%% file: glio2_arxiv_layout.tex
\AtBeginDocument{%
  \definecolor{tmlcncolor}{cmyk}{0.93,0.59,0.15,0.02}%
  \definecolor{NavyBlue}{RGB}{0,86,125}%
}

\usepackage{etoolbox}
\makeatletter
\def\p@subsection{\thesection-}
\def\p@subsubsection{\thesection-\thesubsection}

\patchcmd{\@maketitle}
  {{\receivedfont \ifx\@receiveddate\@empty\else Received\ \@receiveddate\fi\ifx\@revised\@empty\else;\ revised\ \@reviseddate\fi\ifx\@accepteddate\@empty\else;\ accepted\ \@accepteddate\fi\ifx\@publisheddate\@empty\else; Date of publication\nobreakspace\@publisheddate\fi\ifx\@currentdate\@empty\else;\space date of current version\nobreakspace\@currentdate\fi.\space\ifx\@editor\@empty\else\@editor\fi\par}}
  {}
  {}
  {\PackageError{glio2-arxiv}{Could not suppress the template publication history}{Check the version of ieeetj.cls.}}

\def\authorfont{\fontfamily{\sfdefault}\fontsize{10}{13.5}\selectfont\bfseries
  \leftskip0pt plus1fill\rightskip0pt plus1fill}

\patchcmd{\@affil}
  {\textsuperscript{\the\tempcount}}
  {}
  {}
  {\PackageError{glio2-arxiv}{Could not suppress the single affiliation marker}{Check the version of ieeetj.cls.}}

\def\OJlogo{}
\def\seclogo{}
\def\ps@glioarxiv{%
  \let\@oddhead\@empty
  \let\@evenhead\@empty
  \def\@oddfoot{\hbox to\textwidth{\hfil{\rffont\thepage}\hfil}}%
  \let\@evenfoot\@oddfoot
}
\let\ps@plain\ps@glioarxiv
\let\ps@headings\ps@glioarxiv
\AtBeginDocument{\pagestyle{glioarxiv}}
\makeatother

%% file: appendix_body.tex
\appendices
\numberwithin{equation}{section}

\section{GNSS Factor Jacobians}
\label{app:gnss_jacobians}
All Jacobians below use the unit line-of-sight vector $\mathbf{e}_{r}^{s}$ of
\eqref{eq:los}, pointing from the receiver to satellite $s$
(Table~\ref{tab:nomenclature}); the geometric-range derivative is then
$\partial\|{}^{E}\mathbf{p}^{s}-{}^{E}\mathbf{p}_{r}\|/\partial{}^{E}\mathbf{p}_{r}
=-(\mathbf{e}_{r}^{s})^{T}$. The Jacobians of the DD pseudorange residual
$r_{\rho}^{mi}$ (Sec.~\ref{sec:dd_psr}) with respect to the navigation state are
\begin{align}
\mathbf{J}_{{}_{i}^{w}\mathbf{p}}
&= -(\mathbf{e}_{r}^{m}-\mathbf{e}_{r}^{i})^{T}\cdot{}_{w}^{E}\mathbf{R}
\label{eq:app_dd_position}\\
\mathbf{J}_{{}_{i}^{w}\mathbf{R}}
&= (\mathbf{e}_{r}^{m}-\mathbf{e}_{r}^{i})^{T}\cdot{}_{w}^{E}\mathbf{R}
\cdot{}_{i}^{w}\mathbf{R}\bigl\lfloor{}^{i}\mathbf{l}_{r}\bigr\rfloor_{\times}
\label{eq:app_dd_rotation}\\
\mathbf{J}_{{}_{w}^{E}\mathbf{R}}
&= (\mathbf{e}_{r}^{m}-\mathbf{e}_{r}^{i})^{T}\cdot{}_{w}^{E}\mathbf{R}
\bigl\lfloor{}_{i}^{w}\mathbf{R}\cdot{}^{i}\mathbf{l}_{r}
+{}_{i}^{w}\mathbf{p}\bigr\rfloor_{\times}
\label{eq:app_dd_global_rotation}\\
\mathbf{J}_{{}_{w}^{E}\mathbf{p}}
&= -(\mathbf{e}_{r}^{m}-\mathbf{e}_{r}^{i})^{T}
\label{eq:app_dd_global_position}
\end{align}
Only the rover line-of-sight terms appear because the base-station range terms
in \eqref{eq:dd_pseudorange} depend on the fixed base coordinates
${}^{E}\mathbf{p}_{b}$, not on the state, and therefore contribute no Jacobian.

The Jacobians of the Doppler residual $r_{D}^{s}$ (Sec.~\ref{sec:doppler}) with
respect to the local velocity, the body orientation and the gyroscope bias
(both through the lever-arm term), the global extrinsic, and the
constellation-specific receiver clock drift are
\begin{align}
\mathbf{J}_{{}_{i}^{w}\mathbf{v}}
&= (\mathbf{e}_{r}^{s})^{T}\cdot{}_{w}^{E}\mathbf{R}
\label{eq:app_doppler_velocity}\\
\mathbf{J}^{D}_{{}_{i}^{w}\mathbf{R}}
&= -(\mathbf{e}_{r}^{s})^{T}{}_{w}^{E}\mathbf{R}{}_{i}^{w}\mathbf{R}
\Bigl\lfloor\bigl\lfloor{}^{i}\boldsymbol{\omega}\bigr\rfloor_{\times}
{}^{i}\mathbf{l}_{r}\Bigr\rfloor_{\times}\notag\\
&\quad-\frac{({}^{E}\mathbf{v}^{s}-{}^{E}\mathbf{v}_{r})^{T}
\bigl(\mathbf{I}-\mathbf{e}_{r}^{s}(\mathbf{e}_{r}^{s})^{T}\bigr)}
{\|{}^{E}\mathbf{p}^{s}-{}^{E}\mathbf{p}_{r}\|}
{}_{w}^{E}\mathbf{R}{}_{i}^{w}\mathbf{R}
\bigl\lfloor{}^{i}\mathbf{l}_{r}\bigr\rfloor_{\times}
\label{eq:app_doppler_rotation}\\
\mathbf{J}_{{}^{i}\mathbf{b}_{g}}
&= (\mathbf{e}_{r}^{s})^{T}\cdot{}_{w}^{E}\mathbf{R}\cdot{}_{i}^{w}\mathbf{R}
\bigl\lfloor{}^{i}\mathbf{l}_{r}\bigr\rfloor_{\times}
\label{eq:app_doppler_gyro_bias}\\
\mathbf{J}^{D}_{{}_{w}^{E}\mathbf{R}}
&= -(\mathbf{e}_{r}^{s})^{T}{}_{w}^{E}\mathbf{R}\notag\\
&\quad\cdot\Bigl\lfloor{}_{i}^{w}\mathbf{v}+{}_{i}^{w}\mathbf{R}
\bigl\lfloor{}^{i}\boldsymbol{\omega}\bigr\rfloor_{\times}
{}^{i}\mathbf{l}_{r}\Bigr\rfloor_{\times}\notag\\
&\quad-\frac{({}^{E}\mathbf{v}^{s}-{}^{E}\mathbf{v}_{r})^{T}
\bigl(\mathbf{I}-\mathbf{e}_{r}^{s}(\mathbf{e}_{r}^{s})^{T}\bigr)}
{\|{}^{E}\mathbf{p}^{s}-{}^{E}\mathbf{p}_{r}\|}\notag\\
&\quad\cdot{}_{w}^{E}\mathbf{R}
\bigl\lfloor{}_{i}^{w}\mathbf{R}{}^{i}\mathbf{l}_{r}
+{}_{i}^{w}\mathbf{p}\bigr\rfloor_{\times}
\label{eq:app_doppler_global_rotation}\\
\mathbf{J}^{D}_{{}_{w}^{E}\mathbf{p}}
&= \frac{({}^{E}\mathbf{v}^{s}-{}^{E}\mathbf{v}_{r})^{T}
\bigl(\mathbf{I}-\mathbf{e}_{r}^{s}(\mathbf{e}_{r}^{s})^{T}\bigr)}
{\|{}^{E}\mathbf{p}^{s}-{}^{E}\mathbf{p}_{r}\|}
\label{eq:app_doppler_global_position}\\
\mathbf{J}_{\dot{\delta t}_{r}^{\gamma(s)}}
&= -c.
\label{eq:app_doppler_clock_drift}
\end{align}
The body-orientation Jacobian \eqref{eq:app_doppler_rotation} and the gyro-bias
Jacobian \eqref{eq:app_doppler_gyro_bias} both arise from the tangential lever-arm
velocity ${}_{i}^{w}\mathbf{R}\lfloor{}^{i}\boldsymbol{\omega}\rfloor_{\times}
{}^{i}\mathbf{l}_{r}$ of \eqref{eq:antenna_vel}, linear in
${}^{i}\boldsymbol{\omega}=\boldsymbol{\omega}_{m}-{}^{i}\mathbf{b}_{g}$;
the first (velocity-scale) term of \eqref{eq:app_doppler_rotation}, of magnitude
$O(\|{}^{i}\boldsymbol{\omega}\|\,\|{}^{i}\mathbf{l}_{r}\|)$, is the dominant
Doppler rotation coupling. The second terms of \eqref{eq:app_doppler_rotation}
and \eqref{eq:app_doppler_global_rotation} and the translation Jacobian
\eqref{eq:app_doppler_global_position} stem from the line-of-sight dependence on
the rotated antenna position; they share the magnitude
$O\bigl(\|{}^{E}\mathbf{v}^{s}-{}^{E}\mathbf{v}_{r}\|/\rho\bigr)
\sim 10^{-4}\,\mathrm{s}^{-1}$ and are small but retained for consistency with
the pseudorange Jacobians
\eqref{eq:app_dd_global_rotation}--\eqref{eq:app_dd_global_position}.

\newpage
\section{Single-Receiver Pseudorange Factor}
\label{app:single_receiver}
On datasets without a reference station (e.g.\ M3DGR), the double-differenced
pseudorange factor of Sec.~\ref{sec:dd_psr} is replaced by an undifferenced
single-receiver factor built directly on the raw pseudorange model
\eqref{eq:raw_pseudorange}. Because differencing no longer cancels the receiver
clock offset, the per-constellation offset
$\boldsymbol{\delta t}_{r}\in\mathbb{R}^{4}$ is reinstated as a navigation-state
variable in this variant (its drift rate $\boldsymbol{\dot{\delta t}}_{r}$ is
retained as before). The satellite clock $\delta t^{s}$ is applied feed-forward
from the broadcast ephemeris, and the ionospheric, tropospheric, and Sagnac
terms are corrected with standard broadcast/empirical models
($\hat{I}_{r}^{s}$ Klobuchar, $\hat{T}_{r}^{s}$ Saastamoinen, and
$\hat{\Omega}_{r}^{s}$), so the residual for satellite $s$ is
\begin{equation}
\begin{aligned}
r_{\rho}^{s}(\mathcal{X})
&= \|{}^{E}\mathbf{p}^{s}-{}^{E}\mathbf{p}_{r}\|+c\,\delta t_{r}^{\gamma(s)}
  -c\,\delta t^{s}\\
&\quad+\hat{I}_{r}^{s}+\hat{T}_{r}^{s}+\hat{\Omega}_{r}^{s}-\rho_{r}^{s},
\end{aligned}
\label{eq:app_single_receiver_residual}
\end{equation}
with ${}^{E}\mathbf{p}_{r}$ given by \eqref{eq:antenna_pos} and the residual
weighted by the same goGPS variance \eqref{eq:gogps_weight}. Its Jacobians with
respect to ${}_{i}^{w}\mathbf{p}$, ${}_{i}^{w}\mathbf{R}$, and
${}_{w}^{E}\mathbf{T}$ are those of
\eqref{eq:app_dd_position}--\eqref{eq:app_dd_global_position} with the differenced
line of sight $(\mathbf{e}_{r}^{m}-\mathbf{e}_{r}^{i})$ replaced by the single
line of sight $\mathbf{e}_{r}^{s}$, and the clock-offset Jacobian is
$\mathbf{J}_{\delta t_{r}^{\gamma(s)}}=c$. Since the absolute pseudorange
retains the residual atmospheric and orbit/clock errors that the double
difference would otherwise remove, this variant attains broadcast-corrected
single-point-positioning accuracy; the fused estimate nonetheless stays
sub-meter on the short M3DGR sequences because the LiDAR-inertial factors carry
the local geometry while GNSS only bounds the slow absolute drift
(Sec.~\ref{sec:public_benchmark}).

%% file: Bibliography.bib
@preamble{{
\providecommand{\GLIOBibReviewMark}[1]{{\color{red}#1}}
\long\def\GLIOBibReviewItem#1#2\par{\GLIOBibReviewOriginalItem{#1}\ifcsname GLIOBibReviewText@#1\endcsname\csname GLIOBibReviewText@#1\endcsname\else#2\fi\par}
\let\GLIOBibReviewOriginalBibliography\thebibliography
\def\thebibliography#1{\GLIOBibReviewOriginalBibliography{#1}\let\GLIOBibReviewOriginalItem\bibitem\let\bibitem\GLIOBibReviewItem}
\expandafter\def\csname GLIOBibReviewText@fastlio2\endcsname{W.~Xu, Y.~Cai, D.~He, J.~Lin, and F.~Zhang, ``{\GLIOBibReviewMark{FAST-LIO2}}: Fast direct
  {\GLIOBibReviewMark{LiDAR}}\GLIOBibReviewMark{-inertial} odometry,'' \emph{IEEE Transactions on Robotics}, vol.~38,
  no.~4, pp. 2053--2073, 2022.}
\expandafter\def\csname GLIOBibReviewText@glim\endcsname{K.~Koide, M.~Yokozuka, S.~Oishi, and A.~Banno, ``{\GLIOBibReviewMark{GLIM}}: {\GLIOBibReviewMark{3D}} range-inertial
  localization and mapping with {\GLIOBibReviewMark{GPU}}\GLIOBibReviewMark{-accelerated} scan matching factors,''
  \emph{Robotics and Autonomous Systems}, vol. 179, p. 104750, 2024.}
\expandafter\def\csname GLIOBibReviewText@lio-sam\endcsname{T.~Shan, B.~Englot, D.~Meyers, W.~Wang, C.~Ratti, and D.~Rus, ``{\GLIOBibReviewMark{LIO-SAM}}:
  Tightly-coupled {\GLIOBibReviewMark{LiDAR}} inertial odometry via smoothing and mapping,'' in
  \emph{\GLIOBibReviewMark{Proc.} IEEE/RSJ \GLIOBibReviewMark{Int. Conf. Intell. Robots Syst.} (IROS)}, 2020, pp.
  5135--5142.}
\expandafter\def\csname GLIOBibReviewText@zhang2016degeneracy\endcsname{\GLIOBibReviewMark{J.~Zhang, M.~Kaess, and S.~Singh, ``On degeneracy of optimization-based state
  estimation problems,'' in \emph{Proc. IEEE Int. Conf. Robot. Autom. (ICRA)},
  2016, pp. 809--816.}}
\expandafter\def\csname GLIOBibReviewText@GNSS_IMU_FGO_Welson\endcsname{W.~Wen, T.~Pfeifer, X.~Bai, and L.-t. Hsu, ``Factor graph optimization for
  {\GLIOBibReviewMark{GNSS}}\GLIOBibReviewMark{/}{\GLIOBibReviewMark{INS}} integration: A comparison with the extended kalman filter,''
  \emph{Navigation}, vol.~68, no.~2, pp. 315--331, 2021.}
\expandafter\def\csname GLIOBibReviewText@isam2\endcsname{M.~Kaess, H.~Johannsson, R.~Roberts, V.~Ila, J.~Leonard, and F.~Dellaert,
  ``{\GLIOBibReviewMark{iSAM2}}: Incremental smoothing and mapping with fluid relinearization and
  incremental variable reordering,'' in \emph{\GLIOBibReviewMark{Proc}. IEEE \GLIOBibReviewMark{Int. Conf. Robot.
  Autom. (ICRA)}}, 2011, pp. 3281--3288.}
\expandafter\def\csname GLIOBibReviewText@glio\endcsname{X.~Liu, W.~Wen, and L.-T. Hsu, ``{\GLIOBibReviewMark{GLIO}}: Tightly-coupled {\GLIOBibReviewMark{GNSS}}\GLIOBibReviewMark{/}{\GLIOBibReviewMark{LiDAR}}\GLIOBibReviewMark{/}{\GLIOBibReviewMark{IMU}}
  integration for continuous and drift-free state estimation of intelligent
  vehicles in urban areas,'' \emph{IEEE Transactions on Intelligent Vehicles},
  vol.~9, no.~1, pp. 1412--1422, 2024.}
\expandafter\def\csname GLIOBibReviewText@ligo\endcsname{D.~He, H.~Li, and J.~Yin, ``{\GLIOBibReviewMark{LIGO}}: A tightly coupled {\GLIOBibReviewMark{LiDAR}}\GLIOBibReviewMark{-inertial-}{\GLIOBibReviewMark{GNSS}}
  odometry based on a hierarchy fusion framework for global localization with
  real-time mapping,'' \emph{IEEE Transactions on Robotics}, vol.~41, pp.
  1224--1244, 2025.}
\expandafter\def\csname GLIOBibReviewText@li2021gil\endcsname{X.~Li, H.~Wang, S.~Li, S.~Feng, X.~Wang, and J.~Liao, ``{\GLIOBibReviewMark{GIL}}: A tightly
  coupled {\GLIOBibReviewMark{GNSS}} {\GLIOBibReviewMark{PPP}}\GLIOBibReviewMark{/}{\GLIOBibReviewMark{INS}}\GLIOBibReviewMark{/}{\GLIOBibReviewMark{LiDAR}} method for precise vehicle navigation,''
  \emph{Satellite \GLIOBibReviewMark{Navigation}}, vol.~2, no.~1, p.~26, 2021.}
\expandafter\def\csname GLIOBibReviewText@hsu2021urbannav\endcsname{L.-T. Hsu, N.~Kubo, W.~Wen, W.~Chen, Z.~Liu, T.~Suzuki, and J.~Meguro,
  ``{\GLIOBibReviewMark{UrbanNav}}: An open-sourced multisensory dataset for benchmarking
  positioning algorithms designed for urban areas,'' in \emph{\GLIOBibReviewMark{Proc.} 34th \GLIOBibReviewMark{Int.
  Tech. Meeting Satellite Division Inst. Navigation} (ION GNSS+)}, 2021, pp.
  226--256.}
\expandafter\def\csname GLIOBibReviewText@MARS-LVIG\endcsname{H.~Li, Y.~Zou, N.~Chen, J.~Lin, X.~Liu, W.~Xu, C.~Zheng, R.~Li, D.~He, F.~Kong,
  \GLIOBibReviewMark{Y}.~\GLIOBibReviewMark{Cai, Z.~Liu, S.~Zhou, K.~Xue, and F.~Zhang}, ``{\GLIOBibReviewMark{MARS-LVIG}} dataset: A
  multi-sensor aerial robots {\GLIOBibReviewMark{SLAM}} dataset for {\GLIOBibReviewMark{LiDAR}}\GLIOBibReviewMark{-visual-inertial-}{\GLIOBibReviewMark{GNSS}}
  fusion,'' \emph{The International Journal of Robotics Research}, vol.~43,
  no.~8, pp. 1114--1127, 2024.}
\expandafter\def\csname GLIOBibReviewText@zhang2025towards\endcsname{D.~Zhang, J.~Zhang, Y.~Sun, T.~Li, H.~Yin, H.~Xie, and J.~Yin, ``Towards robust
  sensor-fusion ground {\GLIOBibReviewMark{SLAM}}: A comprehensive benchmark and a resilient
  framework,'' in \emph{\GLIOBibReviewMark{Proc.} IEEE/RSJ \GLIOBibReviewMark{Int. Conf. Intell.} Robots \GLIOBibReviewMark{Syst.} (IROS)},
  2025, pp. 8894--8901.}
\expandafter\def\csname GLIOBibReviewText@deschaud2018imls\endcsname{J.-E. Deschaud, ``{\GLIOBibReviewMark{IMLS-SLAM}}: Scan-to-model matching based on {\GLIOBibReviewMark{3D}} data,'' in
  \emph{\GLIOBibReviewMark{Proc.} IEEE \GLIOBibReviewMark{Int. Conf. Robot. Autom.} (ICRA)}, 2018, pp. 2480--2485.}
\expandafter\def\csname GLIOBibReviewText@behley2018suma\endcsname{J.~Behley and C.~Stachniss, ``Efficient surfel-based {SLAM} using {3D} laser
  range data in urban environments,'' in \emph{\GLIOBibReviewMark{Proc.} Robotics: Science and
  Systems (RSS)}, 2018.}
\expandafter\def\csname GLIOBibReviewText@point-lio\endcsname{D.~He, W.~Xu, N.~Chen, F.~Kong, C.~Yuan, and F.~Zhang, ``{\GLIOBibReviewMark{Point-LIO}}: robust
  high-bandwidth light detection and ranging inertial odometry,''
  \emph{Advanced Intelligent Systems}, vol.~5, no.~7, p. 2200459, 2023.}
\expandafter\def\csname GLIOBibReviewText@icp\endcsname{P.~J. Besl and N.~D. McKay, ``Method for registration of {\GLIOBibReviewMark{3-D}} shapes,'' in
  \emph{\GLIOBibReviewMark{Proc. SPIE} Sensor \GLIOBibReviewMark{Fusion} IV: \GLIOBibReviewMark{Control Paradigms} and \GLIOBibReviewMark{Data Structures}},
  vol. 1611, 1992, pp. 586--606.}
\expandafter\def\csname GLIOBibReviewText@loam\endcsname{J.~Zhang \GLIOBibReviewMark{and} S.~Singh, ``{\GLIOBibReviewMark{LOAM}}: {\GLIOBibReviewMark{LiDAR}} odometry and mapping in real-time\GLIOBibReviewMark{,}''
  in \emph{\GLIOBibReviewMark{Proc.} Robotics: Science and \GLIOBibReviewMark{Systems (RSS)}}, 2014.}
\expandafter\def\csname GLIOBibReviewText@lu1997globally\endcsname{F.~Lu and E.~Milios, ``Globally consistent range scan alignment for environment
  mapping,'' \emph{Autonomous \GLIOBibReviewMark{Robots}}, vol.~4, no.~4, pp. 333--349, 1997.}
\expandafter\def\csname GLIOBibReviewText@nscan_matching\endcsname{P.~Biber and W.~Strasser, ``{\GLIOBibReviewMark{nScan-matching}}: Simultaneous matching of multiple
  scans and application to {\GLIOBibReviewMark{SLAM}},'' in \emph{\GLIOBibReviewMark{Proc}. IEEE \GLIOBibReviewMark{Int. Conf. Robot.
  Autom. (ICRA)}}, 2006, pp. 2270--2276.}
\expandafter\def\csname GLIOBibReviewText@gicp\endcsname{A.~Segal, D.~Haehnel, \GLIOBibReviewMark{and} S.~Thrun, ``{\GLIOBibReviewMark{Generalized-ICP}}\GLIOBibReviewMark{,}'' in \emph{\GLIOBibReviewMark{Proc.}
  Robotics: \GLIOBibReviewMark{Science} and \GLIOBibReviewMark{Systems (RSS)}}, 2009.}
\expandafter\def\csname GLIOBibReviewText@biber2003ndt\endcsname{P.~Biber and W.~Stra{\ss}er, ``The normal distributions transform: A new
  approach to laser scan matching,'' in \emph{\GLIOBibReviewMark{Proc.} IEEE/RSJ \GLIOBibReviewMark{Int. Conf. Intell.}
  Robots \GLIOBibReviewMark{Syst.} (IROS)}, vol.~3, 2003, pp. 2743--2748.}
\expandafter\def\csname GLIOBibReviewText@vgicp\endcsname{K.~Koide, M.~Yokozuka, S.~Oishi, and A.~Banno, ``Voxelized {\GLIOBibReviewMark{GICP}} for fast and
  accurate {\GLIOBibReviewMark{3D}} point cloud registration,'' in \emph{\GLIOBibReviewMark{Proc.} IEEE \GLIOBibReviewMark{Int. Conf.
  Robot. Autom.} (ICRA)}, 2021, pp. 11\,054--11\,059.}
\expandafter\def\csname GLIOBibReviewText@liu2021balm\endcsname{Z.~Liu and F.~Zhang, ``{BALM}: Bundle adjustment for {\GLIOBibReviewMark{LiDAR}} mapping,''
  \emph{IEEE Robotics and Automation Letters}, vol.~6, no.~2, pp. 3184--3191,
  2021.}
\expandafter\def\csname GLIOBibReviewText@fflins\endcsname{H.~Tang, T.~Zhang, X.~Niu, L.~Wang, L.~Wei, and J.~Liu, ``{\GLIOBibReviewMark{FF-LINS}}: A
  consistent frame-to-frame \GLIOBibReviewMark{solid-state-}{\GLIOBibReviewMark{LiDAR}}\GLIOBibReviewMark{-inertial} state estimator,''
  \emph{IEEE Robotics and Automation Letters}, vol.~8, no.~12, pp. 8525--8532,
  2023.}
\expandafter\def\csname GLIOBibReviewText@koide2022globally\endcsname{K.~Koide, M.~Yokozuka, S.~Oishi, and A.~Banno, ``Globally consistent and
  tightly coupled {\GLIOBibReviewMark{3D}} {\GLIOBibReviewMark{LiDAR}} inertial mapping,'' in \emph{\GLIOBibReviewMark{Proc. IEEE Int.
  Conf. Robot. Autom.} (ICRA)}, 2022, pp. 5622--5628.}
\expandafter\def\csname GLIOBibReviewText@koide2021globally\endcsname{K.~Koide, M.~Yokozuka, S.~Oishi, and A.~Banno, ``Globally consistent {\GLIOBibReviewMark{3D}}
  {\GLIOBibReviewMark{LiDAR}} mapping with {\GLIOBibReviewMark{GPU}}\GLIOBibReviewMark{-accelerated} {\GLIOBibReviewMark{GICP}} matching cost factors,''
  \emph{IEEE Robotics and Automation Letters}, vol.~6, no.~4, pp. 8591--8598,
  2021.}
\expandafter\def\csname GLIOBibReviewText@cao2022gvins\endcsname{S.~Cao, X.~Lu, and S.~Shen, ``{\GLIOBibReviewMark{GVINS}}: Tightly coupled {\GLIOBibReviewMark{GNSS}}--visual--inertial
  fusion for smooth and consistent state estimation,'' \emph{IEEE Transactions
  on Robotics}, vol.~38, no.~4, pp. 2004--2021, 2022.}
\expandafter\def\csname GLIOBibReviewText@liu2021rawgnssvio\endcsname{J.~Liu, W.~Gao, and Z.~Hu, ``Optimization-based visual-inertial {\GLIOBibReviewMark{SLAM}} tightly
  coupled with raw {\GLIOBibReviewMark{GNSS}} measurements,'' in \emph{\GLIOBibReviewMark{Proc.} IEEE \GLIOBibReviewMark{Int. Conf. Robot.
  Autom.} (ICRA)}, 2021, pp. 11\,612--11\,618.}
\expandafter\def\csname GLIOBibReviewText@zhang2024gnssfgo\endcsname{H.~Zhang, C.-C. Chen, H.~Vallery, and T.~D. Barfoot, ``{\GLIOBibReviewMark{GNSS}}\GLIOBibReviewMark{/multisensor}
  fusion using continuous-time factor graph optimization for robust
  localization,'' \emph{IEEE Transactions on Robotics}, vol.~40, pp.
  4003--4023, 2024.}
\expandafter\def\csname GLIOBibReviewText@li2023p3-lins\endcsname{T.~Li, L.~Pei, Y.~Xiang, X.~Zuo, W.~Yu, and T.-K. Truong, ``{\GLIOBibReviewMark{P3-LINS}}: Tightly
  coupled {\GLIOBibReviewMark{PPP}}\GLIOBibReviewMark{-}{\GLIOBibReviewMark{GNSS}}\GLIOBibReviewMark{/}{\GLIOBibReviewMark{INS}}\GLIOBibReviewMark{/}{\GLIOBibReviewMark{LiDAR}} navigation system with effective
  initialization,'' \emph{IEEE Transactions on Instrumentation and
  Measurement}, vol.~72, pp. 1--13, 2023.}
\expandafter\def\csname GLIOBibReviewText@chang2005mlambda\endcsname{P.~Teunissen, P.~De~Jonge, and C.~Tiberius, ``The least-squares ambiguity
  decorrelation adjustment: its performance on short {\GLIOBibReviewMark{GPS}} baselines and short
  observation spans,'' \emph{Journal of \GLIOBibReviewMark{Geodesy}}, vol.~71, no.~10, pp.
  589--602, 1997.}
\expandafter\def\csname GLIOBibReviewText@glins\endcsname{J.~Liu, C.~Chi, X.~Zhang, B.~Zhang, D.~Li, X.~Zhan, and M.~H. Ang, ``{\GLIOBibReviewMark{GLINS}}:
  {\GLIOBibReviewMark{GNSS}}\GLIOBibReviewMark{-}{\GLIOBibReviewMark{LiDAR}}\GLIOBibReviewMark{-}{\GLIOBibReviewMark{INS}} integrated navigation system,'' \emph{IEEE Robotics and
  Automation Letters}, vol.~11, no.~5, pp. 6472--6479, 2026.}
\expandafter\def\csname GLIOBibReviewText@wang2024givl\endcsname{X.~Wang, X.~Li, H.~Yu, H.~Chang, Y.~Zhou, and S.~Li, ``{\GLIOBibReviewMark{GIVL-SLAM}}: A robust
  and high-precision {\GLIOBibReviewMark{SLAM}} system by tightly coupled {\GLIOBibReviewMark{GNSS}} {\GLIOBibReviewMark{RTK}}, inertial,
  vision, and {\GLIOBibReviewMark{LiDAR}},'' \emph{IEEE/ASME Transactions on Mechatronics},
  vol.~30, no.~2, pp. 1212--1223, 2024.}
\expandafter\def\csname GLIOBibReviewText@li2023fgo\endcsname{X.~Li, H.~Yu, X.~Wang, S.~Li, Y.~Zhou, and H.~Chang, ``{\GLIOBibReviewMark{FGO-GIL}}: Factor graph
  optimization-based {\GLIOBibReviewMark{GNSS}} {\GLIOBibReviewMark{RTK}}\GLIOBibReviewMark{/}{\GLIOBibReviewMark{INS}}\GLIOBibReviewMark{/}{\GLIOBibReviewMark{LiDAR}} tightly coupled integration for
  precise and continuous navigation,'' \emph{IEEE Sensors Journal}, vol.~23,
  no.~13, pp. 14\,534--14\,548, 2023.}
\expandafter\def\csname GLIOBibReviewText@wu2023lio\endcsname{W.~Wu, X.~Zhong, D.~Wu, B.~Chen, X.~Zhong, and Q.~Liu, ``{\GLIOBibReviewMark{LIO-Fusion}}:
  Reinforced {\GLIOBibReviewMark{LiDAR}} inertial odometry by effective fusion with
  {\GLIOBibReviewMark{GNSS}}\GLIOBibReviewMark{/relocalization} and wheel odometry,'' \emph{IEEE Robotics and
  Automation Letters}, vol.~8, no.~3, pp. 1571--1578, 2023.}
\expandafter\def\csname GLIOBibReviewText@he2020integrated\endcsname{G.~He, X.~Yuan, Y.~Zhuang, and H.~Hu, ``An integrated {\GLIOBibReviewMark{GNSS}}\GLIOBibReviewMark{/}{\GLIOBibReviewMark{LiDAR}}\GLIOBibReviewMark{-}{\GLIOBibReviewMark{SLAM}}
  pose estimation framework for large-scale map building in partially
  {\GLIOBibReviewMark{GNSS}}\GLIOBibReviewMark{-denied} environments,'' \emph{IEEE Transactions on Instrumentation and
  Measurement}, vol.~70, pp. 1--9, 2020.}
\expandafter\def\csname GLIOBibReviewText@zhang2021continuous\endcsname{J.~Zhang, W.~Wen, F.~Huang, X.~Chen, and L.-T. Hsu, ``Continuous {\GLIOBibReviewMark{GNSS}}\GLIOBibReviewMark{-}{\GLIOBibReviewMark{RTK}}
  aided by {\GLIOBibReviewMark{LiDAR}}\GLIOBibReviewMark{/inertial} odometry with intelligent {\GLIOBibReviewMark{GNSS}} selection in urban
  canyons,'' in \emph{\GLIOBibReviewMark{Proc.} 34th \GLIOBibReviewMark{Int. Tech.} Meeting Satellite Division \GLIOBibReviewMark{Inst.}
  Navigation (ION GNSS+)}, 2021, pp. 4198--4207.}
\expandafter\def\csname GLIOBibReviewText@chang2019gnss\endcsname{L.~Chang, X.~Niu, T.~Liu, J.~Tang, and C.~Qian, ``{\GLIOBibReviewMark{GNSS}}\GLIOBibReviewMark{/}{\GLIOBibReviewMark{INS}}\GLIOBibReviewMark{/}{\GLIOBibReviewMark{LiDAR}}\GLIOBibReviewMark{-}{\GLIOBibReviewMark{SLAM}}
  integrated navigation system based on graph optimization,'' \emph{Remote
  Sensing}, vol.~11, no.~9, p. 1009, 2019.}
\expandafter\def\csname GLIOBibReviewText@mascaro2018gomsf\endcsname{R.~Mascaro, L.~Teixeira, T.~Hinzmann, R.~Siegwart, and M.~Chli, ``{\GLIOBibReviewMark{GOMSF}}:
  Graph-optimization based multi-sensor fusion for robust {\GLIOBibReviewMark{UAV}} pose
  estimation,'' in \emph{\GLIOBibReviewMark{Proc.} IEEE \GLIOBibReviewMark{Int. Conf. Robot. Autom.} (ICRA)}, 2018, pp.
  1421--1428.}
\expandafter\def\csname GLIOBibReviewText@nubert2022graph\endcsname{J.~Nubert, S.~Khattak, and M.~Hutter, ``Graph-based multi-sensor fusion for
  consistent localization of autonomous construction robots,'' in \emph{\GLIOBibReviewMark{Proc.
  IEEE Int. Conf. Robot. Autom.} (ICRA)}, 2022, pp. 10\,048--10\,054.}
\expandafter\def\csname GLIOBibReviewText@li2022tightly\endcsname{S.~Li, S.~Wang, Y.~Zhou, Z.~Shen, and X.~Li, ``Tightly coupled integration of
  {\GLIOBibReviewMark{GNSS}}, {\GLIOBibReviewMark{INS}}, and {\GLIOBibReviewMark{LiDAR}} for vehicle navigation in urban environments,''
  \emph{IEEE Internet of Things Journal}, vol.~9, no.~24, pp. 24\,721--24\,735,
  2022.}
\expandafter\def\csname GLIOBibReviewText@10161522\endcsname{J.~Beuchert, M.~Camurri, and M.~Fallon, ``Factor graph fusion of raw {\GLIOBibReviewMark{GNSS}}
  sensing with {\GLIOBibReviewMark{IMU}} and {\GLIOBibReviewMark{LiDAR}} for precise robot localization without a base
  station,'' in \emph{\GLIOBibReviewMark{Proc.} IEEE \GLIOBibReviewMark{Int. Conf. Robot. Autom.} (ICRA)}, 2023, pp.
  8415--8421.}
\expandafter\def\csname GLIOBibReviewText@gogps_weighting\endcsname{A.~M. Herrera, H.~F. Suhandri, E.~Realini, M.~Reguzzoni, and M.~C. De~Lacy,
  ``{\GLIOBibReviewMark{goGPS}}: open-source {\GLIOBibReviewMark{MATLAB}} software,'' \emph{GPS \GLIOBibReviewMark{Solutions}}, vol.~20,
  no.~3, pp. 595--603, 2016.}
\expandafter\def\csname GLIOBibReviewText@rtklib\endcsname{T.~Takasu and A.~Yasuda, ``Development of the low-cost {\GLIOBibReviewMark{RTK-GPS}} receiver with
  an open source program package {\GLIOBibReviewMark{RTKLIB}},'' in \emph{\GLIOBibReviewMark{Proc.} International
  \GLIOBibReviewMark{Symposium} on GPS/GNSS}, \GLIOBibReviewMark{2009}.}
\expandafter\def\csname GLIOBibReviewText@vins-mono\endcsname{T.~Qin, P.~Li, and S.~Shen, ``{\GLIOBibReviewMark{VINS-MONO}}: A robust and versatile monocular
  visual-inertial state estimator,'' \emph{IEEE Transactions on Robotics},
  vol.~34, no.~4, pp. 1004--1020, 2018.}
\expandafter\def\csname GLIOBibReviewText@triggs1999bundle\endcsname{B.~Triggs, P.~F. McLauchlan, R.~I. Hartley, and A.~W. Fitzgibbon, ``Bundle
  adjustment\GLIOBibReviewMark{---}a modern synthesis,'' in \emph{\GLIOBibReviewMark{Vision Algorithms: Theory and
  Practice}}\GLIOBibReviewMark{, ser. Lecture Notes in Computer Science, B.~Triggs, A.~Zisserman,
  and R.~Szeliski, Eds}.\hskip 1em plus 0.5em minus 0.4em\relax Springer, \GLIOBibReviewMark{2000,
  vol. 1883}, pp. 298--372.}
\expandafter\def\csname GLIOBibReviewText@grisetti2010tutorial\endcsname{G.~Grisetti, R.~K{\"u}mmerle, C.~Stachniss, and W.~Burgard, ``A tutorial on
  graph-based {\GLIOBibReviewMark{SLAM}},'' \emph{IEEE Intelligent Transportation Systems
  Magazine}, vol.~2, no.~4, pp. 31--43, 2010.}
\expandafter\def\csname GLIOBibReviewText@kummerle2011g2o\endcsname{R.~K{\"u}mmerle, G.~Grisetti, H.~Strasdat, K.~Konolige, and W.~Burgard,
  ``{\GLIOBibReviewMark{g2o}}: A general framework for graph optimization,'' in \emph{\GLIOBibReviewMark{Proc}. IEEE
  \GLIOBibReviewMark{Int. Conf. Robot. Autom. (ICRA)}}, 2011, pp. 3607--3613.}
\expandafter\def\csname GLIOBibReviewText@olson2006fast\endcsname{E.~Olson, J.~Leonard, and S.~Teller, ``Fast iterative alignment of pose graphs
  with poor initial estimates,'' in \emph{\GLIOBibReviewMark{Proc}. IEEE \GLIOBibReviewMark{Int. Conf. Robot. Autom.
  (ICRA)}}, 2006, pp. 2262--2269.}
\expandafter\def\csname GLIOBibReviewText@rosen2019sesync\endcsname{D.~M. Rosen, L.~Carlone, A.~S. Bandeira, and J.~J. Leonard, ``{\GLIOBibReviewMark{SE-Sync}}: A
  certifiably correct algorithm for synchronization over the special euclidean
  group,'' \emph{The International Journal of Robotics Research}, vol.~38, no.
  2-3, pp. 95--125, 2019.}
\expandafter\def\csname GLIOBibReviewText@zhu2022robust\endcsname{F.~Zhu, Y.~Ren, and F.~Zhang, ``Robust real-time {\GLIOBibReviewMark{LiDAR}}\GLIOBibReviewMark{-inertial}
  initialization,'' in \emph{\GLIOBibReviewMark{Proc.} IEEE/RSJ \GLIOBibReviewMark{Int. Conf. Intell.} Robots \GLIOBibReviewMark{Syst.}
  (IROS)}, 2022, pp. 3948--3955.}
\expandafter\def\csname GLIOBibReviewText@tuna2024xicp\endcsname{T.~Tuna, J.~Nubert, Y.~Nava, S.~Khattak, and M.~Hutter, ``{\GLIOBibReviewMark{X-ICP}}:
  Localizability-aware {\GLIOBibReviewMark{LiDAR}} registration for robust localization in extreme
  environments,'' \emph{IEEE Transactions on Robotics}, vol.~40, pp. 452--471,
  2024.}
}}

@string{icra = {Proc. IEEE Int. Conf. Robot. Autom. (ICRA)}}

@string{iros = {Proc. IEEE/RSJ Int. Conf. Intell. Robots Syst. (IROS)}}

@string{rss = {Proc. Robotics: Science and Systems (RSS)}}

@string{iongnss34 = {Proc. 34th Int. Tech. Meeting Satellite Division Inst. Navigation (ION GNSS+)}}

@IEEEtranBSTCTL{IEEEexample:BSTcontrol,
  CTLuse_article_number = "yes",
  CTLuse_paper = "yes",
  CTLuse_url = "yes",
  CTLdash_repeated_names = "no",
  CTLuse_forced_etal = "no",
  CTLmax_names_forced_etal = "10",
  CTLnames_show_etal = "1"
}

@inproceedings{icp,
  title={Method for registration of {3-D} shapes},
  author={Besl, Paul J and McKay, Neil D},
  booktitle = {Proc. SPIE Sensor Fusion IV: Control Paradigms and Data Structures},
  volume={1611},
  pages={586--606},
  year={1992},
}

@inproceedings{loam,
  author = {Zhang, Ji and Singh, Sanjiv},
  title = {{LOAM}: {LiDAR} odometry and mapping in real-time},
  booktitle = rss,
  year = {2014},
  doi = {10.15607/RSS.2014.X.007}
}

@article{fastlio2,
  title={{FAST-LIO2}: Fast direct {LiDAR}-inertial odometry},
  author={Xu, Wei and Cai, Yixi and He, Dongjiao and Lin, Jiarong and Zhang, Fu},
  journal={IEEE Transactions on Robotics},
  volume={38},
  number={4},
  pages={2053--2073},
  year={2022},
}

@article{glio,
  author={Liu, Xikun and Wen, Weisong and Hsu, Li-Ta},
  journal={IEEE Transactions on Intelligent Vehicles}, 
  title={{GLIO}: Tightly-Coupled {GNSS}/{LiDAR}/{IMU} Integration for Continuous and Drift-Free State Estimation of Intelligent Vehicles in Urban Areas}, 
  year={2024},
  volume={9},
  number={1},
  pages={1412--1422},
  doi={10.1109/TIV.2023.3323648}}

@article{ligo,
  title={{LIGO}: A tightly coupled {LiDAR}-inertial-{GNSS} odometry based on a hierarchy fusion framework for global localization with real-time mapping},
  author={He, Dongjiao and Li, Haotian and Yin, Jie},
  journal={IEEE Transactions on Robotics},
  volume={41},
  pages={1224--1244},
  year={2025},
}

@article{glim,
  title={{GLIM}: {3D} range-inertial localization and mapping with {GPU}-accelerated scan matching factors},
  author={Koide, Kenji and Yokozuka, Masashi and Oishi, Shuji and Banno, Atsuhiko},
  journal={Robotics and Autonomous Systems},
  volume={179},
  pages={104750},
  year={2024},
}

@inproceedings{lio-sam,
  title={{LIO-SAM}: Tightly-coupled {LiDAR} inertial odometry via smoothing and mapping},
  author={Shan, Tixiao and Englot, Brendan and Meyers, Drew and Wang, Wei and Ratti, Carlo and Rus, Daniela},
  booktitle = iros,
  pages={5135--5142},
  year={2020},
}

@inproceedings{zhang2016degeneracy,
  title={On degeneracy of optimization-based state estimation problems},
  author={Zhang, Ji and Kaess, Michael and Singh, Sanjiv},
  booktitle = icra,
  pages={809--816},
  year={2016},
}

@article{GNSS_IMU_FGO_Welson,
  title={Factor graph optimization for {GNSS}/{INS} integration: A comparison with the extended Kalman filter},
  author={Wen, Weisong and Pfeifer, Tim and Bai, Xiwei and Hsu, Li-ta},
  journal={Navigation},
  volume={68},
  number={2},
  pages={315--331},
  year={2021},
}

@article{fflins,
  title={{FF-LINS}: A consistent frame-to-frame solid-state-{LiDAR}-inertial state estimator},
  author={Tang, Hailiang and Zhang, Tisheng and Niu, Xiaoji and Wang, Liqiang and Wei, Linfu and Liu, Jingnan},
  journal={IEEE Robotics and Automation Letters},
  volume={8},
  number={12},
  pages={8525--8532},
  year={2023},
}

@article{koide2021globally,
  title={Globally consistent {3D} {LiDAR} mapping with {GPU}-accelerated {GICP} matching cost factors},
  author={Koide, Kenji and Yokozuka, Masashi and Oishi, Shuji and Banno, Atsuhiko},
  journal={IEEE Robotics and Automation Letters},
  volume={6},
  number={4},
  pages={8591--8598},
  year={2021},
}

@inproceedings{koide2022globally,
  title={Globally consistent and tightly coupled {3D} {LiDAR} inertial mapping},
  author={Koide, Kenji and Yokozuka, Masashi and Oishi, Shuji and Banno, Atsuhiko},
  booktitle = icra,
  pages={5622--5628},
  year={2022},
}

@inproceedings{gicp,
  author = {Segal, Aleksandr and Haehnel, Dirk and Thrun, Sebastian},
  title = {{Generalized-ICP}},
  booktitle = rss,
  year = {2009},
  doi = {10.15607/RSS.2009.V.021}
}

@article{dellaert2017factor,
  title={Factor graphs for robot perception},
  author={Dellaert, Frank and Kaess, Michael},
  journal={Foundations and Trends in Robotics},
  volume={6},
  number={1-2},
  pages={1--139},
  year={2017}
}

@article{vins-mono,
  title={{VINS-MONO}: A robust and versatile monocular visual-inertial state estimator},
  author={Qin, Tong and Li, Peiliang and Shen, Shaojie},
  journal={IEEE Transactions on Robotics},
  volume={34},
  number={4},
  pages={1004--1020},
  year={2018},
}

@article{gogps_weighting,
  title={{goGPS}: open-source {MATLAB} software},
  author={Herrera, Antonio M and Suhandri, Hendy F and Realini, Eugenio and Reguzzoni, Mirko and De Lacy, M Clara},
  journal={GPS Solutions},
  volume={20},
  number={3},
  pages={595--603},
  year={2016},
}

@article{chang2005mlambda,
  title={The least-squares ambiguity decorrelation adjustment: its performance on short {GPS} baselines and short observation spans},
  author={Teunissen, PJG and De Jonge, PJ and Tiberius, CCJM},
  journal={Journal of Geodesy},
  volume={71},
  number={10},
  pages={589--602},
  year={1997},
}

@INPROCEEDINGS{isam2,
  title={{iSAM2}: Incremental smoothing and mapping with fluid relinearization and incremental variable reordering},
  author={Kaess, Michael and Johannsson, Hordur and Roberts, Richard and Ila, Viorela and Leonard, John and Dellaert, Frank},
  booktitle = icra,
  pages={3281--3288},
  year={2011},
}

@article{point-lio,
  title={{Point-LIO}: robust high-bandwidth light detection and ranging inertial odometry},
  author={He, Dongjiao and Xu, Wei and Chen, Nan and Kong, Fanze and Yuan, Chongjian and Zhang, Fu},
  journal={Advanced Intelligent Systems},
  volume={5},
  number={7},
  pages={2200459},
  year={2023},
}

@INPROCEEDINGS{vgicp,
  title={Voxelized {GICP} for fast and accurate {3D} point cloud registration},
  author={Koide, Kenji and Yokozuka, Masashi and Oishi, Shuji and Banno, Atsuhiko},
  booktitle = icra,
  pages={11054--11059},
  year={2021},
}

@inproceedings{zhang2021continuous,
  title={Continuous {GNSS}-{RTK} aided by {LiDAR}/inertial odometry with intelligent {GNSS} selection in urban canyons},
  author={Zhang, Jiachen and Wen, Weisong and Huang, Feng and Chen, Xiaodong and Hsu, Li-Ta},
  booktitle = iongnss34,
  pages={4198--4207},
  year={2021}
}

@article{wu2023lio,
  title={{LIO-Fusion}: Reinforced {LiDAR} inertial odometry by effective fusion with {GNSS}/relocalization and wheel odometry},
  author={Wu, Wenhong and Zhong, Xunyu and Wu, Dongjie and Chen, Bushi and Zhong, Xungao and Liu, Qiang},
  journal={IEEE Robotics and Automation Letters},
  volume={8},
  number={3},
  pages={1571--1578},
  year={2023},
}

@article{he2020integrated,
  title={An integrated {GNSS}/{LiDAR}-{SLAM} pose estimation framework for large-scale map building in partially {GNSS}-denied environments},
  author={He, Guojian and Yuan, Xingda and Zhuang, Yan and Hu, Huosheng},
  journal={IEEE Transactions on Instrumentation and Measurement},
  volume={70},
  pages={1--9},
  year={2020},
}

@article{chang2019gnss,
  title={{GNSS}/{INS}/{LiDAR}-{SLAM} integrated navigation system based on graph optimization},
  author={Chang, Le and Niu, Xiaoji and Liu, Tianyi and Tang, Jian and Qian, Chuang},
  journal={Remote Sensing},
  volume={11},
  number={9},
  pages={1009},
  year={2019},
}

@article{wang2024givl,
  title={{GIVL-SLAM}: A robust and high-precision {SLAM} system by tightly coupled {GNSS} {RTK}, inertial, vision, and {LiDAR}},
  author={Wang, Xuanbin and Li, Xingxing and Yu, Hui and Chang, Hanyu and Zhou, Yuxuan and Li, Shengyu},
  journal={IEEE/ASME Transactions on Mechatronics},
  volume={30},
  number={2},
  pages={1212--1223},
  year={2024},
}

@article{li2021gil,
  title={{GIL}: A tightly coupled {GNSS} {PPP}/{INS}/{LiDAR} method for precise vehicle navigation},
  author={Li, Xingxing and Wang, Huidan and Li, Shengyu and Feng, Shaoquan and Wang, Xuanbin and Liao, Jianchi},
  journal={Satellite Navigation},
  volume={2},
  number={1},
  pages={26},
  year={2021},
}

@article{li2022tightly,
  title={Tightly coupled integration of {GNSS}, {INS}, and {LiDAR} for vehicle navigation in urban environments},
  author={Li, Shengyu and Wang, Shiwen and Zhou, Yuxuan and Shen, Zhiheng and Li, Xingxing},
  journal={IEEE Internet of Things Journal},
  volume={9},
  number={24},
  pages={24721--24735},
  year={2022},
}

@article{li2023fgo,
  title={{FGO-GIL}: Factor graph optimization-based {GNSS} {RTK}/{INS}/{LiDAR} tightly coupled integration for precise and continuous navigation},
  author={Li, Xingxing and Yu, Hui and Wang, Xuanbin and Li, Shengyu and Zhou, Yuxuan and Chang, Hanyu},
  journal={IEEE Sensors Journal},
  volume={23},
  number={13},
  pages={14534--14548},
  year={2023},
}

@article{li2023p3-lins,
  title={{P3-LINS}: Tightly coupled {PPP}-{GNSS}/{INS}/{LiDAR} navigation system with effective initialization},
  author={Li, Tao and Pei, Ling and Xiang, Yan and Zuo, Xingxing and Yu, Wenxian and Truong, Trieu-Kien},
  journal={IEEE Transactions on Instrumentation and Measurement},
  volume={72},
  pages={1--13},
  year={2023},
}

@INPROCEEDINGS{10161522,
  author={Beuchert, Jonas and Camurri, Marco and Fallon, Maurice},
  booktitle = icra,
  title={Factor Graph Fusion of Raw {GNSS} Sensing with {IMU} and {LiDAR} for Precise Robot Localization without a Base Station}, 
  year={2023},
  pages={8415--8421},
}

@INPROCEEDINGS{nscan_matching,
  title={{nScan-matching}: Simultaneous matching of multiple scans and application to {SLAM}},
  author={Biber, Peter and Strasser, Wolfgang},
  booktitle = icra,
  pages={2270--2276},
  year={2006},
}

@article{glins,
  author={Liu, Jiahui and Chi, Cheng and Zhang, Xin and Zhang, Binlin and Li, Dongen and Zhan, Xingqun and Ang, Marcelo H.},
  journal={IEEE Robotics and Automation Letters}, 
  title={{GLINS}: {GNSS}-{LiDAR}-{INS} Integrated Navigation System}, 
  year={2026},
  volume={11},
  number={5},
  pages={6472--6479},
  doi={10.1109/LRA.2026.3681161}}

@INPROCEEDINGS{liu2021rawgnssvio,
  title={Optimization-based visual-inertial {SLAM} tightly coupled with raw {GNSS} measurements},
  author={Liu, Jinxu and Gao, Wei and Hu, Zhanyi},
  booktitle = icra,
  pages={11612--11618},
  year={2021},
}

@article{cao2022gvins,
  title={{GVINS}: Tightly coupled {GNSS}--visual--inertial fusion for smooth and consistent state estimation},
  author={Cao, Shaozu and Lu, Xiuyuan and Shen, Shaojie},
  journal={IEEE Transactions on Robotics},
  volume={38},
  number={4},
  pages={2004--2021},
  year={2022},
}

@inproceedings{zhu2022robust,
  title={Robust real-time {LiDAR}-inertial initialization},
  author={Zhu, Fangcheng and Ren, Yunfan and Zhang, Fu},
  booktitle = iros,
  pages={3948--3955},
  year={2022},
}

@inproceedings{zhang2025towards,
  title={Towards robust sensor-fusion ground {SLAM}: A comprehensive benchmark and a resilient framework},
  author={Zhang, Deteng and Zhang, Junjie and Sun, Yan and Li, Tao and Yin, Hao and Xie, Hongzhao and Yin, Jie},
  booktitle = iros,
  pages={8894--8901},
  year={2025},
}

@inproceedings{hsu2021urbannav,
  title={{UrbanNav}: An open-sourced multisensory dataset for benchmarking positioning algorithms designed for urban areas},
  author={Hsu, Li-Ta and Kubo, Nobuaki and Wen, Weisong and Chen, Wu and Liu, Zhizhao and Suzuki, Taro and Meguro, Junichi},
  booktitle = iongnss34,
  pages={226--256},
  year={2021}
}

@inproceedings{rtklib,
  author = {Takasu, Tomoji and Yasuda, Akio},
  title = {Development of the low-cost {RTK-GPS} receiver with an open source program package {RTKLIB}},
  booktitle = {Proc. International Symposium on GPS/GNSS},
  year = {2009}
}

@article{enge1994global,
  title={The global positioning system: Signals, measurements, and performance},
  author={Enge, Per K},
  journal={International Journal of Wireless Information Networks},
  volume={1},
  number={2},
  pages={83--105},
  year={1994},
}

@article{forster2017preintegration,
  author={Forster, Christian and Carlone, Luca and Dellaert, Frank and Scaramuzza, Davide},
  journal={IEEE Transactions on Robotics}, 
  title={On-Manifold Preintegration for Real-Time Visual--Inertial Odometry}, 
  year={2017},
  volume={33},
  number={1},
  pages={1--21},
  doi={10.1109/TRO.2016.2597321}
}

@article{qin2019general,
  title={A General Optimisation-Based Framework for Global Pose Estimation With Multiple Sensors},
  author={Qin, Tong and Cao, Shaozu and Pan, Jie and Shen, Shaojie},
  journal={IET Cyber-Systems and Robotics},
  volume={7},
  number={1},
  pages={e70023},
  year={2025},
}

@inproceedings{mascaro2018gomsf,
  title={{GOMSF}: Graph-optimization based multi-sensor fusion for robust {UAV} pose estimation},
  author={Mascaro, Ruben and Teixeira, Lucas and Hinzmann, Timo and Siegwart, Roland and Chli, Margarita},
  booktitle = icra,
  pages={1421--1428},
  year={2018},
}

@inproceedings{nubert2022graph,
  title={Graph-based multi-sensor fusion for consistent localization of autonomous construction robots},
  author={Nubert, Julian and Khattak, Shehryar and Hutter, Marco},
  booktitle = icra,
  pages={10048--10054},
  year={2022},
}

@incollection{moore2016generalized,
author="Moore, Thomas
and Stouch, Daniel",
title="A Generalized Extended Kalman Filter Implementation for the Robot Operating System",
booktitle="Intelligent Autonomous Systems 13",
year="2016",
publisher="Springer International Publishing",
address="Cham",
pages="335--348",
}

@article{tuna2024xicp,
  author={Tuna, Turcan and Nubert, Julian and Nava, Yoshua and Khattak, Shehryar and Hutter, Marco},
  journal={IEEE Transactions on Robotics}, 
  title={{X-ICP}: Localizability-Aware {LiDAR} Registration for Robust Localization in Extreme Environments}, 
  year={2024},
  volume={40},
  pages={452--471},
  doi={10.1109/TRO.2023.3335691}}

@article{sola2018micro,
  title={A micro lie theory for state estimation in robotics},
  author={Sola, Joan and Deray, Jeremie and Atchuthan, Dinesh},
  journal={arXiv preprint arXiv:1812.01537},
  year={2018}
}

@incollection{triggs1999bundle,
  author = {Triggs, Bill and McLauchlan, Philip F. and Hartley, Richard I. and Fitzgibbon, Andrew W.},
  title = {Bundle adjustment---A modern synthesis},
  editor = {Triggs, Bill and Zisserman, Andrew and Szeliski, Richard},
  booktitle = {Vision Algorithms: Theory and Practice},
  series = {Lecture Notes in Computer Science},
  volume = {1883},
  pages = {298--372},
  year = {2000},
  publisher = {Springer},
  doi = {10.1007/3-540-44480-7_21}
}

@article{grisetti2010tutorial,
  title={A tutorial on graph-based {SLAM}},
  author={Grisetti, Giorgio and K{\"u}mmerle, Rainer and Stachniss, Cyrill and Burgard, Wolfram},
  journal={IEEE Intelligent Transportation Systems Magazine},
  volume={2},
  number={4},
  pages={31--43},
  year={2010},
}

@inproceedings{kummerle2011g2o,
  title={{g2o}: A general framework for graph optimization},
  author={K{\"u}mmerle, Rainer and Grisetti, Giorgio and Strasdat, Hauke and Konolige, Kurt and Burgard, Wolfram},
  booktitle = icra,
  pages={3607--3613},
  year={2011},
}

@inproceedings{olson2006fast,
  title={Fast iterative alignment of pose graphs with poor initial estimates},
  author={Olson, Edwin and Leonard, John and Teller, Seth},
  booktitle = icra,
  pages={2262--2269},
  year={2006},
}

@article{rosen2019sesync,
  title={{SE-Sync}: A certifiably correct algorithm for synchronization over the special Euclidean group},
  author={Rosen, David M and Carlone, Luca and Bandeira, Afonso S and Leonard, John J},
  journal={The International Journal of Robotics Research},
  volume={38},
  number={2-3},
  pages={95--125},
  year={2019},
}

@article{MARS-LVIG,
  author = {Li, Haotian and Zou, Yuying and Chen, Nan and Lin, Jiarong and Liu, Xiyuan and Xu, Wei and Zheng, Chunran and Li, Rundong and He, Dongjiao and Kong, Fanze and Cai, Yixi and Liu, Zheng and Zhou, Shunbo and Xue, Kaiwen and Zhang, Fu},
  title = {{MARS-LVIG} dataset: A multi-sensor aerial robots {SLAM} dataset for {LiDAR}-visual-inertial-{GNSS} fusion},
  journal = {The International Journal of Robotics Research},
  volume = {43},
  number = {8},
  pages = {1114--1127},
  year = {2024},
  doi = {10.1177/02783649241227968}
}

@article{zhang2024gnssfgo,
  title={{GNSS}/multisensor fusion using continuous-time factor graph optimization for robust localization},
  author={Zhang, Haoming and Chen, Chih-Chun and Vallery, Heike and Barfoot, Timothy D},
  journal={IEEE Transactions on Robotics},
  volume={40},
  pages={4003--4023},
  year={2024},
}

@article{lu1997globally,
  title={Globally consistent range scan alignment for environment mapping},
  author={Lu, Feng and Milios, Evangelos},
  journal={Autonomous Robots},
  volume={4},
  number={4},
  pages={333--349},
  year={1997},
}

@inproceedings{deschaud2018imls,
  title={{IMLS-SLAM}: Scan-to-model matching based on {3D} data},
  author={Deschaud, Jean-Emmanuel},
  booktitle = icra,
  pages={2480--2485},
  year={2018},
}

@inproceedings{behley2018suma,
  title={Efficient surfel-based {SLAM} using {3D} laser range data in urban environments},
  author={Behley, Jens and Stachniss, Cyrill},
  booktitle = rss,
  year={2018}
}

@inproceedings{biber2003ndt,
  title={The normal distributions transform: A new approach to laser scan matching},
  author={Biber, Peter and Stra{\ss}er, Wolfgang},
  booktitle = iros,
  volume={3},
  pages={2743--2748},
  year={2003},
}

@article{liu2021balm,
  title={{BALM}: Bundle adjustment for {LiDAR} mapping},
  author={Liu, Zheng and Zhang, Fu},
  journal={IEEE Robotics and Automation Letters},
  volume={6},
  number={2},
  pages={3184--3191},
  year={2021},
}

@incollection{odijk2017differential,
  title={Differential positioning},
  author={Odijk, Dennis and Wanninger, Lambert},
  booktitle={Springer handbook of global navigation satellite systems},
  pages={753--780},
  year={2017},
  publisher={Springer}
}
